Karlsruhe Institute of Technology

# Fully Convolutional Neural Networks for Dynamic Object Detection in Grid Maps

Master's Thesis of

Florian Pierre Joseph Piewak

Department of Computer Science
Institute for Anthropomatics
and
FZI Research Center for Information Technology

| | |
|---|---|
| Reviewer: | Prof. Dr.–Ing. J. M. Zöllner |
| Second reviewer: | Prof. Dr.–Ing. R. Dillmann |
| Advisors: | Dipl.–Inform. Michael Weber |
| | Dipl.–Ing. Timo Rehfeld |

Research Period: December 13th, 2015 – June 12th, 2016

# Fully Convolutional Neural Networks for Dynamic Object Detection in Grid Maps

by

Florian Pierre Joseph Piewak

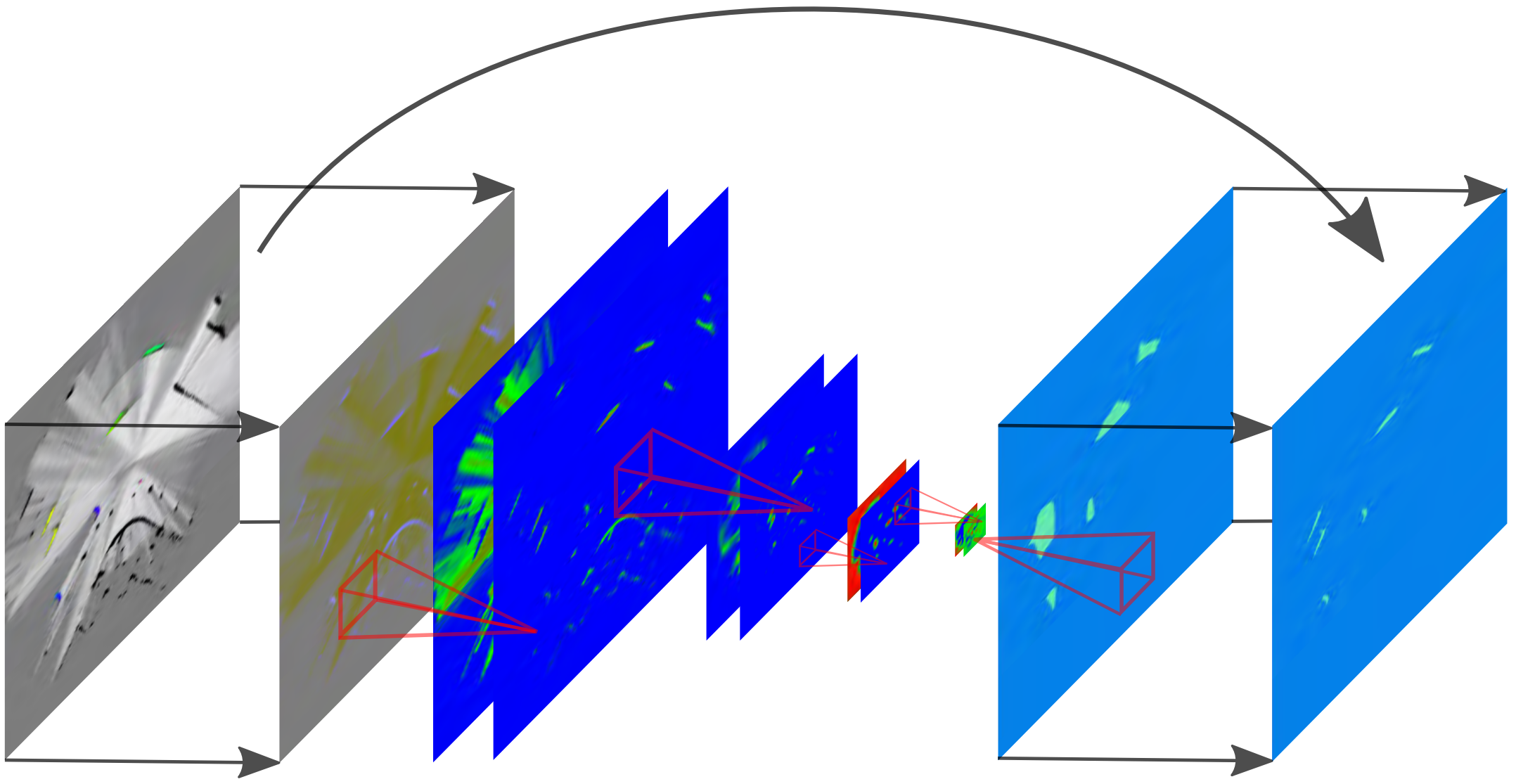

**Master's Thesis**

June 2016

Mercedes-Benz
Research & Development North America, Inc.

FZI

Department Technical Cognitive Systems
FZI Research Center for Information Technology

## Affirmation

Ich versichere wahrheitsgemäß, die Arbeit selbstständig verfasst, alle benutzten Hilfsmittel vollständig und genau angegeben und alles kenntlich gemacht zu haben, was aus Arbeiten anderer unverändert oder mit Abänderungen entnommen wurde sowie die Satzung des KIT zur Sicherung guter wissenschaftlicher Praxis in der jeweils gültigen Fassung beachtet zu haben.

Sunnyvale,
June 2016

*Florian Pierre Joseph Piewak*

# Abstract

One of the most important parts of environment perception is the detection of obstacles in the surrounding of the vehicle. To achieve that, several sensors like radars, LiDARs and cameras are installed in autonomous vehicles. The produced sensor data is fused to a general representation of the surrounding. In this thesis the dynamic occupancy grid map approach of Nuss et al. [37] is used while three goals are achieved. First, the approach of Nuss et al. [36] to distinguish between moving and non-moving obstacles is improved by using Fully Convolutional Neural Networks to create a class prediction for each grid cell. For this purpose, the network is initialized with public pre-trained network models and the training is executed with a semi-automatic generated dataset. The pre-trained Convolutional Neural Networks are trained on colored images, which is another kind of input as used in this thesis. For this reason, several parameters of the Fully Convolutional Neural Networks like the network structure, different combination of inputs and the weights of labels are optimized. The second goal is to provide orientation information for each detected moving obstacle. This could improve tracking algorithms, which are based on the dynamic occupancy grid map. The orientation extraction based on the Convolutional Neural Network shows a better performance in comparison to an orientation extraction directly over the velocity information of the dynamic occupancy grid map. A general problem of developing machine learning approaches like Neural Networks is the number of labeled data, which can always be increased. For this reason, the last goal is to evaluate a semi-supervised learning algorithm, to generate automatically more labeled data. The result of this evaluation shows that the automated labeled data does not improve the performance of the Convolutional Neural Network. All in all, the best results are combined to compare the detection against the approach of Nuss et al. [36] and a relative improvement of 34.8% is reached.

# Zusammenfassung

Ein elementarer Bestandteil der Umgebungswahrnehmung eines Fahrzeuges, ist die Erkennung von Verkehrsteilnehmern. Dies wird mittels unterschiedlicher Sensoren, wie zum Beispiel Radar, LiDAR oder Kamera realisiert, wobei die aufgenommen Daten zu einer allgemeinen Umgebungsrepräsentation fusioniert werden. In dieser Arbeit wird die Umgebungsrepräsentation von Nuss et al. [37] genutzt, welche eine Implementierung einer dynamischen Belegungsrasterkarte vorstellen. Die vorliegende Arbeit besteht aus drei Teilen. Als Erstes wird der Ansatz von Nuss et al. [36] zum Unterscheiden zwischen bewegten und nicht bewegten Objekten durch das Einsetzen von Convolutional Neural Networks (CNNs) erweitert. Dabei besteht die Ausgabe des CNNs aus einer Klassifikation für jede Rasterzelle. Die Initialisierung der CNNs erfolgt durch publizierte vortrainierte Netze, welche anschließend durch semi-automatisch erzeugte Daten trainiert werden. Die publizierten vortrainierten Netze sind mit farbigen Eingabebildern trainiert wurden, was einen anderen Eingabetyp als in dieser Arbeit darstellt. Aus diesem Grund werden diverse Parameter des CNNs, wie zum Beispiel die Netzwerkstruktur, die Kombination der Eingabedaten oder die Gewichtung der Trainingsklassen, optimiert. Als zweiter Teil wird die Orientierungsextraktion der erkannten Objekte betrachtet, welche bestehende (auf der dynamischen Belegungsrasterkarte basierende) Tracking-Algorithmen verbessern kann. Das Ergebnis der Orientierungsextraktion zeigt eine leichte Verbesserung basierend auf dem CNN im Vergleich zur Extraktion der Orientierung direkt aus der dynamischen Belegungsrasterkarte. Eine generelle Herausforderung beim Entwickeln von Maschinellen Lernen Ansätze ist die limitierte Anzahl verfügbarer Trainingsdaten. Aus diesem Grund ist der dritte Teil der Arbeit die Evaluation eines semi-überwachten Lernalgorithmus, welcher automatisch zusätzliche Daten klassifiziert und damit die Anzahl verfügbarer Trainingsdaten vergrößert. Das Ergebnis zeigt keine Verbesserung des CNNs bei Verwendung der automatisch erzeugten Trainingsdaten. Zum Schluss werden die besten Ergebnisse der Optimierungen aus dem ersten Teil vereint und mit dem Ansatz von Nuss et al. [36] verglichen, wobei eine relative Verbesserung von $34,8\%$ erreicht wird.

## Acknowledgment

First, I would like to thank all, who supported and motivated me during my master thesis.

I would also like to acknowledge my supervising professor Prof. Dr.–Ing. J. M. Zöllner, who supervised me on the part of the Karlsruhe Institute of Technology (KIT) and proposed to write my thesis at Mercedes-Benz Research and Development North America (MBRDNA). In several deep discussions, he had passionate participation and motivating input to develop the topic of my thesis.

Additionally, I would like to thank my on-site supervisor Dipl.–Ing. Timo Rehfeld, who supervised me on the side of MBRDNA, and Dipl.–Inform. Michael Weber, who supervised me on the side of FZI Research Center for Information Technology. They supported me every time, whenever I ran into a problem or had a question about my research or writing. They consistently allowed this thesis to be my own work, but steered me in the right direction whenever they thought I needed it.

I would also thank Prof. Dr.–Ing. R. Dillmann, who agreed to be my second supervisor at Karlsruhe Institute of Technology (KIT).

In addition I would like to thank MBRDNA and my team "Autonomous Driving - Environment Perception", who gave me the opportunity to make research in California within an enjoyable atmosphere and a great infrastructure to train and execute machine learning approaches.

Furthermore, I would like to acknowledge Christian Olms, who supported me to hand in my thesis, while being located on another continent.

Finally, I must express my very profound gratitude to my parents for providing me with unfailing support and continuous encouragement throughout my years of study and through the process of researching. This accomplishment would not have been possible without them.

Thank you.

Sunnyvale, *Florian Pierre Joseph Piewak*
June 2016

# Contents

# 1. Introduction

Every day people are injured by car accidents. Some of these injuries result in death so that 1.25 million road traffic fatalities happened globally in 2013 [55, p. 2], what can be seen in Figure 1.1. There the increase of traffic deaths has to be observed related to the increase of 16% of registered vehicles worldwide between 2010 and 2013. As a result the number of relative traffic fatalities decreased, but is still high and has to be reduced absolutely with new approaches for vehicle safety. The Swedish Parliament for instance introduced "Vision Zero" in 1997 [48], with the suggestion that "no one will be killed or seriously injured within the road transport system" [48]. To fulfill this vision, it is important to conduct research in the field of driver assistance systems and autonomous driving. Especially because most of the accidents are caused by human errors [53], vehicles should be able to recognize and react on dangerous situations or drive autonomously.

All the research in the field of driver assistance systems and autonomous driving shows that environment perception is an elementary part for these autonomous or driver assistance systems. Not only to localize the vehicle in a global map, also to detect and react on dynamic obstacles in the surrounding of the vehicle such as pedestrians or cars, a stable environment perception should be achieved. To recognize them, a variety of sensors such as cameras, radars and LiDARs are used. The information of the different sensors is fused together to obtain a representation of the entire surrounding. This representation is commonly a grid mapping approach [16, 46]. Based on this grid map moving objects and non-moving objects, which are called dynamic and static, can be distinguished with different approaches [36, 50, 51]. This information is used to track the object over time [56] and calculate reactions.

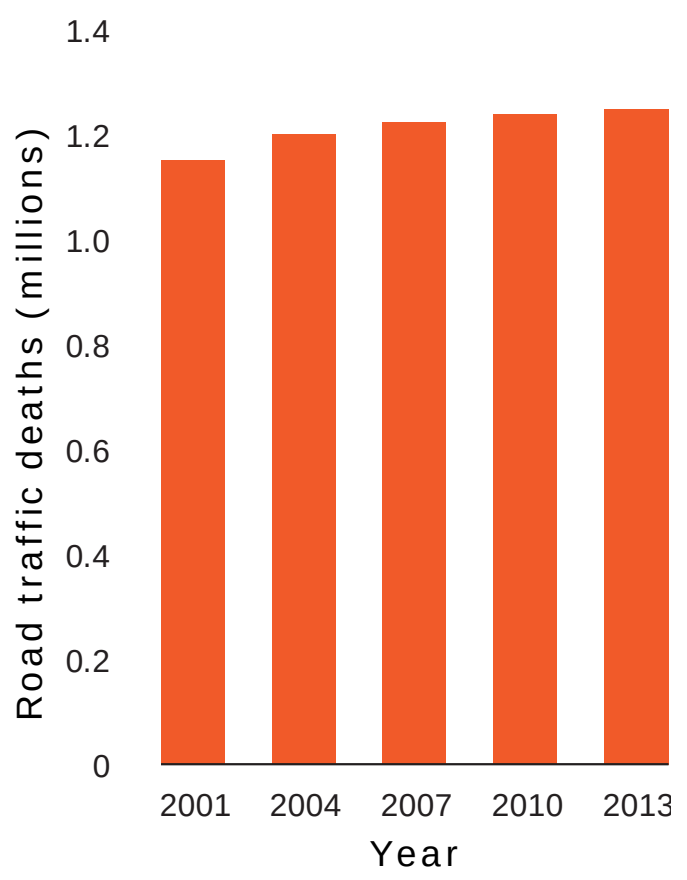

Figure 1.1.: Number of worldwide road traffic fatalities [55, p. 2].

Tracking in general works well, but has its limits, especially if the input data can be improved. For this purpose, Convolutional Neural Networks (CNNs) are used in this thesis to detect moving obstacles within the grid map. The advantage of CNN approaches is that with sufficient data the solution of a specified problem would be calculated implicit. The general obstruction is to generate and label the data. Especially for deep learning approaches like CNNs large amounts of training samples are required to generate good results. Typically the data is generated by human annotators, which is a time and cost expensive task. Furthermore, the annotators have to be skilled to recognize and label the data in a correct way. To reduce this expensive task semi supervised learning algorithms were introduced in 1965 [40] and became more important since the 1990s [32, 33, 35]. By using this approach, the machine learning algorithms could handle expensive labeled data as well as cheap unlabeled data. Especially in the current decade, unlabeled data became easy to generate in a large scale regardless of which area of life. So it is not a problem to get pictures from Google, videos from YouTube or other unlabeled data for machine learning approaches.

Several goals should be achieved in this thesis. First of all the CNN should improve current implementations of moving object detections. For this purpose, pre-trained CNNs are used and optimized. On the one hand the accuracy has to be optimized and on the other hand the processing time has to be observed. The training data for the fine tuning of the CNN is generated in a semi-automatic manner. A second goal is the evaluation if orientation extraction of moving objects with the CNN can improve the current orientation extraction. The last goal is to automatically generate new training data from unlabeled data and to evaluate the improvement of the CNN by restarting trainings with a larger dataset.

The structure of this thesis correspond to these different objectives. After a background section (Chapter 2), the methodology of each of these three goals is presented (Chapter 3). Afterwards, the realization with the semi-automatic data generation is described (Chapter 4). Then the results of the three goals are presented (Chapter 5). Finally the thesis is summarized (Chapter 6) and an outlook is given (Chapter 7).

# 2. Background

In this chapter, the background of the thesis and the whole system is described. First, a brief overview over the development of autonomous driving is given. Then a description of the environment perception with grid maps is provided. Afterwards the CNNs and the semi supervised learning are introduced and finally an angle representation for the orientation extraction and receiver operator characteristic are discussed.

## 2.1. Trend of Autonomous Driving

Already in the 1990s the European PROMETHEUS project was one of the first autonomous driving projects with an autonomous drive from Munich in Germany to Odense in Denmark [15, 20, 14]. In this project several vehicles like a Mercedes 500 SEL and a 10-ton bus (see Figure 2.1) of Daimler-Benz AG were equipped with different sensors. The results were autonomous driving systems for different situations like lane following or lane changing on the freeway entrance without traffic. Other results were detection and tracking of up to five Obstacles with CCD-cameras.

After that, several projects and challenges were created to promote research in this field. For example the Defense Advanced Research Projects Agency (DARPA) started a challenge in off-road scenarios in 2004 [47]. The goal of the challenge was to develop an autonomous vehicle capable of traversing off-road terrain. The autonomous vehicles had to navigate 142 miles through the Mojave Desert in no more than ten hours. After repetition of the challenge in 2005, five of 23 teams finished the course. The winner used different sensors like radars, lasers, cameras and inertial measurement units (IMU). These measurements were transferred to a grid map for each sensor type (see Figure 2.2), which were combined to a fused grid map. Based on this map the drivable path was generated.

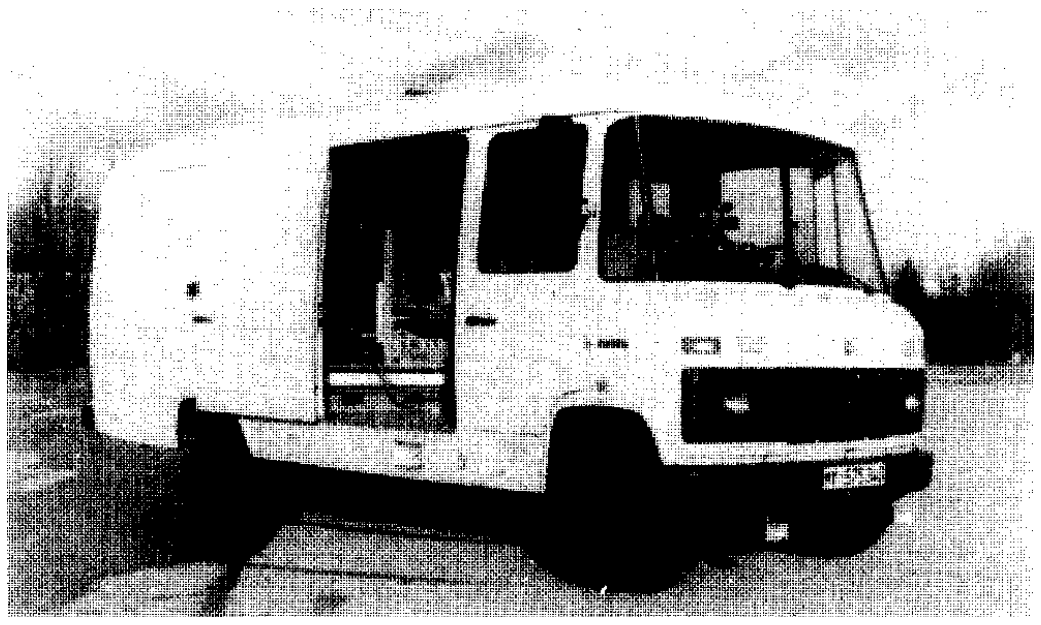

Figure 2.1.: Testing vehicle within PROMETHEUS project 1990 [15, p. 1275].

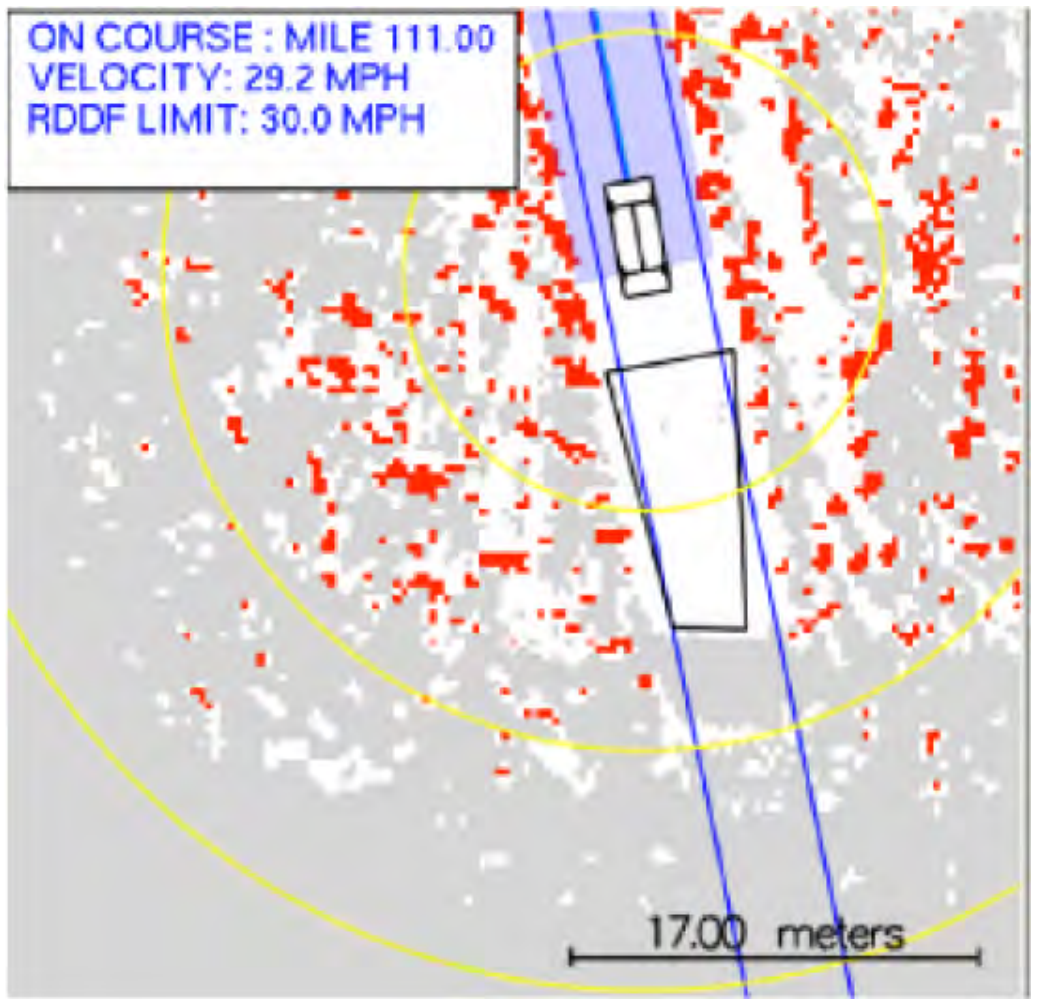

Figure 2.2.: Laser grid map of STANLEY, the winner of DARPA Challenge 2005 [47, p. 674].

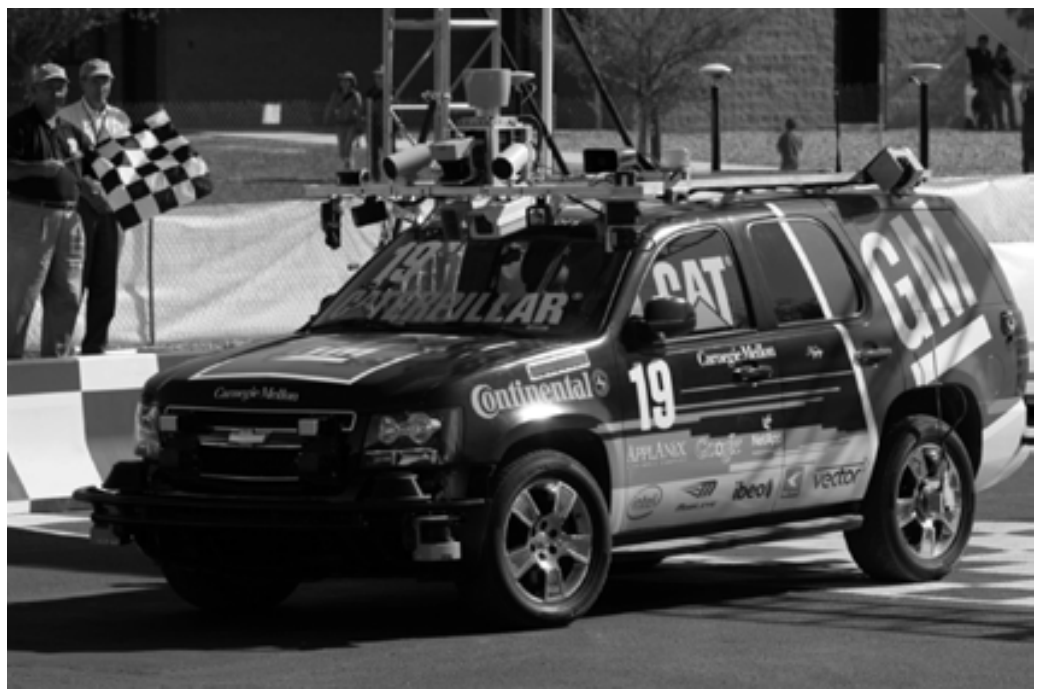

Figure 2.3.: Winner of the DARPA Urban Challenge 2007 with several sensors on top of the roof [49, p. 19].

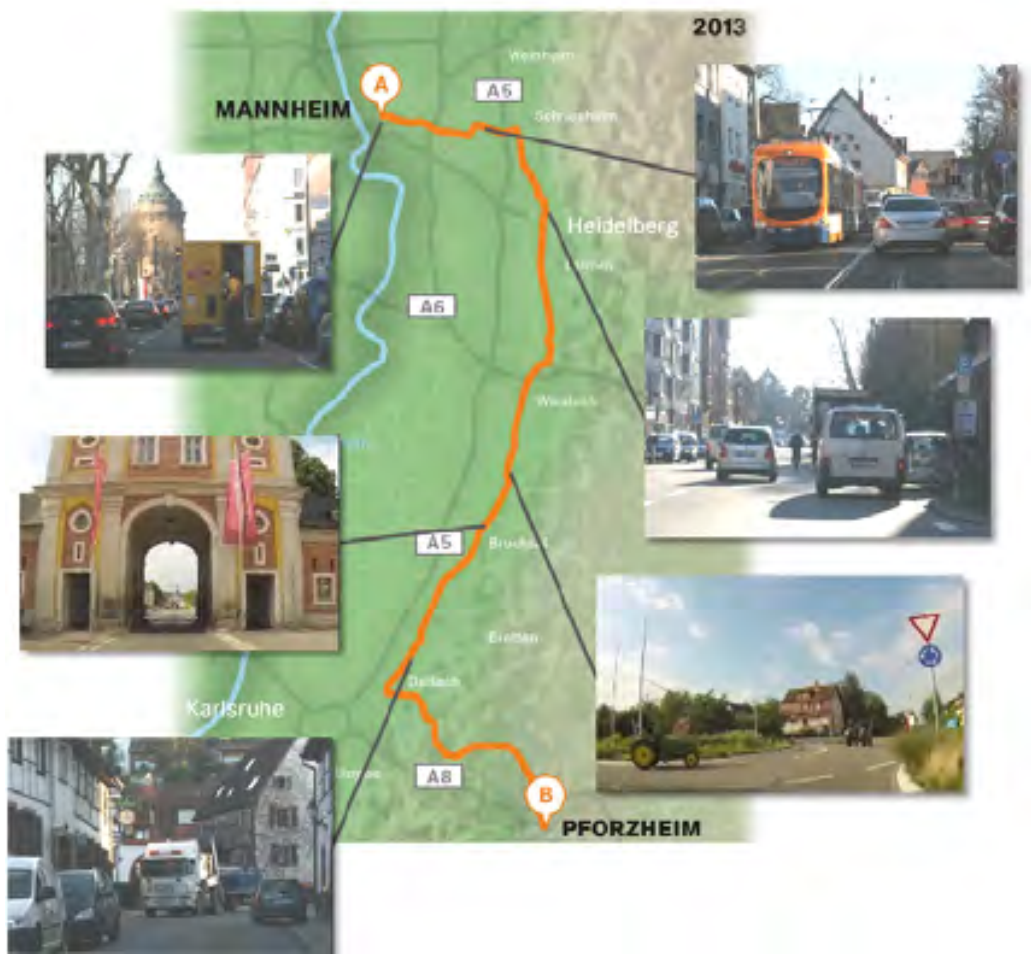

Figure 2.4.: Path of Bertha, an autonomous S-Class with close-to-market sensors in 2013 [59, p. 2].

To extend the scenario, the DARPA Urban Challenge was launched in 2007 [49]. There, autonomous vehicles had to drive a 30 miles race through urban traffic in and around the former George Air Force Base at speeds up to 30 miles per hour (mph). This challenge was especially created to build autonomous systems that could react to dynamic environments. The tasks are among other things, obeying traffic rules while driving along roads, merging with moving traffic, parking and independently handling unusual situations. Six of eleven teams reached the end of the course in 2007 and showed the state-of-the-art of autonomous driving development (see Figure 2.3).

Also beside such challenges, car manufacturers like Daimler presented their expertise: Bertha, an autonomous S-Class, was released in 2013 and drove the Bertha Benz Memorial route [59] through urban streets and rural roads, including traffic lights, roundabouts and pedestrian crossings (see Figure 2.4). The specialty of this vehicle was the sensor configuration. In contrast to the DARPA challenges, where each car had specific sensor racks on top of the roof, Bertha was only equipped with close-to-market sensors like stereo cameras, mono cameras and radars (short and long-range). Finally, Bertha drove 64 miles without an intervention of a human.

All these autonomous vehicles were designed for specific situations and scenarios. To integrate these approaches in mass-production vehicles, the algorithms have to be generalized, which is the current research of several companies within the automotive or software industry. They are testing these algorithms for example on roads in California, what is regularly reported by the California Department of Motor Vehicles (DMV) [44].

## 2.2. Representation of the Environment

Recognizing the environment is an essential part of autonomous driving. Based on this information object detection, tracking, and path planning algorithms can be applied. A well-established environment perception approach is grid mapping, which was first introduced in the field of robotics [16, 46]. Grid maps divide the environment into grid cells and estimate their properties like the occupancy state. The first approaches generated grid maps only for stationary environments and are called static occupancy grid map (SOG). Later approaches detected or tracked moving grid cells and inserted this information as features into the grid map [11, 37]. These approaches are called dynamic occupancy grid map (DOG). In the next subsections, the SOG and DOG formulation by Nuss et al. [37] is introduced briefly, as it is the foundation of the work presented in this thesis.

### 2.2.1. Static Occupancy Grid Map

The goal of the static occupancy grid map (SOG) is to find occupied and free areas in the environment. This can be provided with a probability $P(O_k^c)$ for the occupancy states $o_k^c \in \{O, F\}$ at time $k$ for each grid cell $c$ in a static environment. The probability can be updated with independent measurements $z$ by means of a Bayesian filter [46, p. 23]. With an initial estimation probability $P(O_k^c) = 0.5$ the update equation becomes:

$$P(O_{k+1}^c) = \frac{P(O_k^c)P(O_{k+1}^c|z_{k+1})}{P(O_k^c)P(O_{k+1}^c|z_{k+1}) + P(F_k^c)P(F_{k+1}^c|z_{k+1})} \quad . \qquad [2.1]$$

Here $P(F_k^c)$ represents the probability for the free space, which is the counter hypothesis:

$$P(F_k^c) = 1 - P(O_k^c) \quad . \qquad [2.2]$$

$P(O_{k+1}^c|z_{k+1})$ and $P(F_{k+1}^c|z_{k+1})$ describe the inverse sensor model, which is the likelihood of the state, given the current measurement $z_{k+1}$. For further details, the reader is referred to [46] and [37].

The problem of the SOG is the movement of objects within the grid map. Especially to detect and react on these objects in traffic scenarios, they have to be recognized.

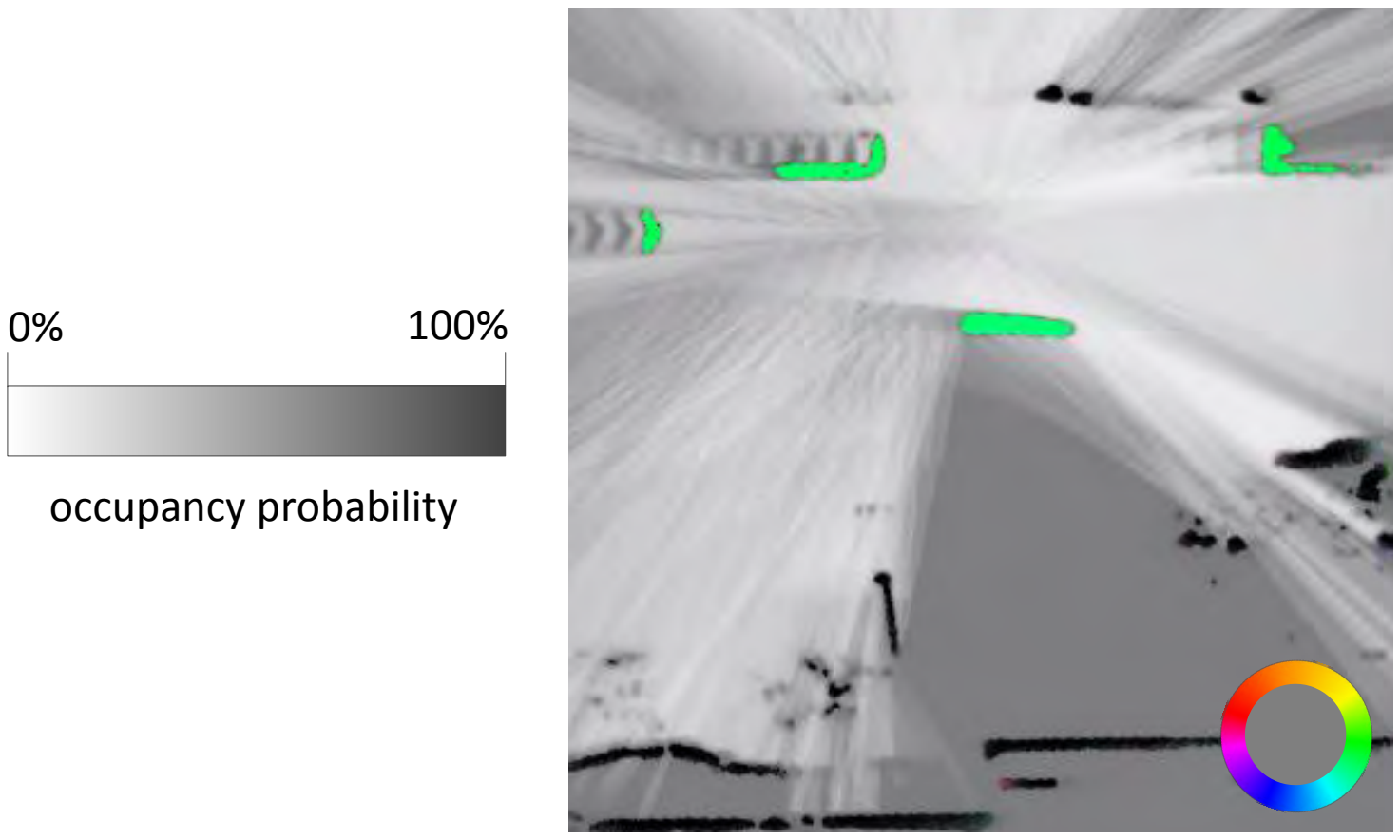

Figure 2.5.: Example of a dynamic occupancy grid map (DOG): the orientation of the velocity is encoded in a color (see the circle) and the occupancy probability is encoded with gray scale values.

### 2.2.2. Dynamic Occupancy Grid Map

In this thesis, the dynamic occupancy grid map approach of Nuss et al. [37] is used. There, a particle filter predicts dynamic and occupied states for each grid cell, where the cell state $X_k^c$ at time $k$ for a grid cell $c$ is given as

$$X_k^c = (o_k^c, x_k^c) \quad . \tag{2.3}$$

The term $o_k^c$ is the occupancy state as in the SOG and $x_k^c$ is the dynamic state, which is defined as a two-dimensional position $p_x$, $p_y$ and a two dimensional velocity $v_x$, $v_y$, i.e.

$$x = (p_x, p_y, v_x, v_y)^T \quad . \tag{2.4}$$

The cell state is represented by particles of the particle filter. Each particle $i$ in a cell $c$ at time $k$ contain velocities, which corresponds to the dynamic state, and has a weight $w_k^{c,i}$. The sum of these weights over a cell correspond to the occupancy probability of the cell. An example of a DOG can be recognized in Figure 2.5.

The processing of the particle filter is divided into the prediction, update and resampling step, which are described in the following subsections. The prediction and the update step can also be recognized in Figure 2.6. For further details, the reader is referred to [37] and [36].

#### Prediction

Each particle is propagated according to a constant velocity and constant orientation model including a discrete, unbiased, white, Gaussian noise $v$. This noise produces diversity over the particles, even if there are some similar particles resampled in further steps. The model transition is defined as:

$$x_{k+1} = f(x_k, v_k) = \begin{pmatrix} 1 & 0 & T & 0 \\ 0 & 1 & 0 & T \\ 0 & 0 & 1 & 0 \\ 0 & 0 & 0 & 1 \end{pmatrix} x_k + v_k \quad . \qquad [2.5]$$

$T$ represents the time difference between the time step $k$ and $k+1$. With this propagation a set of particles can move from cell $a$ to cell $c$, what is denoted as $\Gamma^{c \leftarrow a}_{k+1|k}$. The set of weights of these particles is stated as ${}^{w}\Gamma^{c \leftarrow a}_{k+1|k}$ and a probability for the transition event $E^{c \leftarrow a}_{k+1|k}$ that the occupancy of cell $a$ moves to cell $c$ can be defined by

$$P(E^{c \leftarrow a}_{k+1|k}) = \sum_{w^{a,i}_{k+1|k} \in {}^{w}\Gamma^{c \leftarrow a}_{k+1|k}} w^{a,i}_{k+1|k} \quad . \qquad [2.6]$$

Based on this definition the a priori occupancy probability can be specified with the assumption that all transitions are statistically independent by

$$P(O^{c}_{k+1|k}) = 1 - \prod_{1 \leq a \leq C} \left(1 - P(E^{c \leftarrow a}_{k+1|k})\right) \quad , \qquad [2.7]$$

where $C$ defines the total number of grid cells.

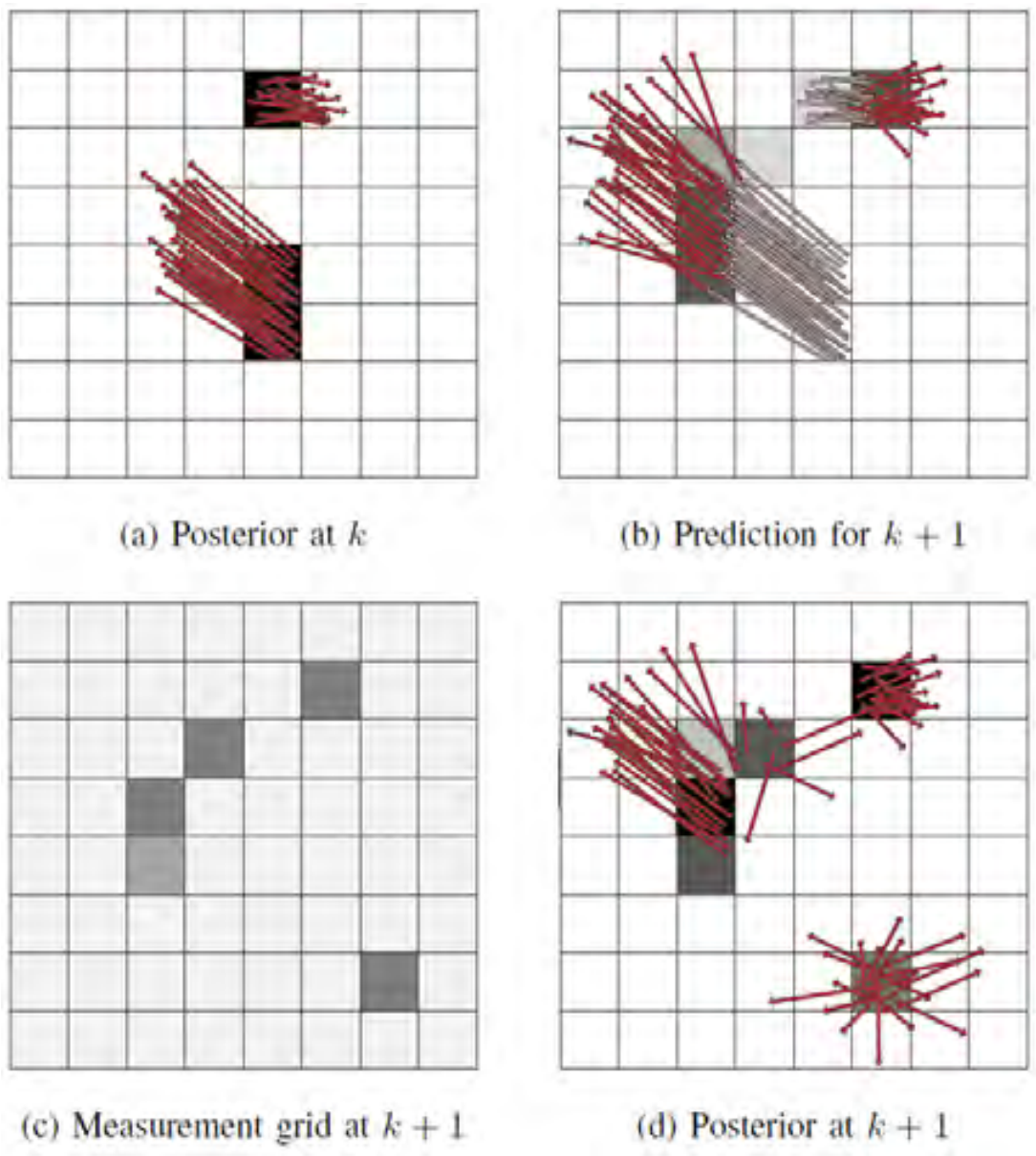

Figure 2.6.: Different steps of the particle filter within the DOG: (a) describes the first state which is used to create the prediction (b). After a measurement (c), the actual state is updated (d). The color of the grid cell describes the occupancy probability (the darker, the higher the probability is). The arrows describe the dynamic state of each particle. [36]

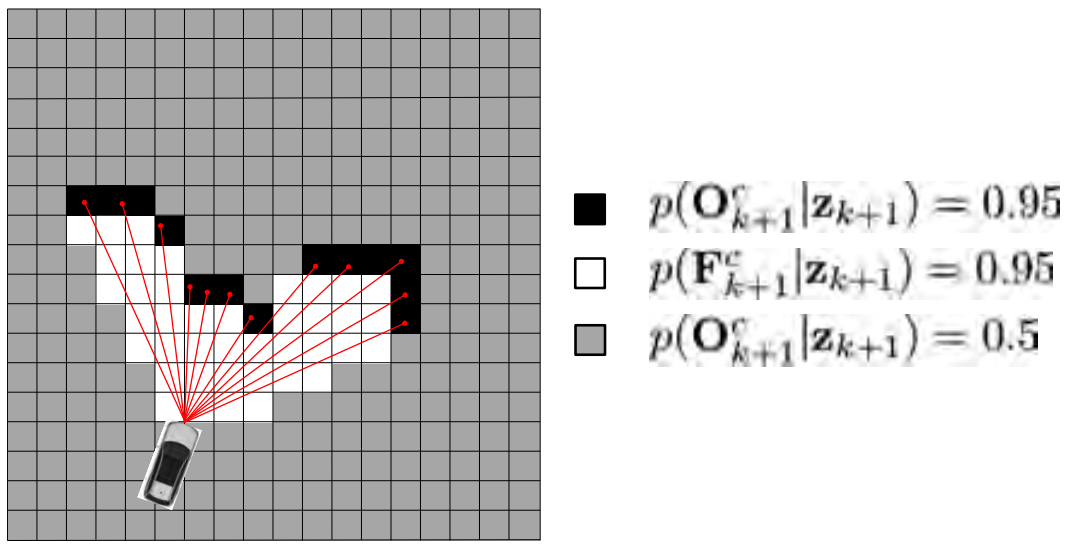

Figure 2.7.: Inverse sensor model for a laser measurement as a measurement grid [37, p. 1078].

### Update

To update the prediction of each particle, measurements of several sensors are combined. The resulting grid map is called measurement grid. To generate a measurement grid from several sensors, their individual inverse sensor model is used. An example of an inverse sensor model can be seen in Figure 2.7, where a laser measurement with multiple beams is shown. A high occupancy probability is created for grid cells in the neighborhood of sensor detections. Low occupancy probabilities are created for grid cells in between the sensor detection and the vehicle. All other grid cells are set to an occupancy probability of 0.5 to represent uncertainty about the occupancy state. Similar to the SOG approach, the measurement grid and the predicted occupancy states are then combined with a binary Bayes filter to obtain the posteriori occupancy probability:

$$P(O^c_{k+1|k+1}) = \frac{P(O^c_{k+1|k})P(O^c_{k+1}|z_{k+1})}{P(O^c_{k+1|k})P(O^c_{k+1}|z_{k+1}) + P(F^c_{k+1|k})P(F^c_{k+1}|z_{k+1})} \ . \qquad [2.8]$$

The dynamic state of a particle is updated by a measurement through a weight update of each particle. The updated unnormalized weight $\widetilde{w}^{c,i}_{k+1|k+1}$ is

$$\widetilde{w}^{c,i}_{k+1|k+1} = \beta^c_{k+1}\ w^{c,i}_{k+1|k+1} = w^{c,i}_{k+1|k}\ p(z_{k+1}|x^{c,i}_{k+1|k}) \ . \qquad [2.9]$$

Unnormalized means that the sum of each particle weight $\widetilde{w}^{c,i}_{k+1|k+1}$ within a grid cell does not represent the occupancy probability. To get a normalized updated weight $w^{c,i}_{k+1|k+1}$, the normalization factor $\beta^c_{k+1}$ can be calculated with

$$\beta^c_{k+1} = \frac{1}{P(O^c_{k+1|k+1})} \sum_{\widetilde{w}^{c,i}_{k+1|k+1} \in {}^{\widetilde{w}}\Gamma^{c\leftarrow}_{k+1|k}} \widetilde{w}^{c,i}_{k+1|k+1} \ , \qquad [2.10]$$

where ${}^{\widetilde{w}}\Gamma^{c\leftarrow}_{k+1|k}$ describes the set of unnormalized weights of the particles in ${}^{w}\Gamma^{c\leftarrow}_{k+1|k}$.

### Resampling

Resampling is a main part of classical particle filters. It is used to eliminate particles and reproduce others based on their weights. This includes the reproduction of one particle multiple times. As a result, degeneracy is avoided and convergence to a stable velocity is evoked. At the same time new particles are inserted. The weight of new spawn particles is dependent of the ratio of measured and predicted occupancy probability and the weights of survived particles. This can be defined as follow

$$\frac{\sum w_{new}}{\sum w_{survived}} = \gamma \frac{P(O^c_{k+1|k+1})}{P(O^c_{k+1|k})} \quad . \qquad [2.11]$$

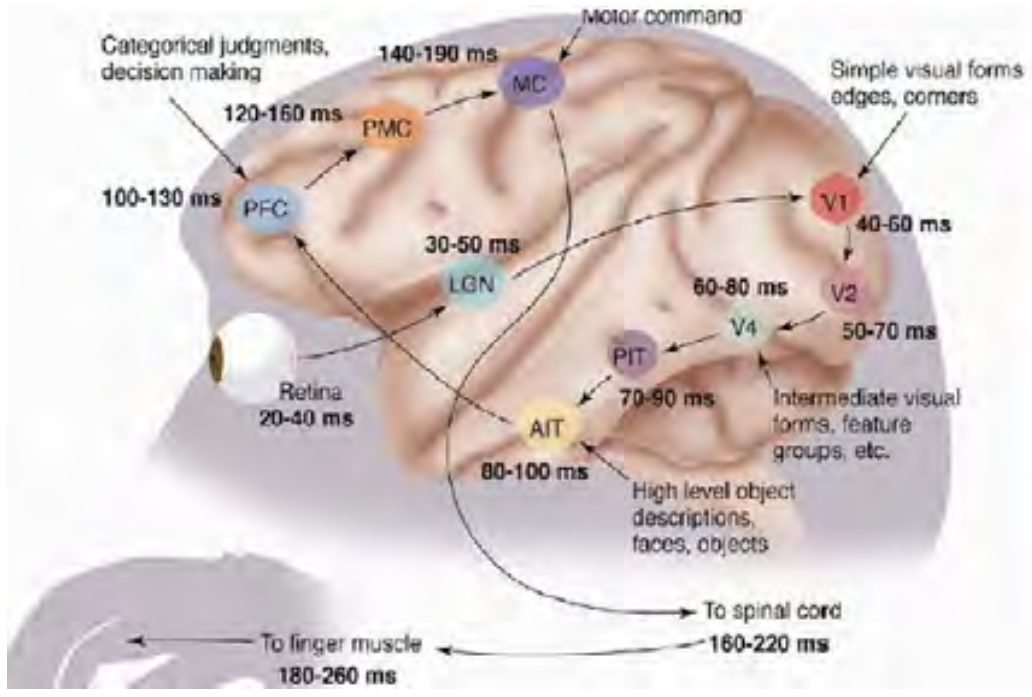

Figure 2.8.: Image processing and reaction in a monkey brain [45].

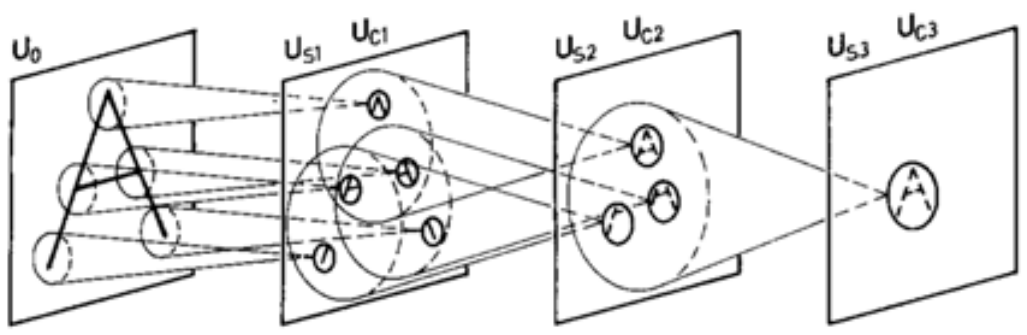

Figure 2.9.: One of the first technical realization of visual recognition of mammal brains [21, p. 198].

Especially if a high occupancy is measured but no particles were predicted into the corresponding cell, new particles are produced.

## 2.3. Convolutional Neural Networks

In this section, the Convolutional Neural Networks (CNNs) are described in context of this thesis. After a brief motivation different layer types are discussed. Afterwards, the development of Fully Convolutional Neural Networks (FCNs) followed by basic training algorithms is provided. Then possible integrations into the current environment perception system are introduced. For further details, the reader is referred to [23].

### 2.3.1. Motivation

Since the invention of computers there is an effort to create intelligent machines that are able to solve problems adaptively without human interaction. The inspiration for the realization of these ideas mostly came from nature. CNNs are motivated by the visual recognition part of the brain of mammals. As shown in Figure 2.8, the brain is structured in different sections to recognize objects and create a body reaction. Each section has a specific task to solve. Especially in the first levels of recognition (Retina - LGN - V1 - V2 - V4 - PIT - AIT) different levels of object detection can be found. The first layers are used to detect edges, corners and simple shapes. Based on these detections more complex shapes can be recognized. Only at the last layers (AIT) real object representations are produced. Afterwards, this representation is used to generate decisions and reactions executed by actors like fingers.

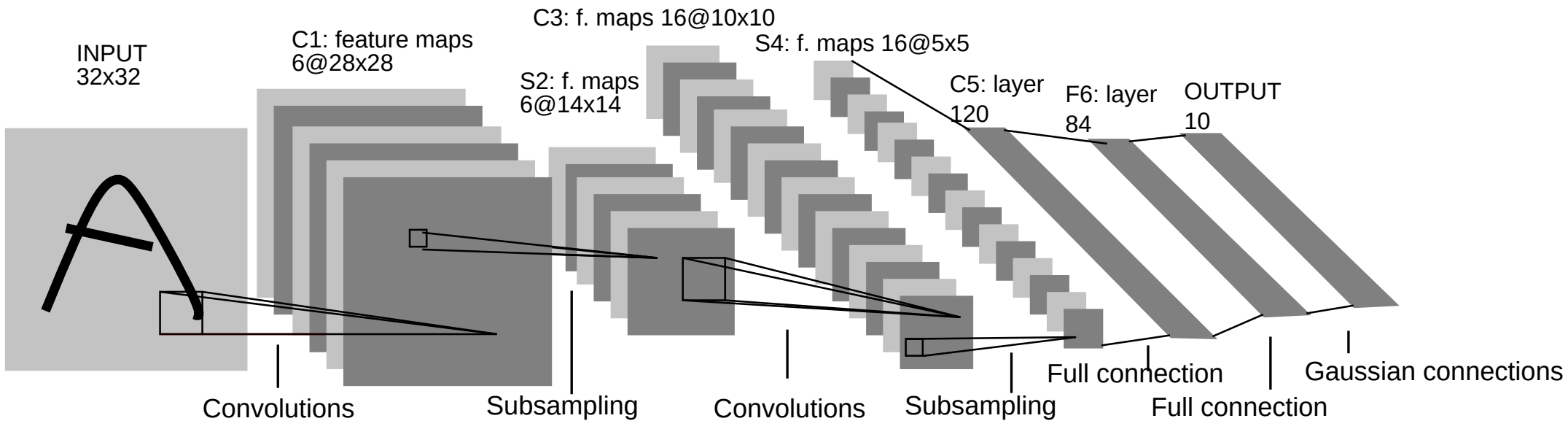

Figure 2.10.: One of the first Convolutional Neural Networks trained in a supervised manner to recognize characters [27, p. 7].

This layered detection and recognition structure is used for several years in the computer vision field. Two-dimensional filters are used to detect simple structures like corners and edges. Based on this information a higher level of object detection is calculated [43]. The problem of these approaches is that the filter types have to be chosen and optimized by hand. In contrast, CNNs allow to learn and optimize the filters automatically. Based on first technical realization of the brain visual recognition from Fukushima [21] (see Figure 2.9), LeCun et al. [27] created one of the first CNNs that was trained in a supervised manner. Their CNN was trained to recognize handwritten characters, which can be seen in Figure 2.10. Based on this paper several deeper CNN structures were created to solve classification tasks based on two-dimensional input images like AlexNet [26], VGG-net [41] or GoogLeNet [42].

### 2.3.2. Layer

The CNNs consist of three-dimensional feature maps which can be connected with different layer types. The most important layers (convolutional layer, pooling layer and deconvolutional layer) are discussed in the following subsections.

#### Convolutional layer

The convolutional layer includes the filters $f(i,j)$, which are also used in several non-machine learning computer vision techniques. There a filter is a combination of weights, what represent a specific feature extractor. The filters of size $n_1 \times n_2$, which are usually squared, are used to convolve the input feature map $g(i,j)$. The input feature map at the first layer can be a two-dimensional grayscale image or an RGB-image, where the color channel represents the third dimension. The convolution equation is given by

$$h(i,j) = f * g = \int_k \int_l f(i-k, j-l) g(k,l) \quad . \qquad [2.12]$$

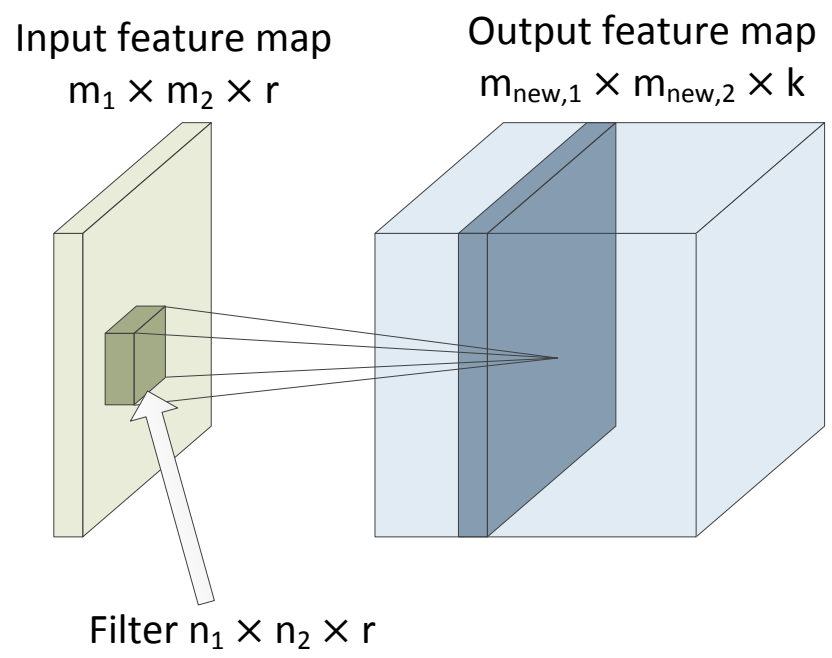

Figure 2.11.: Principle of a convolutional layer.

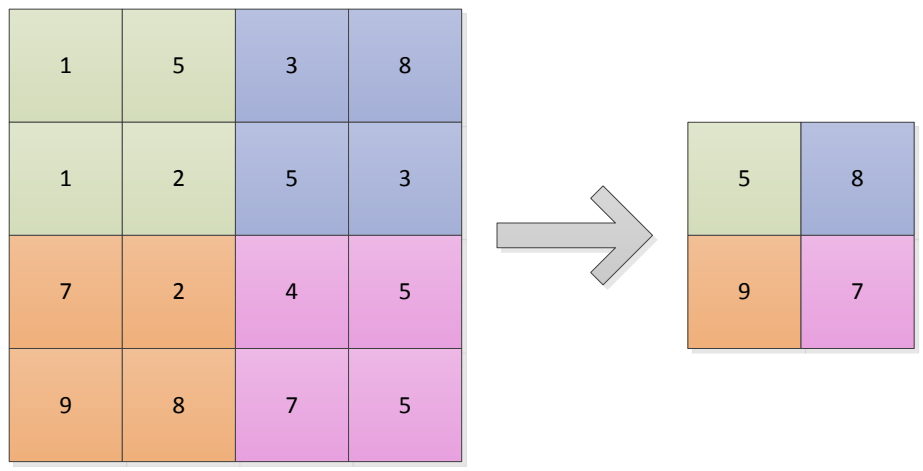

Figure 2.12.: Example of a max pooling layer with stride two.

This equation becomes a sum by the discretization of the feature map in pixels:

$$h(i,j) = f * g = \sum_k \sum_l f(i-k, j-l) g(k,l) \quad . \qquad [2.13]$$

This convolution creates a two-dimensional output, what can be seen in Figure 2.11. By using $k$ filters per convolutional layer the resulting feature map becomes three-dimensional with the size $m_{new} \times m_{new} \times k$. The edge of the resulting feature map $m_{new}$ can be calculated via

$$m_{new} = \frac{m - n + 2p}{s} + 1 \quad , \qquad [2.14]$$

where $s$ describes the stride, i.e. the step size to use for the iteration over the feature map. A larger stride produces a smaller output map. $p$ describes the padding, which can be used to resize the feature map to a specific size. The most common padding approach is the zero padding. The padding is usually used to rebuild the same output size as the input size.

### Pooling Layer

With the filters within the convolutional layers several features are extracted. But these features are related to a specific size of the input feature map. To obtain features of different map sizes, the feature map can be reduced by using pooling layers. They fuse squares of pixels with different possible approaches like mean or maximum, what can be seen in Figure 2.12.

### Deconvolutional Layer

With the discussed layers the size of the feature maps becomes smaller in deeper layers of the network. Consequently, this process has to be reversed to obtain a pixelwise prediction as output. For this purpose Long et al. [28] used the deconvolutional layer, which was already introduced by Zeiler et al. [57]. This layer uses bilinear upscaling to initialize the deconvolutional weights. They produce a larger output map, what you can see in Figure 2.13. An upscaled pixel value can be calculated based on the four surrounding pixel values with

$$R_0 = \frac{e_x}{d_x + e_x} Q_{00} + \frac{d_x}{d_x + e_x} Q_{10} \qquad [2.15]$$

$$R_1 = \frac{e_x}{d_x + e_x} Q_{01} + \frac{d_x}{d_x + e_x} Q_{11} \qquad [2.16]$$

$$P = \frac{e_y}{d_y + e_y} R_0 + \frac{d_y}{d_y + e_y} R_1 \qquad [2.17]$$

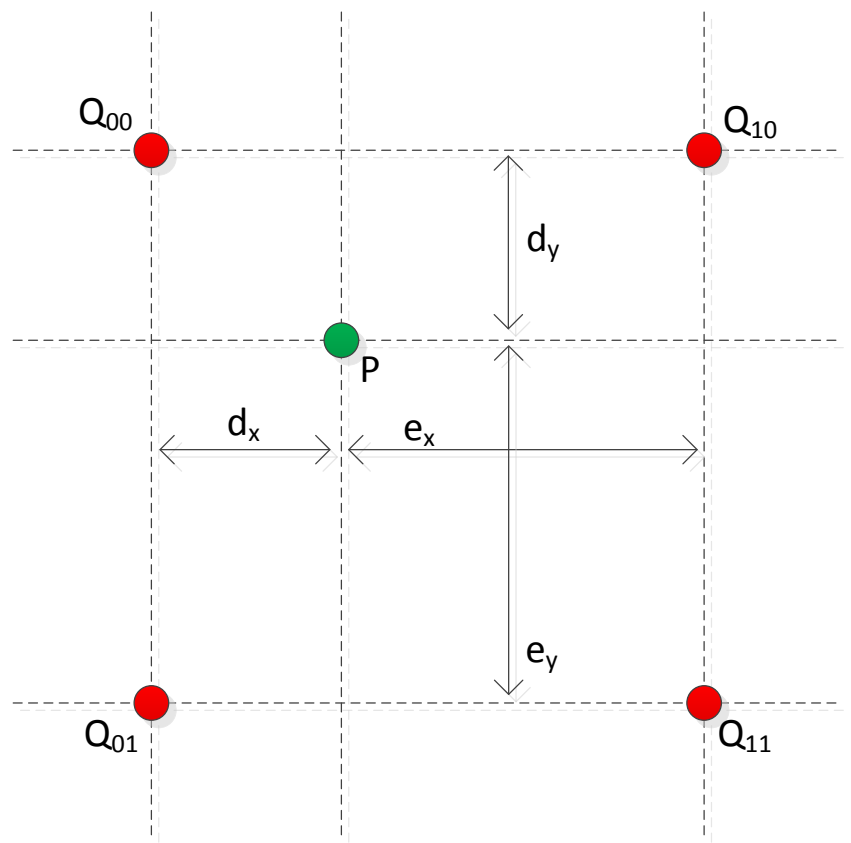

Figure 2.13.: Bilinear upscaling, the red points are the original points and the green point is an example for an interpolated target pixel.

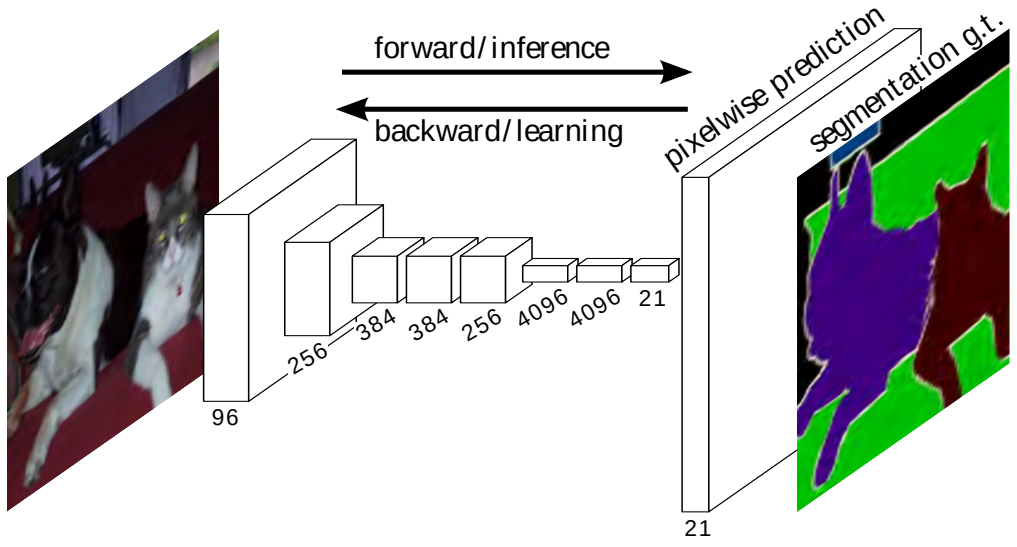

Figure 2.14.: One of the first Fully Convolutional Neural Networks to create a pixelwise predictions [28, p. 3431].

### 2.3.3. Fully Convolutional Network

All the networks described in Section 2.3.1 classify the entire image, because of the whole structure with fully connected layers in the end. For semantic segmentation a sliding window approach could be used to create a pixelwise prediction. This is in general not suitable caused by the computational expense what arises by executing the CNN once for each pixel. For this purpose, the basic idea of Fully Convolutional Nerual Networks (FCNs) were introduced by Matan et al. [30] for the one dimensional output case and by Wolf and Platt [54] for the two dimensional output case.

The basic idea of a fully convolutional network structure is the size independence of the input image. This can only be proceed, if the fully connected layers of conventional CNNs are converted to convolutional layers with a specific filter size. The fully connected layers can be interpreted as convolutional layer with a filter size of the last feature map. For this reason, the two types of layer can easily be transformed by just copying the weights. The final output of a FCN is a feature map, which can produce different output feature map sizes in dependence of the input size.

Caused by the convolutions of different layers, a pixelwise prediction can not be established in the first versions of FCNs. For this reason, Long et al. [28] introduced later a fully convolutional network structure with deconvolutional layers, which was trained end-to-end for semantic segmentation. This combination of the layers facilitates a two-dimensional output of the neural network for each class, where the output has the same size as the input (see Figure 2.14). As a result, a pixelwise classification is produced by executing the CNN only once.

### 2.3.4. Training

The training of CNNs is responsible for tuning and optimizing the filters of convolutional and deconvolutional layers, where each filter consists of adaptable weights. This is done by using a gradient descent algorithm. The applied algorithm is called backpropagation. There, an overall error is calculated, which is called loss. This loss is used to adapt the weights for example of the convolutional and deconvolutional layers by propagating back the error through the different layers. As an example the multinomial logistic loss is described.

The multinomial logistic loss is used to tune and optimize the parameters of the CNN with a multi class output $y^i \in \{1,\ldots,K\}$ for $K$ classes. For this purpose a training set $\{\left(x^{(1)},y^{(1)}\right),\ldots,\left(x^{(m)},y^{(m)}\right)\}$ with $m$ samples is given, which contains the input $x$ and the desired output $y$. The hypothesis $h_\theta$ describes the class probability and is defined with

$$h_\theta(x) = \begin{bmatrix} P(y=1|x;\theta) \\ P(y=2|x;\theta) \\ \vdots \\ P(y=K|x;\theta) \end{bmatrix} = \frac{1}{\sum_{j=1}^{K} \exp\left(\theta^{(j)\top}x\right)} \begin{bmatrix} \exp\left(\theta^{(1)\top}x\right) \\ \exp\left(\theta^{(2)\top}x\right) \\ \vdots \\ \exp\left(\theta^{(K)\top}x\right) \end{bmatrix} , \quad [2.18]$$

where $\theta$ represents the parameters of the model, what is in this thesis the weights of the CNN. As a result the cost function $J(\theta)$ is defined as

$$J(\theta) = -\left[\sum_{i=1}^{m}\sum_{k=1}^{K} 1\{y^{(i)}=k\}\log P\left(y^{(i)}=k|x^{(i)};\theta\right)\right] , \quad [2.19]$$

where $1\{.\}$ is the “indicator function”, so that

$$1\{a\ true\ statement\} = 1 \quad \text{and} \quad [2.20]$$

$$1\{a\ false\ statement\} = 0\ . \quad [2.21]$$

The last part of the equation $P\left(y^{(i)}=k|x^{(i)};\theta\right)$ represents the probability of a specific class, which can be extracted from Equation 2.18:

$$P\left(y^{(i)}=k|x^{(i)};\theta\right)=\frac{\exp\left(\theta^{(k)\top}x^{(i)}\right)}{\sum\limits_{j=1}^{K}\exp\left(\theta^{(j)\top}x^{(i)}\right)} \qquad [2.22]$$

The cost function has to be minimized to generate an optimal output. For this reason the multinomial loss is propagated back through all layers, to adapt the CNN as a model. For further details, the reader is referred to [34].

In general a training of CNNs need a large amount of labeled data, especially if the network is trained from scratch. For this reason pre-trained networks are used to initialize the CNNs. Pre-trained networks are CNNs, which are trained and optimized on public datasets like Pascal VOC [18] or Imagenet [13]. These public datasets contains different images for classification or segmentation approaches. After initializing the weights of the CNN by using pre-trained CNNs, the fine tuning can be executed, to adapt the CNN to the own dataset.

### 2.3.5. Integration of Convolutional Neural Networks

One of the most important parts of environment perception is detection of other obstacles in the surrounding of the vehicle. To achieve that, several sensors like radars, LiDARs and cameras are installed in autonomous vehicles. The produced sensor data is fused to a general representation of the surrounding of the vehicle as described in Section 2.2.2. Based on the DOG, clusters of occupied grid cells are extracted and marked as dynamic or static [39]. These clusters are then used to instantiate or update object-level tracks which incorporate higher level object knowledge such as vehicle dynamics [56], that are not taken into account in the DOG.

To distinguish between dynamic and static clusters only the velocities of particles per grid cell are taken into account [36]. This can cause some problem with noise or incorrect measurements. To improve the whole toolchain and generate more accurate clusters this thesis proposes a deep learning approach with convolutional neural networks (CNNs) to create a pixelwise classification of dynamic and static parts of the grid map. Possible integrations in the toolchain can be seen in Figure 2.15 and Figure 2.16. In Figure 2.15 the classification of the CNN is used in parallel to the toolchain to improve the current clustering algorithm. With this approach the clustering algorithm can generate clusters as before and uses the CNN simply as an improvement. Another approach is to integrate the CNN completely into the toolchain like in Figure 2.16. As a result, the cluster algorithm do not need the whole grid map as an input what can end in a reduction of processing time.

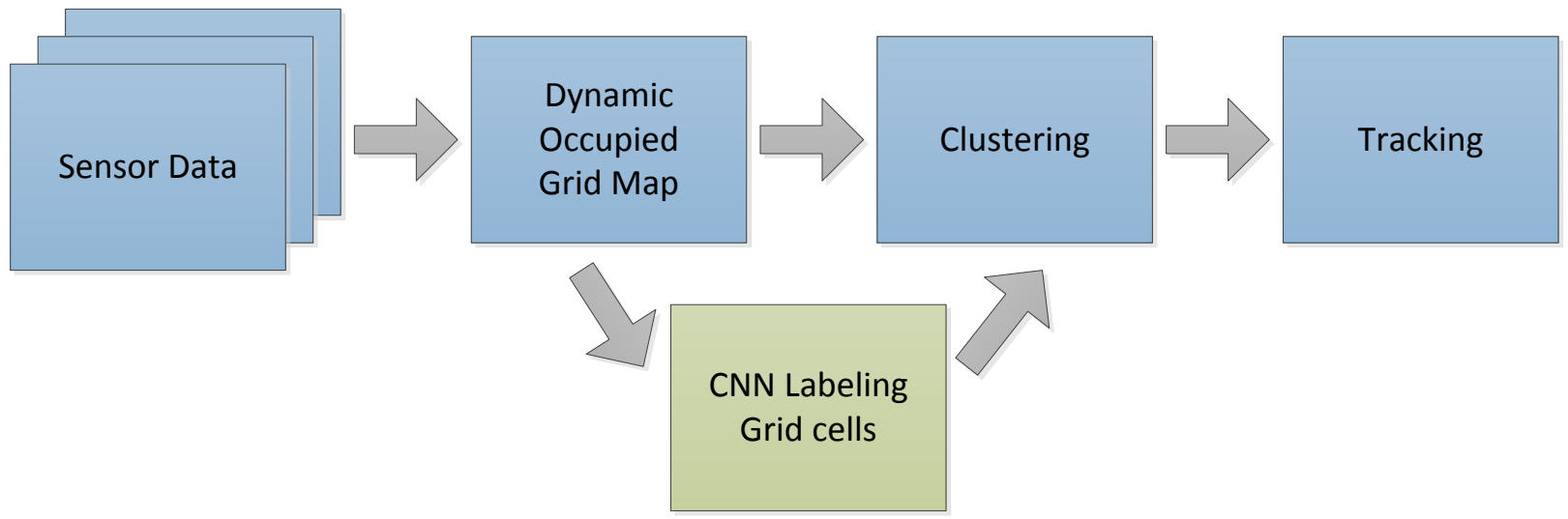

Figure 2.15.: Parallel integration of Convolutional Neural Networks (CNNs).

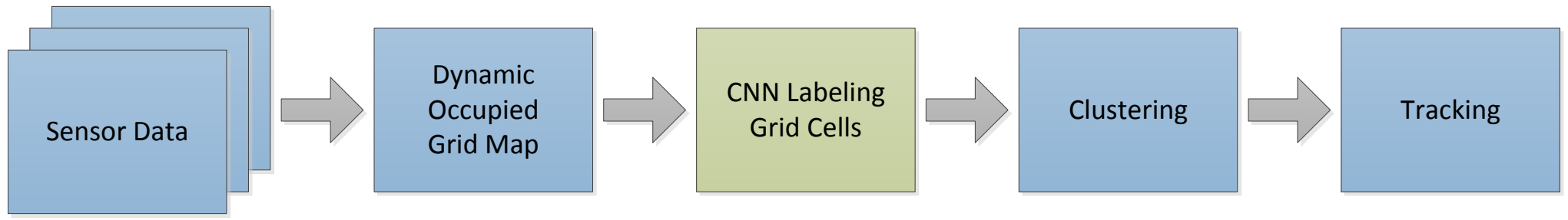

Figure 2.16.: Full integration of Convolutional Neural Networks (CNNs).

## 2.4. Semi Supervised Learning

In contrast to static modeling approaches, which solves problems by using manual defined equations as a model, machine learning approaches can potentially determine better results by using adaptive models, especially when the static model is not trivial to find. For this purpose machine learning approaches need representative data to extract meaningful representations. In general two possible learning approaches exist:

- supervised learning
- unsupervised learning

Supervised learning uses labeled data, i.e. input data and the corresponding desired output data. Based on this, the machine learning approach regresses a representation to map the input to the desired output. Unsupervised learning uses unlabeled data, i.e input data without any specific target output. Most of the unsupervised learning approaches recognize clusters or detect structures within the data. This type of learning is usually used in the field of data mining. These two presented learning approaches are combined in different ways to more complex approaches like active learning or semi supervised learning. For further details, the reader is referred to [4].

A major problem with supervised machine learning approaches is the generation of labeled data. Especially if the problem is slightly different to problems in the literature, public datasets like Pascal VOC [18], Imagenet [13] or Citycapes [10] cannot be used. An own dataset has to be created, what ends in a manual annotation. For this annotation the annotators have to be skilled to recognize and label the data in a correct way, what is a cost and time expensive task. To reduce this problem, semi supervised learning algorithms were introduced in 1965 [40] and became more important since the 1990s [32, 33, 35]. There, a combination of labeled and unlabeled data is used to train the model. The goal of these semi supervised algorithms is that only a few data is labeled and together with a large number of unlabeled data the machine learning approach becomes more accurate. This results in a less cost and time expensive task, especially because unlabeled data is nowadays available in large amounts.

## 2.5. Orientation Extraction

Orientation angle extraction from images is usually applied in the field of head pose estimation [2, 5]. There, the orientation is often split in different bins and a classifier is trained for each separate bin. After that a linear regression over the bins can be calculated. These approaches restrict the capability of machine learning algorithms, especially if they are able to directly regress a continuous orientation range. The problem that occurs is the periodicity of the orientation angle $\phi$. It is important that a smooth transition from 359° to 0° is encoded. This periodicity causes that gradient descent approaches (like in the training step of CNNs) cannot be used. For this purpose, Beyer et al. [3] introduced the Biternion representation, which is based on quaternions. There, the orientation $q$ is represented as

$$q = (\cos\phi, \sin\phi) \quad . \qquad [2.23]$$

With this formulation the periodicity is solved. The difference is that the machine learning approach has to regress two variables (sine and cosine) in range of $-1$ and 1. These two regressions could then be recombined to an angle $\phi$ with

$$\phi = \arctan\frac{\sin\phi}{\cos\phi} \qquad [2.24]$$

## 2.6. Receiver Operator Characteristic

For evaluating the parameters of binary classification algorithms and to benchmark different approaches usually the Receiver Operator Characteristic (ROC) curve is used. This curve represents the relationship between the true positive rate (TPR) and the false positive rate (FPR) of an algorithm (see Figure 2.17). One specific configuration of the classification algorithm produces one point within the ROC space. To generate a curve, one parameter of the algorithm is iterated over his whole range. This parameter is usually a threshold of the output probability. The goal of each classification algorithm is to reach a TPR of 100% and a FPR of 0% what can be seen in Figure 2.17 as the green line. The red diagonal represents a random process.

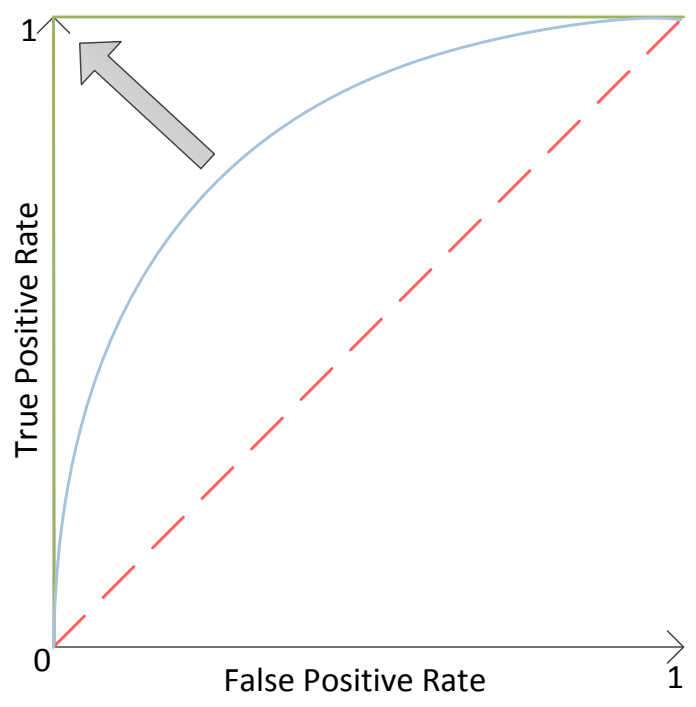

Figure 2.17.: Receiver Operator Characteristic curve: the red line represents a random process, the green curve represents a perfect classifier and the blue line represents a classification algorithm, which can be optimized.

## 2.7. Related Works

The detection of other obstacles in the surrounding of a vehicle is an elementary part of environment perception. Especially moving objects have to be detected to generate reactions for driver assistance systems or autonomous driving according to the environment. For this purpose several publications exist about the detection of moving objects with different sensor types. For example Vu et al. [50] implemented a dynamic obstacle detection by using a measurement grid of a laser sensor. They used the movements of the occupied grid cells in dependence of the free space information to detect dynamic obstacles. These dynamic objects were verified with the measurements of radar sensors. Another approach from Asvadi et al. [1] uses a Velodyne sensor to generate a 2.5 D grid map, where dynamic objects are detected by recognizing movements of grid cells over time.

The mentioned papers are using specific sensors to distinguish between dynamic and static objects. Nuss et al. [36] proposed a method to generate a dynamic object detection based on a dynamic occupancy grid map (DOG), what is described in Section 2.2.2. This allows a sensor independent recognition of moving objects. Some problems like false positives at clutter occurs by using this approach in different scenarios. For this purpose this thesis extends this approach by using the DOG as an input for a Convolutional Neural Network to detect moving objects as one of the goals.

In the last few years deep and Convolutional Neural Networks (CNNs) were in the center of research. Especially with the rising computational power deeper networks were feasible with lower training and execution time, what Cireşan et al. [7] proved with a speed improvement of factor 50. There a pixelwise labeling of images from an electron microscopy to classify membrane and non-membrane pixels was introduced with CNNs. To obtain this pixelwise prediction a sliding window approach was used, where each sub image created the label for one output pixel.

Other approaches like the Hypercolumns of Hariharan et al. [24] use a CNN for feature generation. These features are produced over all the CNN layers by upscaling smaller layers and combining all the layers pixelwise. This produces a feature vector for each pixel, which can be used to train linear classifiers. A slightly different approach is the Region-based Convolutional Network from Girshick et al. [22] which creates classified bounding boxes instead of pixelwise predictions. There, the CNN is used to create feature vectors of generated region proposals which are provided to support vector machines (SVMs). Based on these SVMs the regions are classified.

One of the problems of these approaches is that for each pixel or each region proposal the classifier has to be executed. That produces a computational overhead, which can be reduced. For this purpose Long et al. [28] introduced Fully Convolutional Neural Networks (FCNs) in combination with deconvolutional layer, which were trained end-to-end for segmentation tasks. They transfered the standard AlexNet [26], VGG-net [41], and GoogLeNet [42] to FCNs and compared them based on the Pascal VOC 2012 dataset [18]. To create a fully convolutional network for pixelwise prediction, they used deconvolution layers to rebuild the input image size as a 2D output for each class. The deconvolution layers are implemented as up-sampling layers and the output is a pixelwise class prediction. This reduces the computational overhead so that for a pixelwise prediction the net has to be executed only once. To refine the segmentation result, several deconvolution layers combined with convolutional layers were stacked. This results in more detailed segmentation structures at the output layer.

Based on this approach, several extensions and applications were developed. Fischer et al. [19] applied a similar FCN structure for calculation of optical flow. The input was a combination of two images to create the optical flow between these pictures. Another extension is the refinement with Conditional Random Fields (CRFs) [6, 58]. There, a CRF is used to produce more detailed structure of the segmented objects.

All of these approaches use predefined datasets like Pascal VOC [18], Imagenet [13] or Citycapes [10] to evaluate the neural networks. There are only a few approaches, which uses different input images. For example Wang et al. [52] used a pre-trained network, which was trained on the cifar-10 dataset [25], to detect oil tanks on satellite images. Another approach from Maturana et al. [31] used an three dimensional occupancy grid map generated by an LiDAR sensor. Based on this grid map different objects like pedestrians or toilets are classified.

In general there was no approach found, which uses two dimensional DOGs (see chapter 2.2.2) as an input for a CNN. For this reason, the datasets for training and testing has to be created and public datasets can not be used. Because of the amount of data which is needed for training, a semi-supervised learning approach is followed as one goal of the thesis. This is a current topic in the literature, but most of the approaches use weakly labeled data to generate pixelwise segmentation labels or use unsupervised pre-training. For example Dai et al. [12] use manual bounding box labels to generate a pixelwise ground truth automatically. Another approach is a combination of weakly- and semi-supervised approaches [38]. There Papandreou et al. combined pixel-level, image-level and bounding box level annotations to create more data for training. Unsupervised pre-training approaches use unlabeled data for the initialization of the CNN and train afterwards on the labeled data, like Masci et al. [29]. However all these approaches need a specific kind of labels for the main training step (image-level, bounding box, etc.).

In 1965, Scudder [40] introduced a self-learning algorithm, where an “adaptive pattern-recognition machine” was created. To obtain better results, unlabeled data was labeled by the machine itself. This new data was then used to train the machine for obtaining more accurate results. After that the labeling step starts again so that finally the amount of labeled data rises by obtaining better results with the “adaptive pattern-recognition machine”. This approach will be applied in this thesis to obtain more accurate results on the one hand and more labeled data on the other hand.

# 3. Methodology

In this chapter, the methodological background of the three goals of this thesis are described. First the segmentation of the grid map with CNNs and their parameter optimization is introduced. Afterwards the orientation extraction is presented. Finally the semi-supervised learning approach is described.

## 3.1. Segmentation

The first goal of this thesis is to create a pixelwise classification of the DOG, as shown in Figure 3.1, to support clustering of dynamic obstacles. To this end, the target set consists of two classes, dynamic (moving objects) and static (non-moving objects), where background is part of the static class. The classification process is divided into three steps. First the DOG is preprocessed to obtain a specific input for the CNN. Based on this the CNN creates the pixelwise classification. Finally, the intersection between the segmented image and an image of occupied grid cells is calculated. Doing so produces sharper labeling borders and rejects false positives which appear by labeling non-occupied grid cells as dynamic.

To achieve the classification task, Fully Convolutional Neural Networks (FCNs) are used to reduce the computationally expensive task of sliding window approaches (as mentioned in Section 2.3.3). Based on Long et al. [28], the popular network structures VGG-net [41] and Alexnet [26] are converted to FCNs with deconvolutional layers, which are using bilinear upscaling. This produces a two-dimensional output of the network and creates the desired pixelwise classification by using the same input and output size. For initialization of the CNN weights, pre-trained networks are used for reducing the training time. These pre-trained networks were trained on the public Imagenet [13] dataset, which contains colored images of different categories. Even with a different type of input data, the pre-trained networks can reduce the training time. This is caused by the pre-trained filters of the convolutional layers, which contain specific shape representations and which should improve the segmentation task.

The transfer of the pre-trained networks to a new scope requires several parameter optimizations, which are discussed in the next subsections. Each parameter is optimized by its own to finally merge the optimized parameters together to obtain an optimized FCN. Therefore an independence of parameters is assumed.

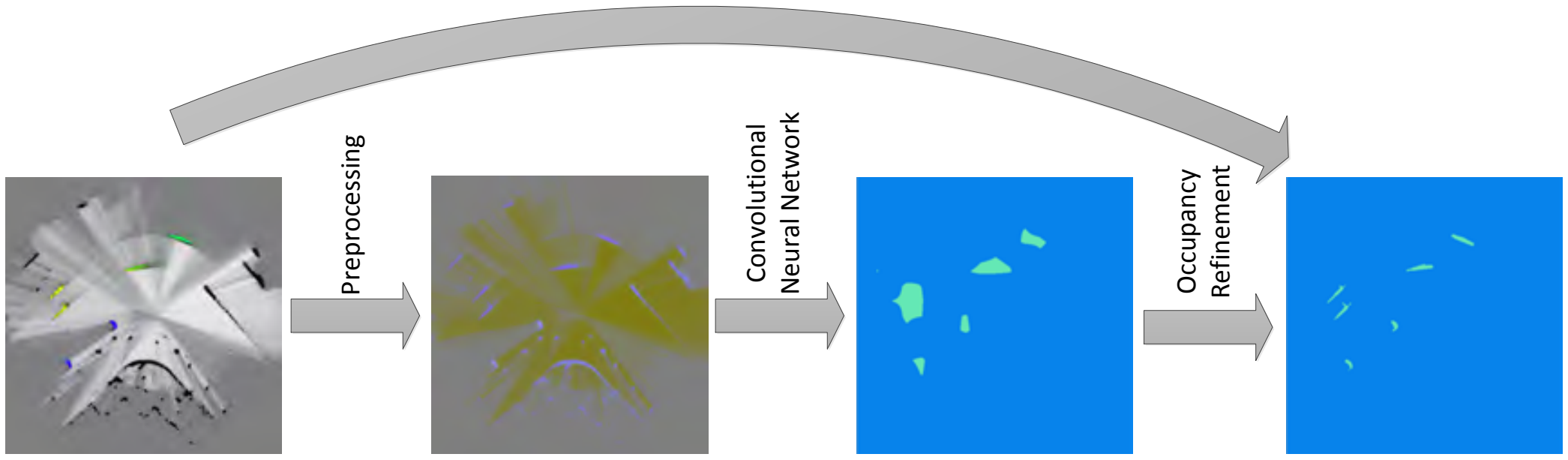

Figure 3.1.: Overview of segmentation steps: the raw DOG is preprocessed, to extract a specific input for the CNN. Afterwards the output of the network is refined with the occupancy information of the DOG. The final output contains the static labels (blue) and the dynamic labels (turquoise).

### 3.1.1. Input

Based on the input data, the CNN calculates the pixelwise classification. This data has to contain sufficient information to extract an accurate prediction. For this reason different input data type combinations are provided and evaluated against each other. The following data types can be provided from the DOG:

- Occupancy probability (with and without freespace information)
- Mean velocity in x direction over all particles in one grid cell
- Mean velocity in y direction over all particles in one grid cell
- Variance of velocity in x direction over all particles in one grid cell
- Variance of velocity in y direction over all particles in one grid cell
- Mahalanobis distance of overall velocity in one grid cell.

As described in Section 2.2.2, the DOG contains several particles in each grid cell. These particles have weights, which represents the occupancy probability. This probability is combined with the freespace information, which is provided as well. The result is a value between 0.0 and 1.0, which represents the occupancy state: 0.0 represents free space and 1.0 represents an obstacle, which is shown in Figure 2.5. If the freespace information is not provided, the value is only between 0.5 and 1.0.

Each particle describes a dynamic state as well (see Section 2.2.2). This is described with the velocity in $x$ and $y$ direction. For one grid cell the mean velocity and the variance of the velocities of the particles can be calculated, which represents an uncertainty for the velocities. Additionally, a mahalanobis distance $m$ of the mean overall velocity $\overline{v}_{overall}$ to the velocity $v_{reference} = 0$ can be calculated, as mentioned in [36]:

$$m^2 = \overline{v}_{overall}^T \Sigma^{-1} \overline{v}_{overall} \quad , \tag{3.1}$$

where $\Sigma$ represents the covariance matrix of the velocities.

The structure of the used CNN is designed based on public network structures like VGG-net [41] and Alexnet [26]. This means the input has to be a colored image, otherwise the weights of pre-trained networks cannot be used. For this purpose the three color channels (blue, green, and red) are used to encode the discussed input data. The following combinations of input data are provided (B - G - R):

1. $Occ_{free}$ - $v_x$ - $v_y$
2. $Occ_{free}$ - $v_{x,norm}$ - $v_{y,norm}$
3. $Occ_{nonfree}$ - $v_{x,norm}$ - $v_{y,norm}$
4. $Occ_{free}$ - $v_{overall}$ - $Var_{overall}$
5. $Occ_{free}$ - $v_{overall}$ - $m$

In every combination the occupancy probability is included. The reason for this is that the CNN should use this occupancy information to include shape recognition of the obstacles for the prediction. The third combination excludes the freespace information to analyze the influence of freespace information. All the other parameters are used to represent the movement of grid cells. Besides the velocity itself, the normalized velocities $v_{x,norm}$ and $v_{y,norm}$ can be provided. The normalized velocity in $x$ and in $y$ direction can be calculated accordingly, where the normalized velocity in $x$ direction is defined as

$$v_{x,norm} = \frac{v_x}{\sqrt{Var(v_x)}} \quad . \tag{3.2}$$

The normalization should help to distinguish between clutter and real moving objects. This is caused by the DOG: grid cells, which belong to clutter, contain particles with different velocities. This can potentially be detected with the variance over these velocities. Additionally, in combination four and five, the parameters for the uncertainty are provided in a separate input channel. One of these parameters is the overall variance, which is calculated with

$$Var_{overall} = Var(v_x) + 2Cov(v_x, v_y) + Var(v_y) \quad . \tag{3.3}$$

Another parameter is the mahalanobis distance (see Equation 3.1). The problem of these combinations is the restriction of three input channels. This causes the velocities to be combined to the overall velocity $v_{overall}$.

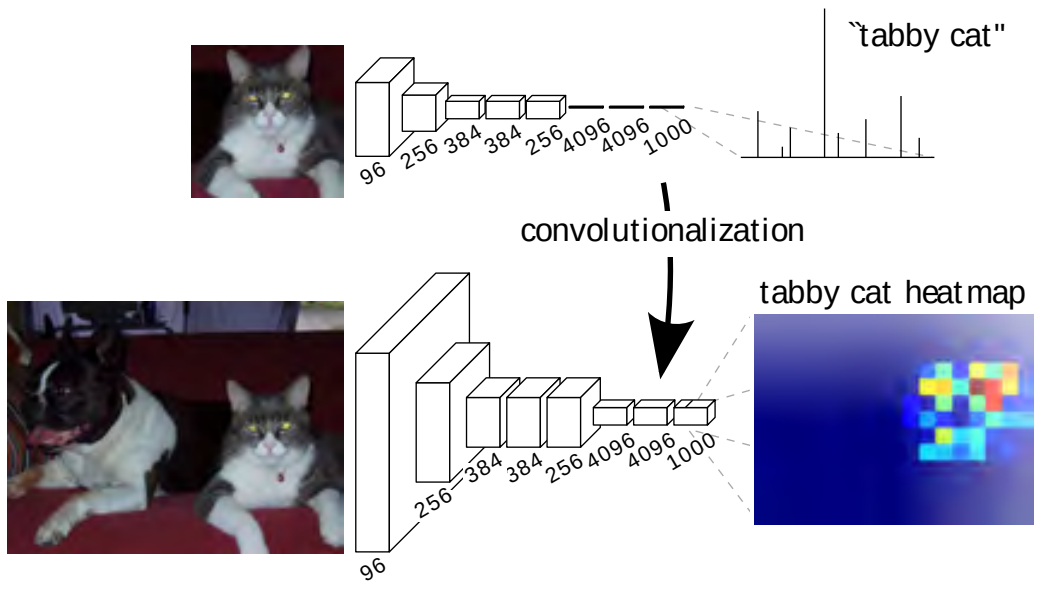

Figure 3.2.: Conversion of fully connected layers to fully convolutional layers [28, p. 3433].

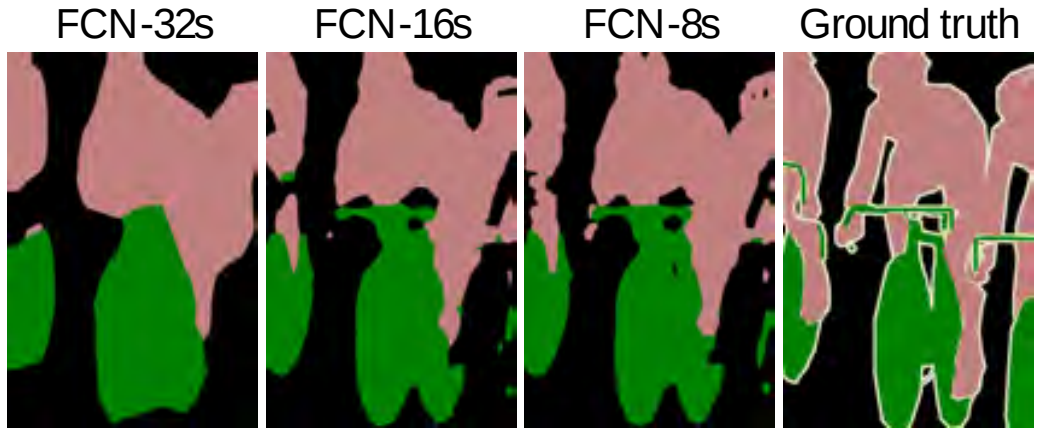

Figure 3.3.: Refinement of the segmentation with the deep jet and a resulting smaller stride at the last layers. The first images show the output from the stride 32, 16, and 8 network [28, p. 3436].

### 3.1.2. Range

The CNN is initialized with the weight of pre-trained networks, which were trained with colored images from the public Imagenet [13] dataset. This means that every input channel of the CNN has a range of $[0,255]$, i.e. the input values from the previous section have to be discretized to this range. The occupancy probability with a range of $[0,1]$ can easily be transferred to the required range by using a multiplication factor of 255. For the other parameters, a mapping to the required range has to be applied. For this purpose, the following input limits $l$ are evaluated and transferred to the range of $[0,255]$:

$$\begin{gathered} l = [-t,t] \quad , \\ t \in 5,10,15,20,25 \quad . \end{gathered} \tag{3.4}$$

These ranges are applied to the normalized and unnormalized velocities, the variances, and the mahalanobis distance.

### 3.1.3. Structure

Based on Long et al. [28], public network stuctures are used and converted to fully convolutional layers as mentioned in Section 2.3.3. This is executed by transferring weights of fully connected layers to convolutional layers as shown in Figure 3.2. Two network structures are used in this thesis: VGG-net [41] and Alexnet [26]. VGG-net is deeper than Alexnet, what means that it contains more layers. That causes on the one side a longer execution time. On the other side Long et al. [28] showed a higher accuracy.

The original VGG-net contains 16 weight layers (convolution and fully connected) and five max pooling layers. Each convolution layer besides the first has a stride of one, a filter size of three, and a padding of one. This ensures that input and output of a convolutional layer have the same size. All the pooling layers have a filter size and a stride of two, what produces a reduction of the input size by factor two. For further details, the reader is referred to [41].

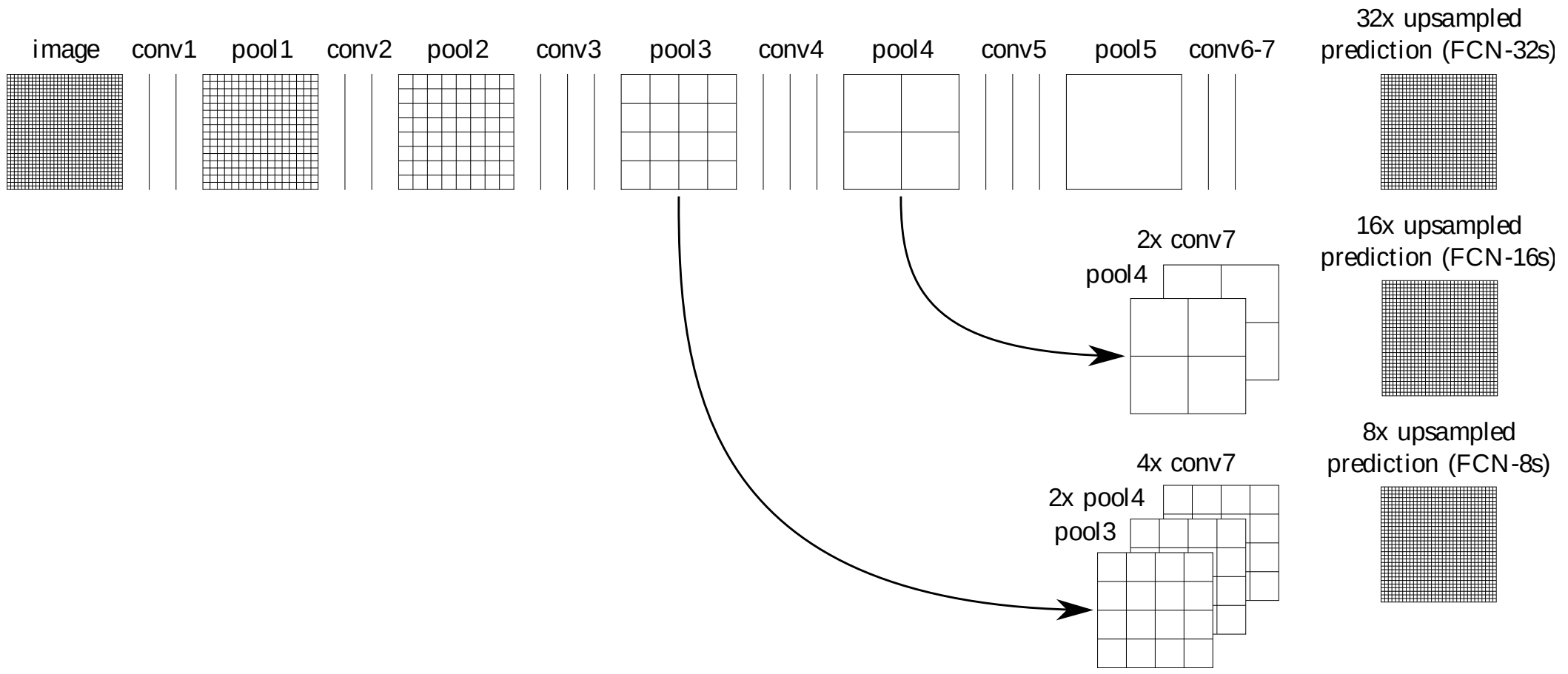

Figure 3.4.: Deep jet introduced to the VGG-net: pooling and prediction layers are shown as grids that reveal relative spatial coarseness, while convolutional layers are shown as vertical lines. First row (FCN-32s): up-sampling with a stride 32 in a single step. Second row (FCN-16s): Combining predictions from both the final layer and the pool4 layer, at stride 16, to predict finer details, while retaining high-level semantic information. Third row (FCN-8s): Additional predictions from pool3, at stride 8, provide further precision. [28, p. 3435].

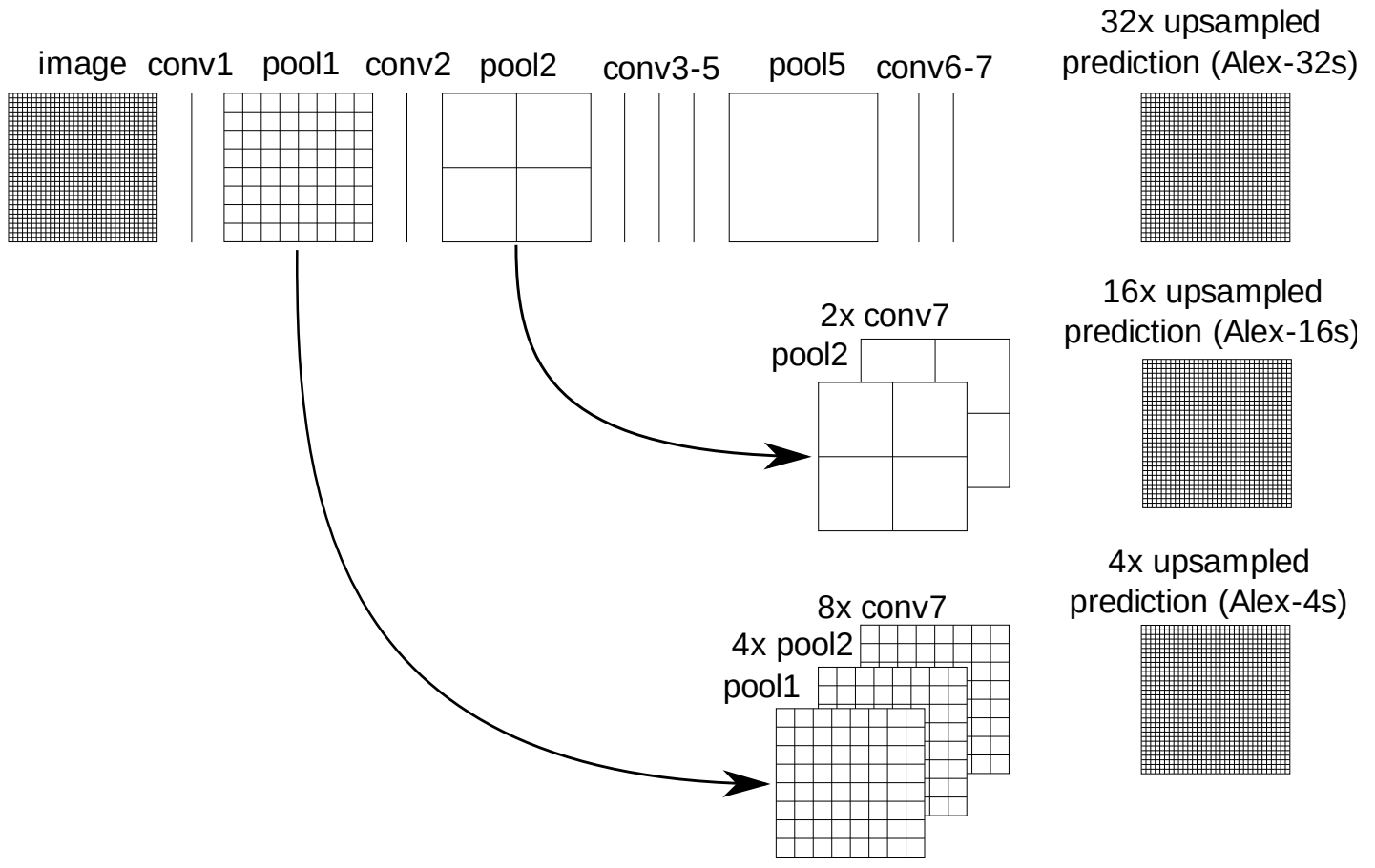

Figure 3.5.: Deep jet introduced to the Alexnet: pooling and prediction layers are shown as grids that reveal relative spatial coarseness, while convolutional layers are shown as vertical lines. First row (Alex-32s): up-sampling with a stride 32 in a single step. Second row (Alex-16s): Combining predictions from both the final layer and the pool2 layer, at stride 16, to predict finer details, while retaining high-level semantic information. Third row (Alex-4s): Additional predictions from pool1, at stride 4, provide further precision. (modeled after [28, p. 3435]).

The original Alexnet is not as stepwise structured as the VGG-net. It contains seven weight layers (convolution and fully connected) and three pooling layers, but the smallest feature map has the same size as the VGG-net. This is possible because of the first convolutional layer, which has a filter size of eleven and a stride of 4. In total Alexnet is much smaller, what results in a shorter execution time. This should be used as a reference in terms of accuracy.

As last layer, a deconvolutional layer, as described in Section 2.3.2, is introduced. Caused by the network structure, the single deconvolutional layer has a stride of 32 to upscale the image to the input size. This limits the scale of details of the segmentation because the inner representation is 32 times smaller than the final output. To improve this, Long et al. [28] introduced a stepwise deconvolution, called deep jet. There, each deconvolution step is fused with lower layers to recombine higher semantic information with lower shape information, as shown in Figure 3.4. The last deconvolution layer ten uses a smaller stride, what produces a finer segmentation output (see Figure 3.3). For the VGG-net, three variations were introduced:

- VGG-net with a stride of 32 at the last deconvolutional layer (FCN-32s)
- VGG-net with a stride of 16 at the last deconvolutional layer (FCN-16s)
- VGG-net with a stride of 8 at the last deconvolutional layer (FCN-8s)

The training of the structures with smaller strides can be achieved with different approaches, which are proposed from Long et al. [28]. On the one side the network can be trained non-incremental. That means the network with the smaller strides is initialized with the pre-trained network and trained afterwards for certain iterations. On the other side the network can be trained incremental. That means the network with the largest stride is trained first. Afterwards this network is used to initialize the network with the next smaller stride and so on.

In this thesis, deep jet is additionally applied to Alexnet to achieve a finer segmentation output. Because of the different network structure, the possible strides differ from the ones applied to VGG-net. The main structure of the Alexnet is shown in Figure 3.5. Three possible deep jet versions are evaluated:

- Alexnet with a stride of 32 at the last deconvolutional layer (ALEX-32s)
- Alexnet with a stride of 16 at the last deconvolutional layer (ALEX-16s)
- Alexnet with a stride of 4 at the last deconvolutional layer (ALEX-4s)

### 3.1.4. Zooming into the Input

The size of the DOG is predefined. That means the dimension of obstacles like vehicles or pedestrians is predefined as well. Additionally the filter sizes of the CNN are also constant. That causes the network to recognize only specific sizes of objects, especially with the initialization with the pre-trained network. To evaluate different sizes of obstacles, the DOG is cropped in the middle and scaled afterwards to the origin input dimensions, as shown in Figure 3.6. That produces an adaptable obstacle size while using a constant image size for the network, so the network structure has not to be adapted for each input size. The following crop sizes $cs$ are evaluated:

$$cs \in \{300 \times 300, 400 \times 400, 500 \times 500, 600 \times 600\} \qquad [3.5]$$

where the origin image size is $600 \times 600$. This implies that the crop size $300 \times 300$ is a doubling of the object size and the crop size $600 \times 600$ is the origin object size.

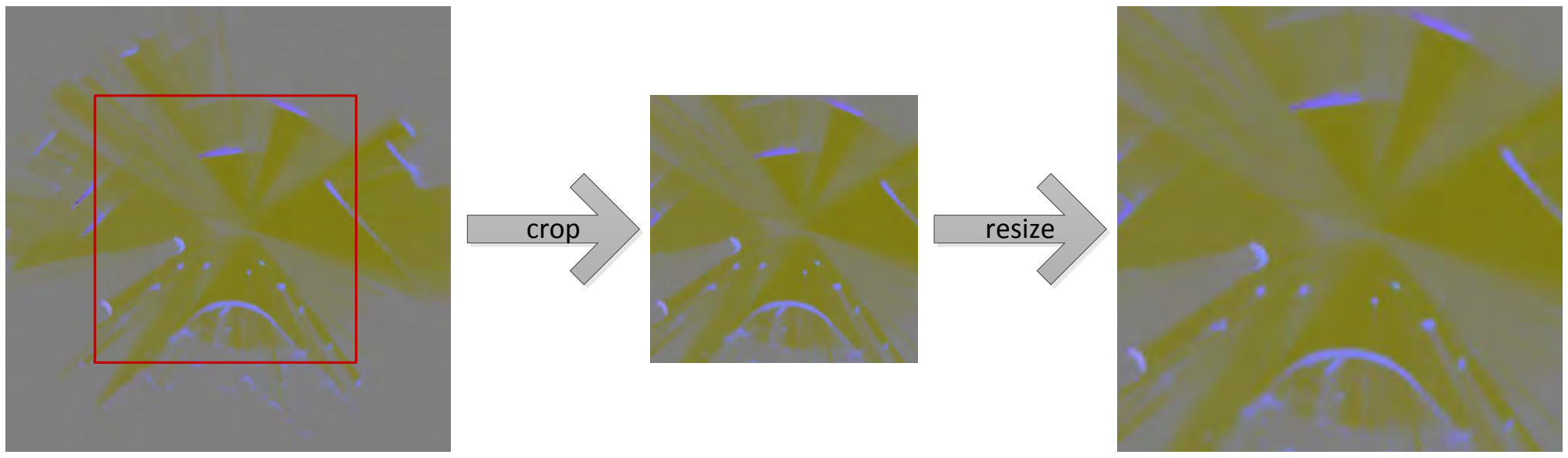

Figure 3.6.: Zooming into the input: first the input is cropped and afterwards it is resized to change the size of the obstacles by using a constant input size for the CNN.

### 3.1.5. Learning Rate

Long et al. [28] and Cordts et al. [9] provided learning rates to train the Fully Convolutional Neural Networks. These learning rates were unnormalized and accordingly dependent on the input image size. In this thesis, these learning rates are transferred to normalized values, to get rid of numerical issues and to being able using backpropagation independent of the total number of pixels or weights of pixels. The transferred learning rates are fixed learning rates, where the momentum and the weight decay is set according to Long et al. [28].

The type of input images differs from the type used by Long et al. and Cordts et al. For this purpose the learning rates are evaluated in a smaller and a larger range around the proposed values. The evaluated learning rates $\mu$ are

$$\mu \in \{2.14 \times 10^{-6}, 7.14 \times 10^{-6}, 2.14 \times 10^{-5}, 3.6 \times 10^{-5}, 7.14 \times 10^{-5}, 2.14 \times 10^{-4}\} \ , \qquad [3.6]$$

where the learning rates $2.14 \times 10^{-5}$ and $3,6 \times 10^{-5}$ represent the transferred learning rates of Cordts et al. and Long et al.

Additionally a step learning rate policy is evaluated, where a higher learning rate can be chosen at the beginning of the training. Afterwards the learning rate is decreased in steps over the iterations. It is expected that the pre-trained weights of the first layers can be easier adapted to the different type of input with a higher learning rate at the beginning of the training. After the reduction of the learning rate the last layers should be finetuned to reach a convergence. To achieve that, the start learning rate is chosen ten times higher than the learning rate of the fixed learning rate policy. Thereby the step size is chosen in relation to the total training iterations, so that the learning rate decreases to the learning rate of the fixed learning rate policy after half of the training iterations.

### 3.1.6. Weight Matrix

As it is shown in Figure 2.5 or Figure 3.1, the number of static pixels (including the background) is much higher than the number of dynamic pixels. That can cause problems with the backpropagation algorithm to a higher affinity to the static class. For this purpose a weight matrix $C$ is introduced in the cost function $J(\theta)$ of the multinomial logistic loss (see Section 2.3.4):

$$J(\theta) = -\left[\sum_{i=1}^{m}\sum_{k=1}^{K} C^{\left(y^{(i)}\right)} 1\{y^{(i)} = k\} \log P\left(y^{(i)} = k | x^{(i)}; \theta\right)\right] \quad , \qquad [3.7]$$

where $C$ contains a weight for each labeled class:

$$C = \left[c^{(1)} c^{(2)} \dots c^{(K)}\right] \quad . \qquad [3.8]$$

## 3.2. Orientation Extraction

The CNN from Section 3.1 creates a pixelwise prediction for the DOG. This segmentation should improve a cluster and tracking algorithm as described in Section 2.3.5. To provide more information to the tracking algorithm, the orientation of the object should be extracted. The orientation in this context is represented by the moving direction. To extract the moving direction of the obstacle two approaches are compared to each other: orientation extraction through the CNN and over the mean velocities in $x$ and $y$ direction, what is discussed in the further subsections.

### 3.2.1. Orientation Extraction with Convolutional Neural Networks

One approach is to extract the orientation through the CNN. This causes some problems with the periodicity of the angle, as described in Section 2.5. For this purpose the biternion representation with sine and cosine of Beyer et al. [3] is used. To regress the angle, the network structure is adapted as shown in Figure 3.7. There, the structure of the pre-trained network is not changed, besides the transformation to a FCN. The extension with deconvolutional layers in the end of the network is replicated twice to create the same structure once for the regression of the sine and once for the regression of the cosine. Only the last layer, which is a softmax layer for the classification, is skipped for the regression. This produces a pixelwise regression for the sine and cosine, which can be recombined to an angle. To produce a clusterwise orientation extraction, clusters can be produced over the pixelwise classification output of the network. Then a mean of sine and cosine can be calculated over the clusters to extract the cluster angles.

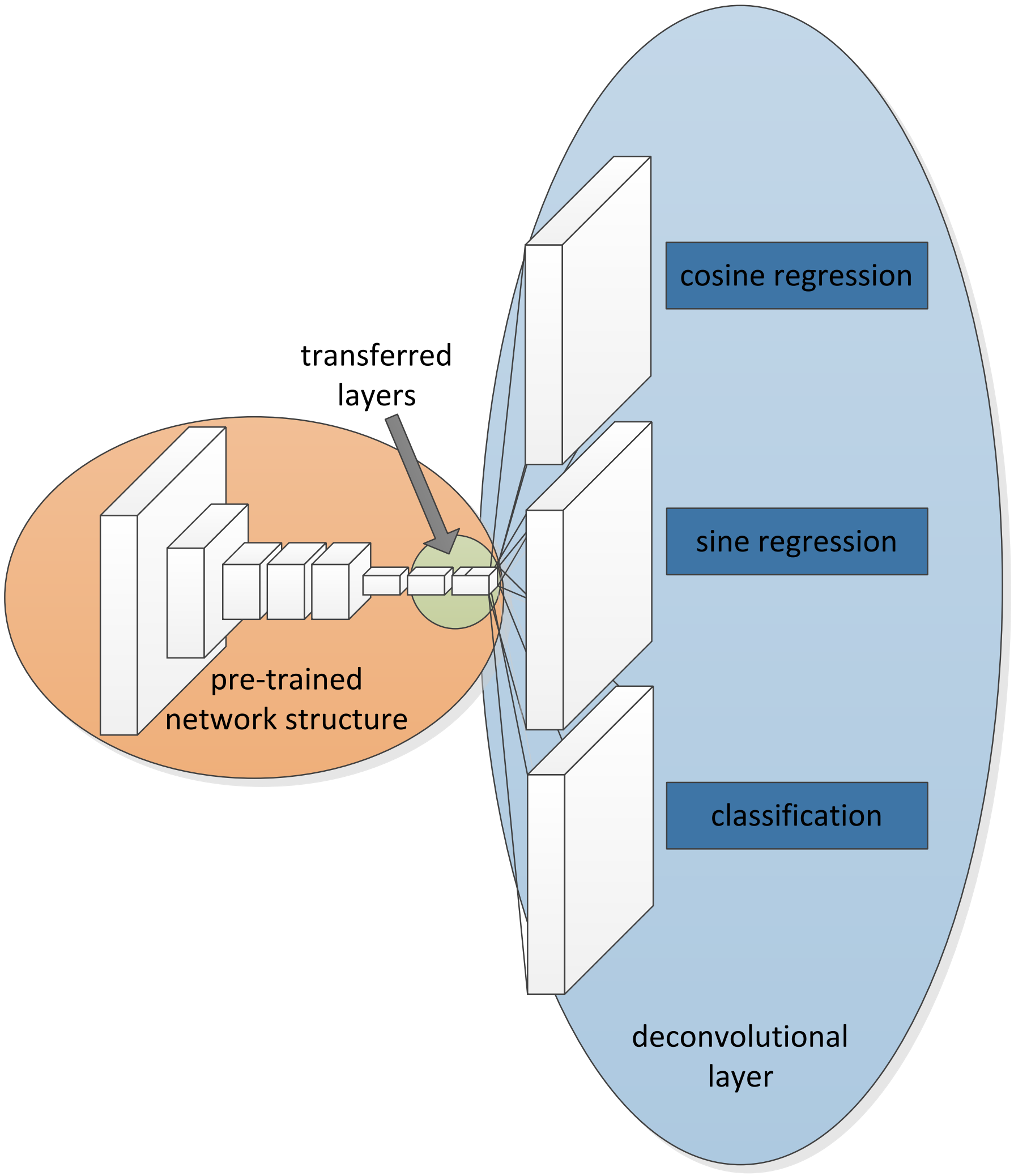

Figure 3.7.: Adaption of CNNs for orientation extraction: two more output maps (deconvolutional layer) are introduced for the regression of sine and cosine. The pre-trained network structure is not changed, besides the last layers, which are transfered from fully connected to fully convolutional layers.

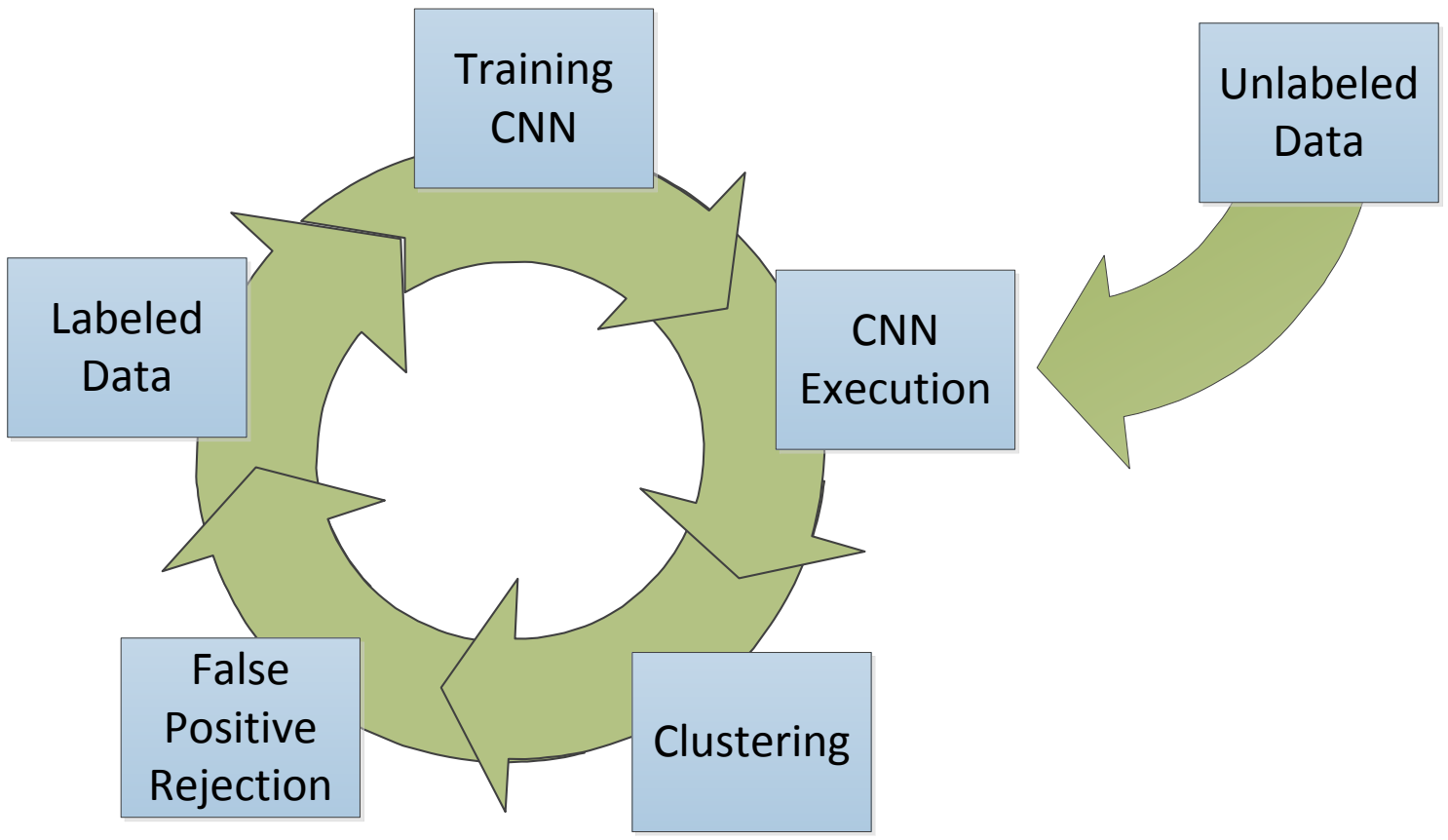

Figure 3.8.: Semi-supervised training loop.

### 3.2.2. Orientation Extraction over Velocities

Another approach is to extract the orientation by directly using the velocity information, which is provided by the DOG. The output of the pixelwise classification (see Section 3.1) is used to create clusters of dynamic objects. Based on this information a mean of the velocities $\overline{v}$ in $x$ and $y$ direction can be calculated, which can be used to extract the angle $\phi$ of the clusters with

$$\phi = \arctan\left(\frac{\overline{v}_y}{\overline{v}_x}\right) \quad . \qquad [3.9]$$

## 3.3. Semi-Supervised Learning

A time and cost expensive task of training deep neural networks is the labeling of data (see Section 2.4). For this purpose a semi-supervised approach based on Scudder et al. [40] is evaluated. There, the machine learning approach is trained on manually labeled data. Afterwards, unlabeled data is labeled by the machine learning approach automatically. This produces a larger training database, which should improve the accuracy. Especially to reduce clutter, the machine learning approach can generalize over a lager dataset.

Additionally, to reduce false detected objects in the automatic labeling step, the labeled data is corrected. For this purpose, the pixelwise labeled data is clustered and a parameter is calculated for each cluster, which is used to reject false positives. As parameter the mean normalized velocity of a cluster and a combination $p$ of the mahalanobis distance $m$ (see Equation 3.1) and mean velocity $\overline{v}$ of a cluster is used.

$$p = m\overline{v} \tag{3.10}$$

This parameter is chosen, because of the velocity independent measure with the mahalanobis distance on the one side and the velocity itself on the other side. That produces a parameter, where the value decreases with a smaller velocity and a smaller uncertainty by using the mahalanobis distance. Finally, obstacles that move either slowly or have a high uncertainty can potentially be filtered, what increases the quality of the automatic labeled data. The whole semi-supervised training loop is shown in Figure 3.8.

# 4. Realization

In this chapter the realization of the methods is discussed. First, the data generation is described. Afterwards, the proceeding for the comparison with other approaches is defined. Finally, the used framework and the training hardware are presented.

## 4.1. Data Generation

In this section, the data generation for training and evaluation is defined. First, the route of the recording vehicle is described. Afterwards, the semi-automatic labeling and the split of data are introduced. Finally, the rotation of the training data to receive a higher orientation variance is specified.

### 4.1.1. Recording Route

For training machine learning approaches, it is important to generate a large database with high diversity in the data. In order to obtain this data, a route was recorded 4.3 miles through different areas in Sunnyvale (California), as shown in Figure 4.1. The first area (green ellipse) is a six-lane street, with a speed limit of 45 miles per hour. There, several passing maneuvers of the recording vehicle and from other vehicles are captured. Additionally, some traffic lights are located in this area, where several vehicles crossed the street. After a left turn the recording vehicle entered the downtown area, which is zoomed in Figure 4.1. There, pedestrians, bicycles, and vehicles with a lower speed were recorded as well as parked vehicles, which were not moving. The recording vehicle also entered a limited traffic zone (orange ellipse) and a parking lot of a shopping center (red ellipse) with several pedestrian crossings to generate enough data. Finally, the recording vehicle drove back on the six-lane street.

The whole recording contains different scenarios like larger streets, traffic lights, pedestrian crossings, and limited traffic zones, as shown in Figure 4.3. This should result in a certain diversity of the data to facilitate generalization of urban areas for the machine learning approach.

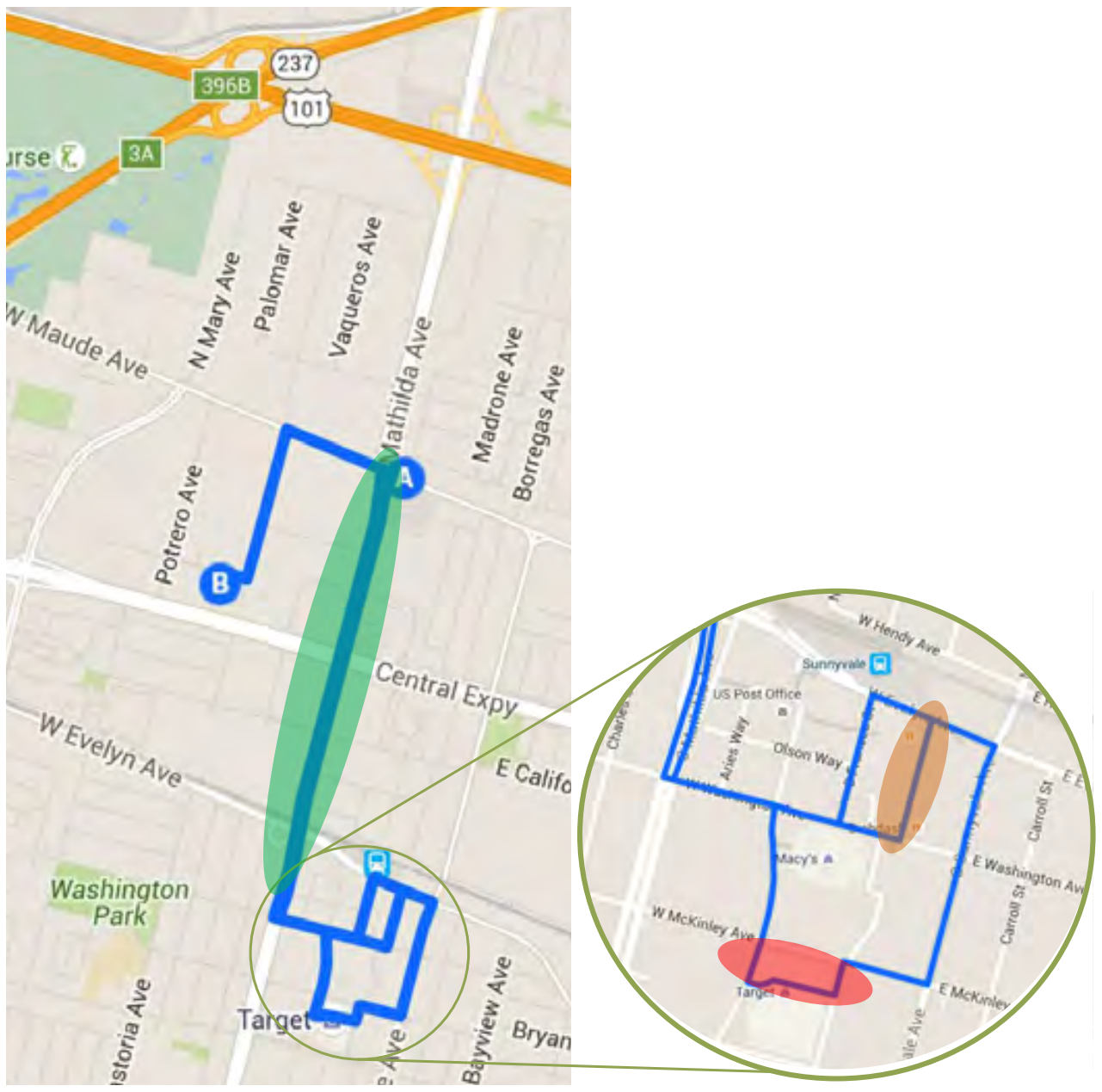

Figure 4.1.: Route for recording the data: the route was defined through different areas in Sunnyvale (California): a six-lane street (green ellipse), a limited traffic zone (orange ellipse), and a parking lot of a shopping center. The zoomed area is the downtown area of Sunnyvale.

### 4.1.2. Semi-Automatic Data Labeling

After recording the data, the labels have to be created. For this purpose, a semi-automatic way is followed. Labels are first generated automatically, clustered, and afterwards corrected manually, as shown in Figure 4.2.

To distinguish between dynamic and static objects in an automated way, the appraoch of Nuss et al. [36] is used as baseline method. There, the velocities of the particles within a grid cell of the DOG and their variances are fused to the mahalanobis distance, which was already defined in Equation 3.1. The mahalanobis distance describes the distance of a point to a distribution, where the standard deviation $\sigma$ is taken into account. Therefore, a mahalanobis distance of 3 corresponds to a distance of $3\sigma$ of the value to the distribution. In case of the DOG, Nuss et al. [36] calculate the distance of the velocity $v_{reference} = 0$ to the distribution of the particle velocities within a grid cell. As a result, the mahalanobis distance is used as a measure for the certainty of movement, what is shown in Figure 4.4. Based on this information, a threshold to distinguish between dynamic and static obstacles was calculated by Nuss et al.

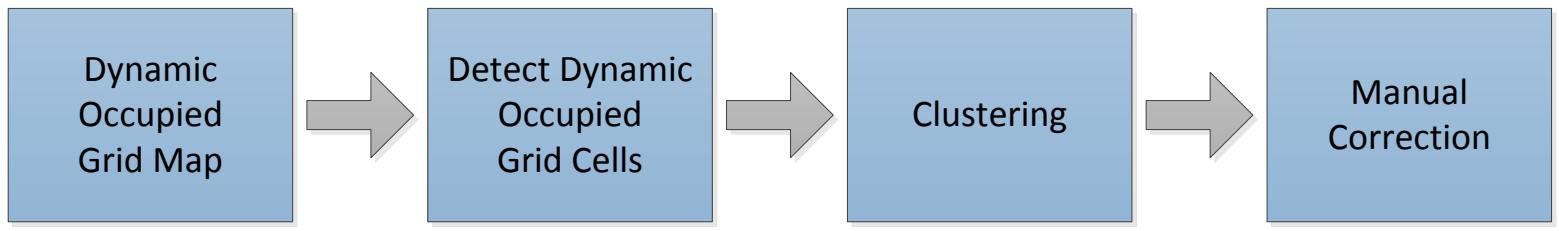

Figure 4.2.: Overview of semi-automatic labeling toolchain.

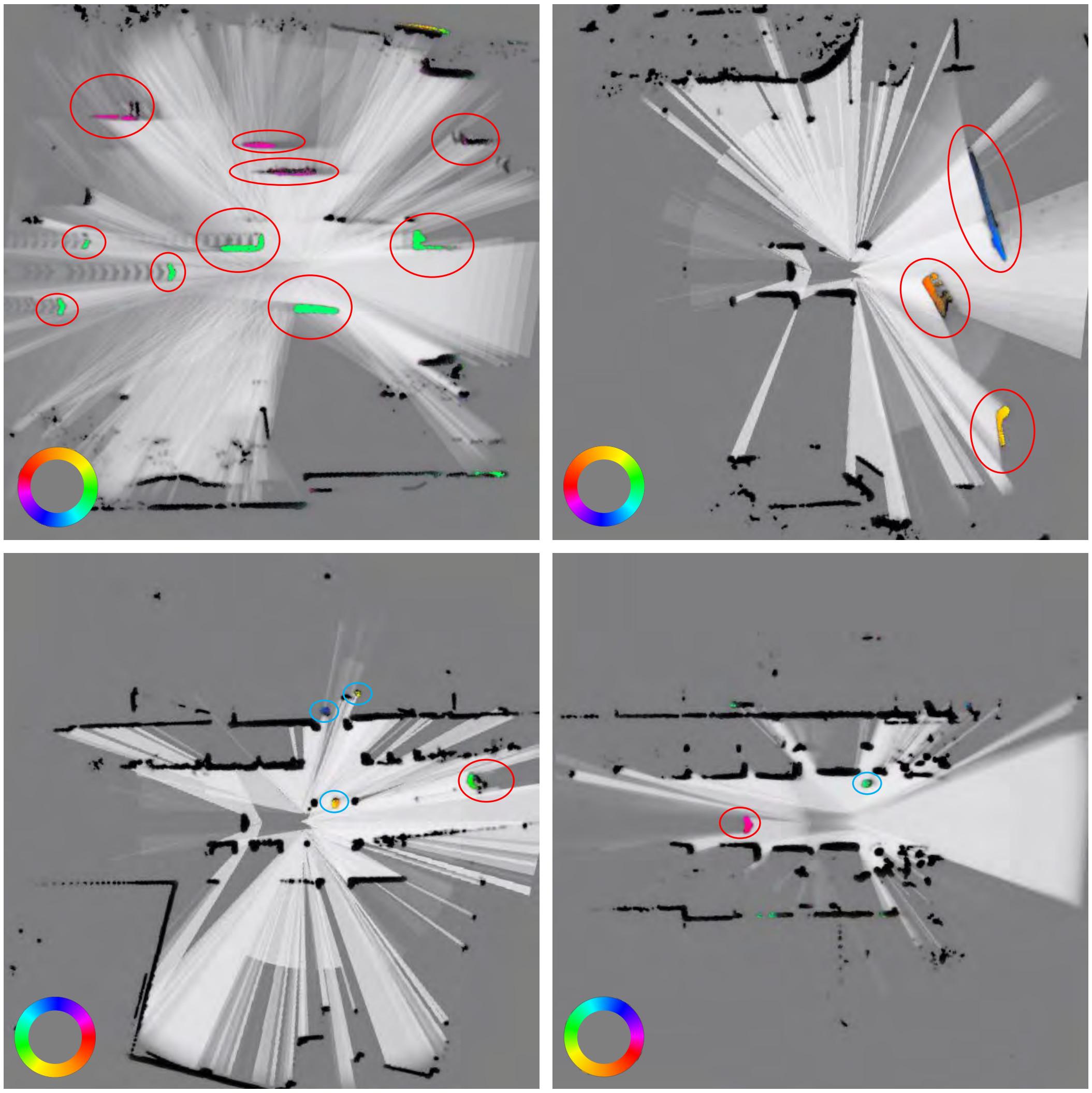

Figure 4.3.: Examples of grid map recordings: the recording vehicle is placed in the middle of each image and its moving direction is to the right. Moving vehicles are labeled with a red ellipse, moving pedestrians are labeled with a blue ellipse, and the colored areas encode the velocity information including the angle, which is shown within the circle. The upper left image shows a six-lane street, the upper right image shows a left turn on a traffic light, the lower left image shows a pedetrian crossing, and the lower right image shows a limited traffic zone.

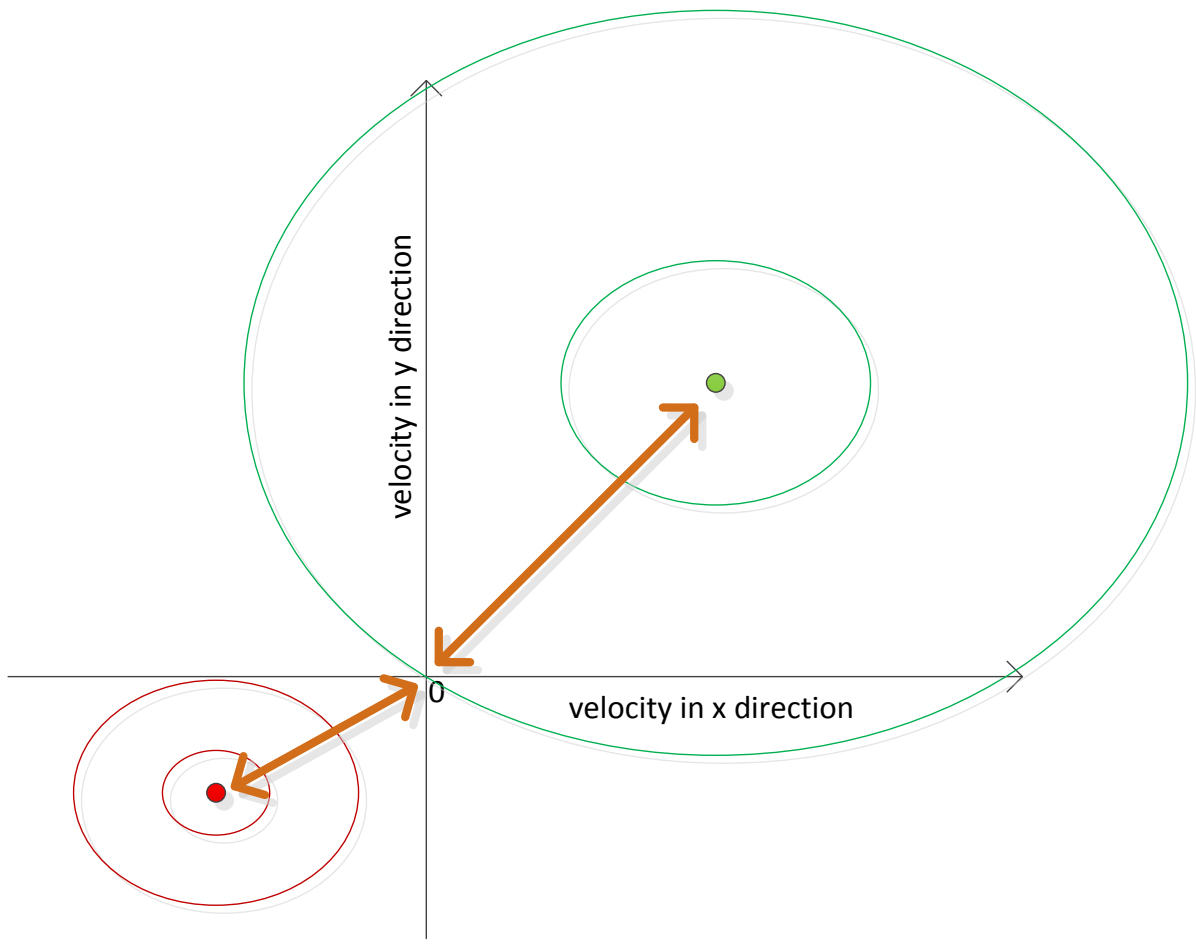

Figure 4.4.: Example of the mahalanobis distance: the green point describes a grid cell with a high mean velocity and a high variance of the particle velocities. The red point describes a grid cell with a lower mean velocity and a lower variance of the particle velocities. As a result the red point has a greater mahalanobis distance than the green point.

The pixelwise decision for dynamic obstacles has to be improved to remove false positives and false negatives. For this purpose, the output of the pixelwise decision is clustered based on the DBSCAN algorithm introduced by Ester et al. [17] and transferred to the DOG by Peukert [39]. Afterwards, the convex hull of these clusters is provided to correct the data manually. Additionally, the orientation for each cluster is extracted by using the mean velocity in $x$ and $y$ direction over the cluster of grid cells, like in Section 3.2.2. This is used to generate the ground truth for the evaluation of the orientation extraction.

The manual correction finalizes the ground truth for evaluation of the different goals of this thesis. During this process, not all of the data is used. Especially if the recording vehicle had to wait on a traffic light for a left turn or if the recorded images are too similar, the annotation is skipped. Overall, 3 457 images were labeled with this semi-automatic approach.

### 4.1.3. Splitting the Data

The goal, while training machine learning approaches, is to generalize and not to overfit over the training data. Overfitting means in this context, that the machine learning approach adapts perfectly to the training data, but does not generalize over it, as shown in Figure 4.5.

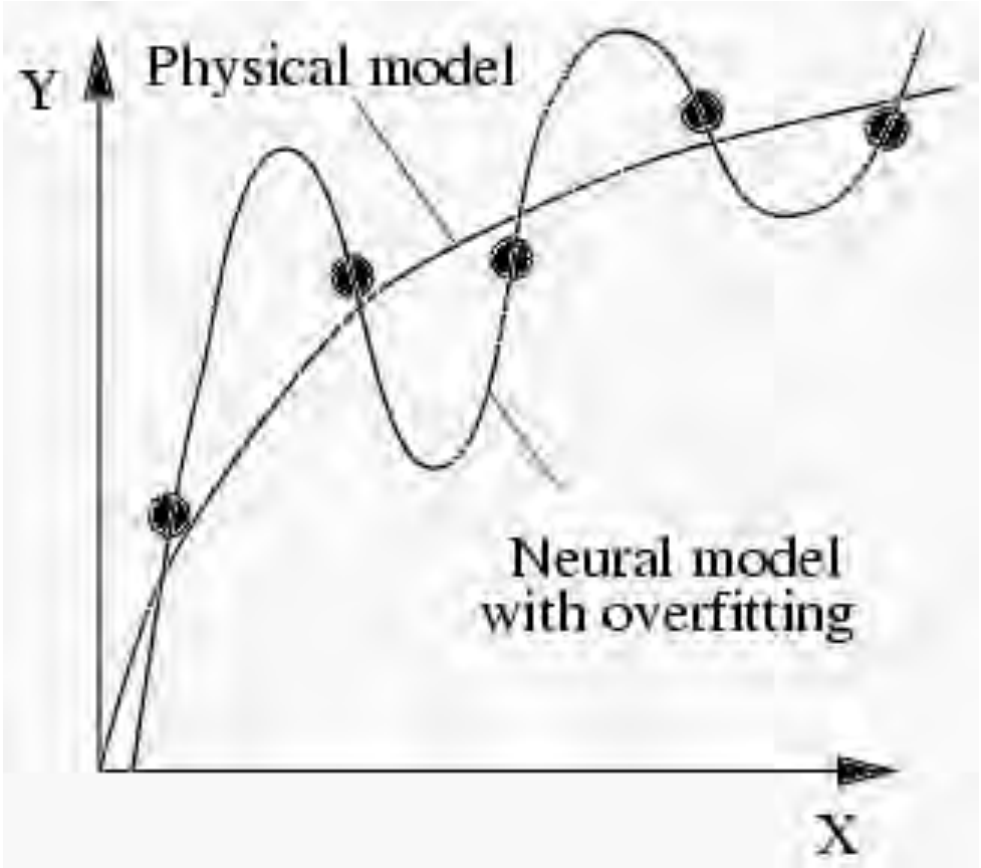

Figure 4.5.: Overfitting on a physical model [8].

| dataset | number of images | scenarios |
|---|---|---|
| testing | 369 | pedestrian crossing, six-lane street, limited traffic zone |
| validation | 343 | parking of a shopping center, left turn on traffic light of a six-lane street, right turn in the downtown area |
| training | 2745 | rest of the recording |

Table 4.1.: Split of data with the different scenarios.

To evaluate the generalization, the data has to be split in a training and a test dataset, where the test dataset is not used for training. Furthermore, the parameters of the CNN are optimized on an additional dataset. For this purpose a validation dataset is created. This allows to train the CNN on the training dataset, to optimize the parameters of the CNN by evaluating on the validation dataset, and finally to evaluate the CNN against other approaches on the test dataset.

The problem of splitting the data is that the number of training samples decreases. Therefore, the validation and test dataset should not be as large as the training dataset. On the other side, the validation and test dataset have to be large enough to be representative. In this thesis, a split of roughly 80% - 10% - 10% for training, validation, and test data is chosen.

The labeled data is generated from one recording, where around 100 milliseconds are between two images. As a result, the split of the data cannot be established randomly. Otherwise, the datasets have a high correlation and the split will not be meaningful. For reducing the correlation, the split is done manually, where two policies are complied. First, the selected data has to be representative and should contain a large street scenario, pedestrians, and scenarios with a lot of clutter. The second policy is, that between the different sets of data the appearance should change, so that correlation decreases. This is ensured by using a minimum time difference between two images of different datasets. Finally the data is divided into three dataset as mentioned in Table 4.1.

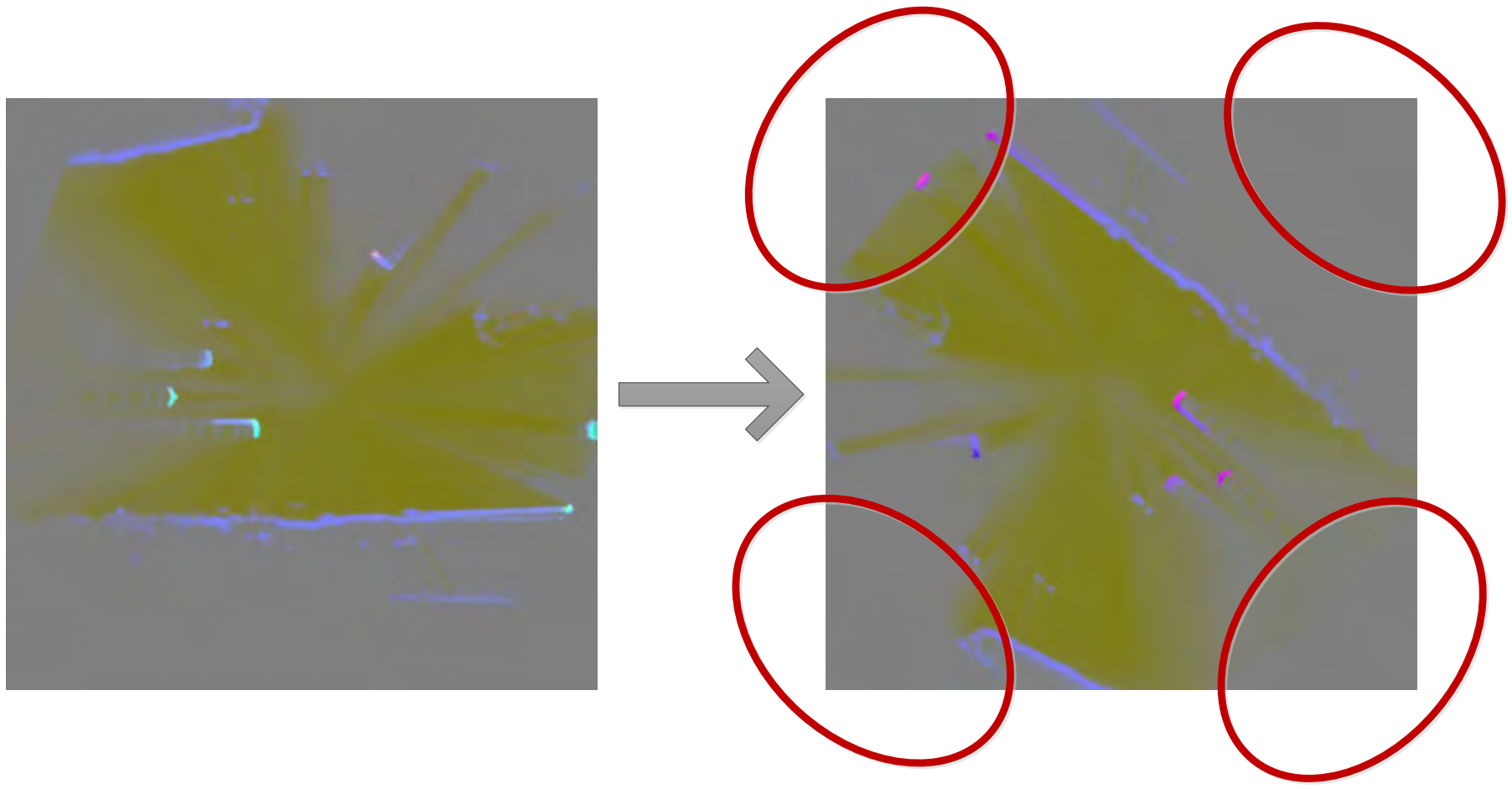

Figure 4.6.: Rotation of a training images rotated 140° counter clockwise: the corners (red ellipses) are filled with an uncertainty occupancy $P(O_k^c) = 0.5$ and a velocity, variance and mahalanobis distance of zero.

### 4.1.4. Rotation

Within traffic scenarios, vehicles can appear at different positions and with different orientations. This has also to be encoded in the training dataset. The different positions are captured with the different scenarios within the recording. For the different orientations, the recordings have to be much longer to produce different views of orientations. For this purpose the orientation variance is introduced afterwards in the training dataset by rotating the images in ten degree steps. This produces on the one side rotation variance in the dataset and on the other side the dataset grows by the factor of 36. Especially for the orientation extraction (see Section 2.5) it is required to have high variation of the orientation angle within the dataset.

The rotation of the images is calculated on the input images of the CNN. These images contain the velocity in certain directions. Thus, the velocity information has to be rotated as well. The corners of the rotation, where no information is provided, are set to an uncertainty occupancy $P(O_k^c) = 0.5$ and a velocity, variance and mahalanobis distance of zero, as shown in Figure 4.6. As a result, the datasets size increases according to Table 4.2. The network is trained on images with rotation and the evaluation is done on original images. That means the rotation of the testing dataset is never used.

## 4.2. Comparison to other Approaches

After the optimization of the CNN parameters, the approach of this thesis is compared to Nuss et al. [36] as a baseline method of grid cell classification of dynamic and static objects, which is described in Section 4.1.2. This approach represents the state of the art, where the mahalanobis distance based on velocities is used as distinguishing criterion. To evaluate the approaches against each other, the training is based on another dataset than the the optimization of the CNN: the training dataset and the validation dataset are merged to create the final training dataset. The evaluation is based on the test dataset, as shown in Figure 4.7. As a result, the final comparison is decorrelated to the optimization steps of the CNN.

| dataset | number of images (original) | number of images (with rotation) |
|---|---|---|
| testing | 369 | not used |
| validation | 343 | 12 348 |
| training | 2 745 | 98 820 |

Table 4.2.: Increse of dataset size by rotation of data: the rotated images are used for training and the original images are used for evaluation. For the final comparison to other approaches the network is trained on the rotated training and validation dataset and evaluated on the original testing dataset. All other optimization and evaluation is done on the original validation dataset, by using the rotated training dataset for training.

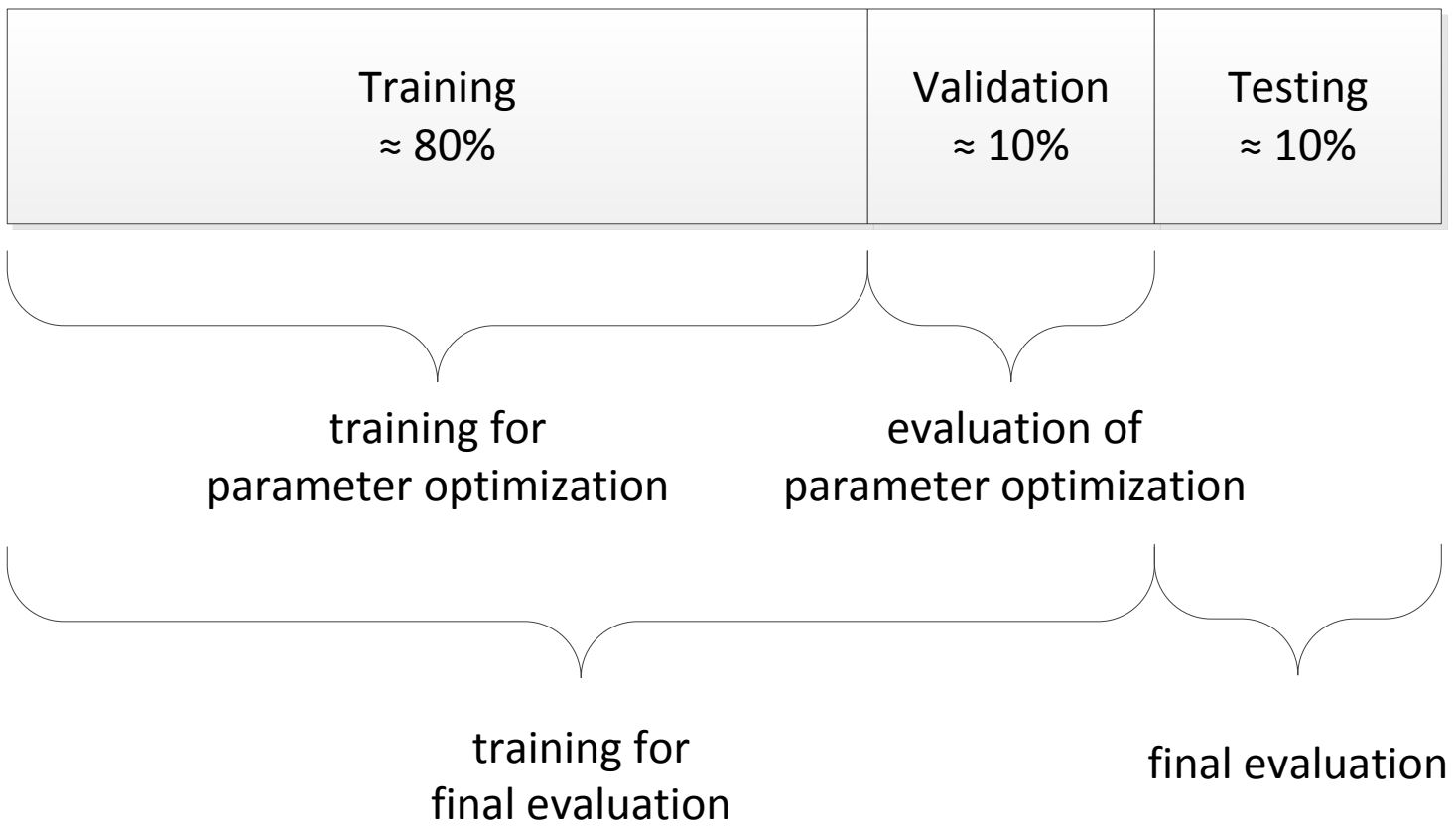

Figure 4.7.: Utilization of the different datasets.

## 4.3. Framework and Hardware

Timing of the CNN is one part of the thesis. Therefore the used Framework as well as the hardware is important. The CNN is developed with the branch “future” of the caffe framework of Long et al. [28]. This branch contains additional layers, like the deconvolutional layer for the FCNs. The branch is adapted at some point to extend the features, like the weight matrix of Section 3.1.6. The CNNs are trained on a Nvidia Geforce GTX 980 Ti and the timing is measured on a Nvidia Geforce GTX Titan X. The used Nvidia driver version is 352.93 with a CUDA version of 7.5 and the operating system is Linux Ubuntu 14.04.

# 5. Evaluation

In this chapter, the evaluation of the goals of this thesis is discussed. First, the segmentation of the DOG is evaluated. Afterwards, the results of the angle regression and the semi supervised learning approach are described. Finally the CNN is compared to the referenced pixelwise classification approach of Nuss et al [36].

## 5.1. Segmentation

In this section, the results of the segmentation are presented. First, a general description of the trainings of the CNNs is given. Afterwards the different optimization parameters of Section 3.1 are evaluated. Each comparison of different configurations is made with ROC curves, which are already described in Section 2.6. As mentioned in Section 3.1.6, the number of background grid cells in the DOG is much higher than the number of occupied grid cells, but in this thesis the focus is the recognition of dynamic objects. For this purpose, the generation of ROC curves uses only occupied grid cells for the pixelwise evaluation to be independent of background of the DOG.

### 5.1.1. Training Configuration

All the parameters are optimized in the same way. For every parameter, the pre-trained CNN is converted to FCN as discussed in Section 2.3.4 and is then trained for 200 000 iterations. Afterwards, the CNN is evaluated with the validation dataset to optimize each parameter. A typical learning curve is shown in Figure 5.1. There, the convergence of the accuracy, precision and recall can be recognized. That means the CNN does not overfit to the training data, especially because the accuracy does not decrease in the end of the training. Additionally, the figure shows a quick convergence, what signifies a fast learning progress.

| Parameter | Value |
|---|---|
| Input | $Occ_{free}$ - $v_{x,norm}$ - $v_{y,norm}$ |
| Input Range | $[-10, 10]$ |
| Network Structure | FCN-32s |
| Crop Size | $500 \times 500$ |
| Learning Rate | fixed policy with $2.14 \times 10^{-5}$ |
| Weight Factor | 60 |

Table 5.1.: Network configuration for the learning curve in Figure 5.1.

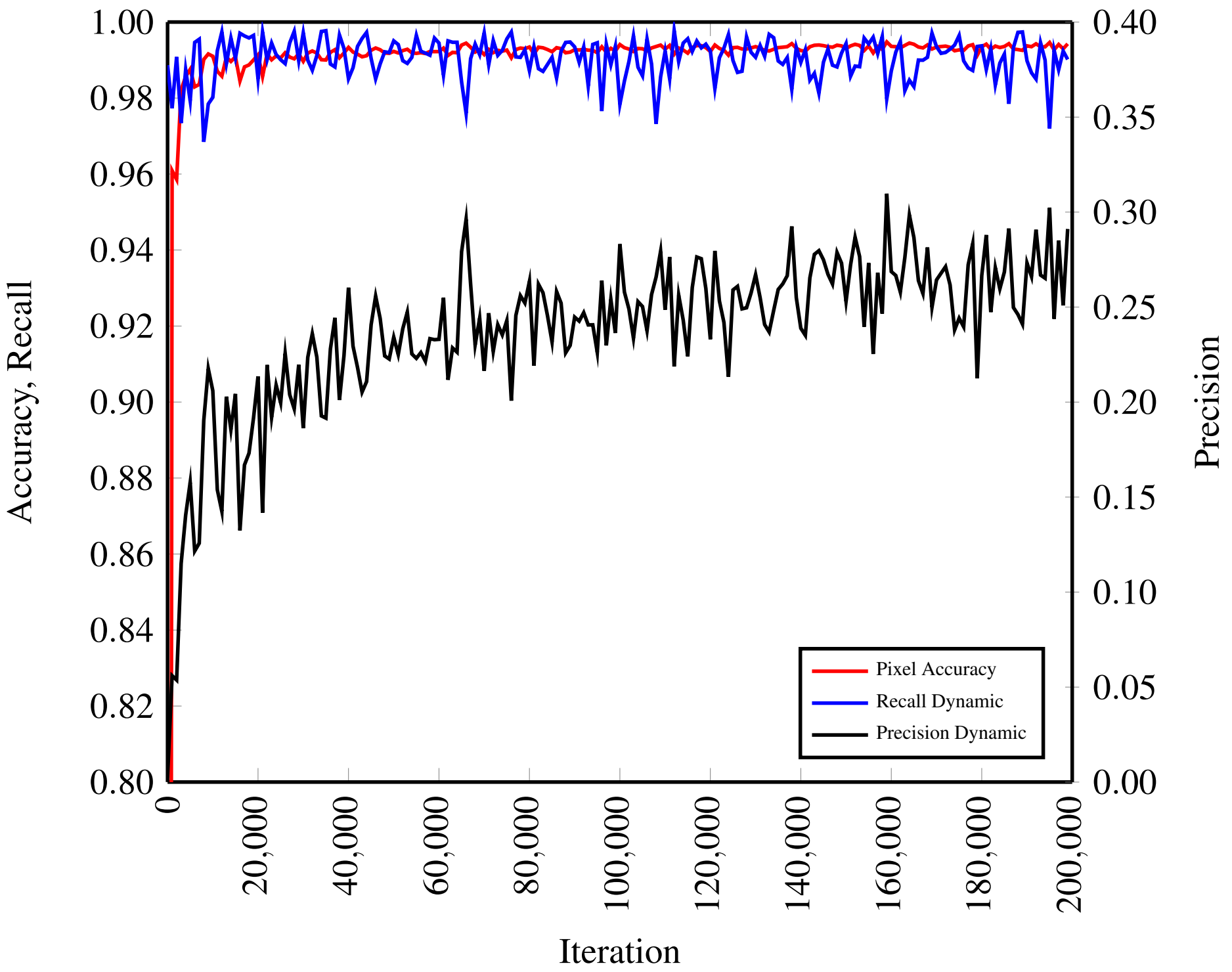

Figure 5.1.: A typical learning curve with the network configuration of Table 5.1.

The learning curve includes also some noise. This noise of the data is caused by the training type: in each iteration only one training sample is used to adapt the network weights ($batchsize = 1$). This results (together with no weight accumulation) in a noisy learning curve.

For every parameter optimization, the CNN is trained once to make the training time feasible. This can cause problems if the training of the network is not deterministic. For this purpose a network configuration (see Table 5.1) was trained several times, what is shown in Figure 5.2. There, a slightly indeterminism can be recognized, what is caused by the random initialization of the newly inserted layer while converting the pre-trained CNN structure to an FCN, but the differences are not essential. As a result, a single training of one network configuration will not influence the outcome of the optimization process.

### 5.1.2. Input

In this subsection, the results of the input combinations of Section 3.1.1 are discussed. The network configuration for this evaluation is shown in Table 5.2. Figure 5.3 shows the ROC curves of the input configurations.

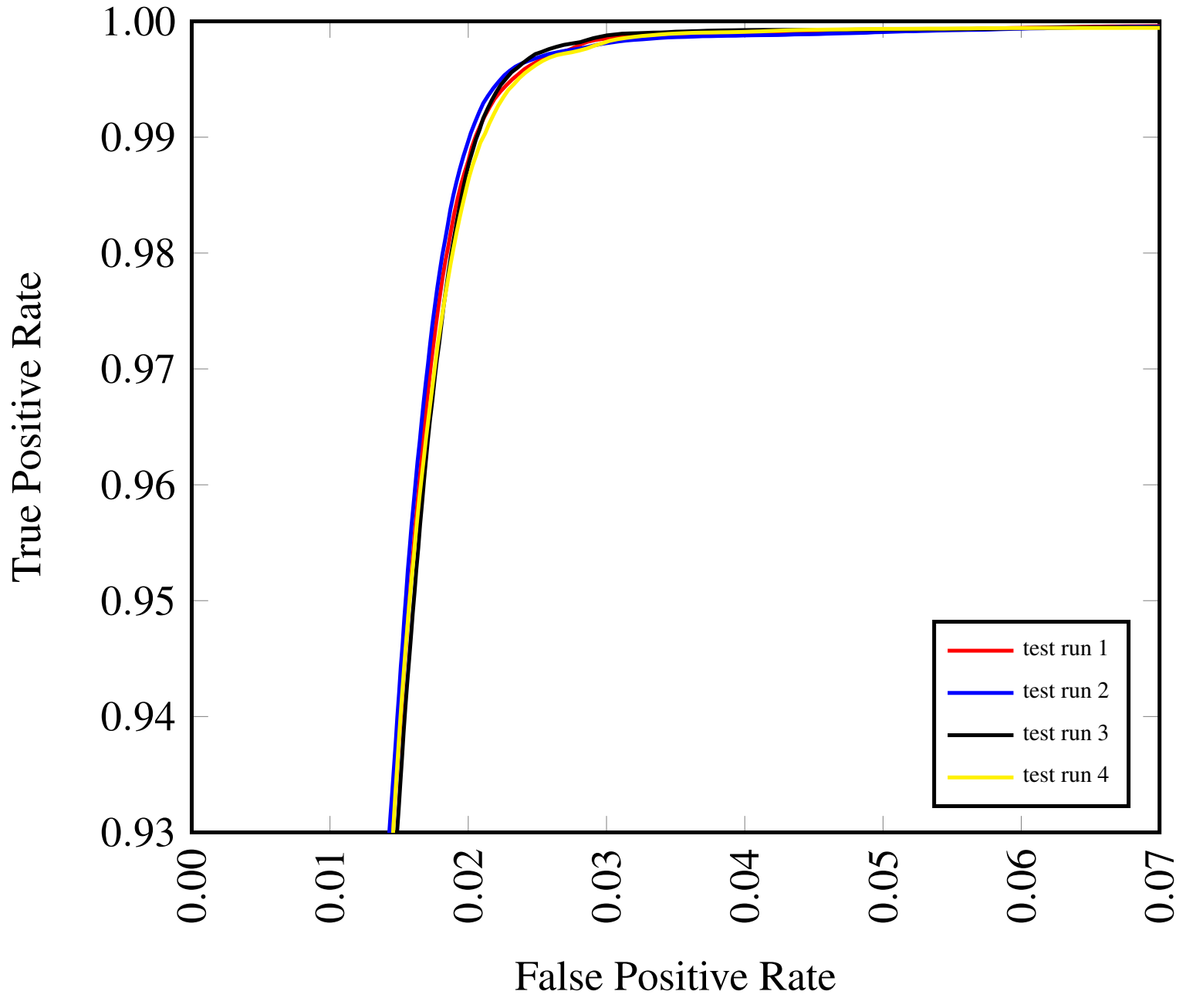

Figure 5.2.: ROC curves of multiple trainings of the same network configuration.

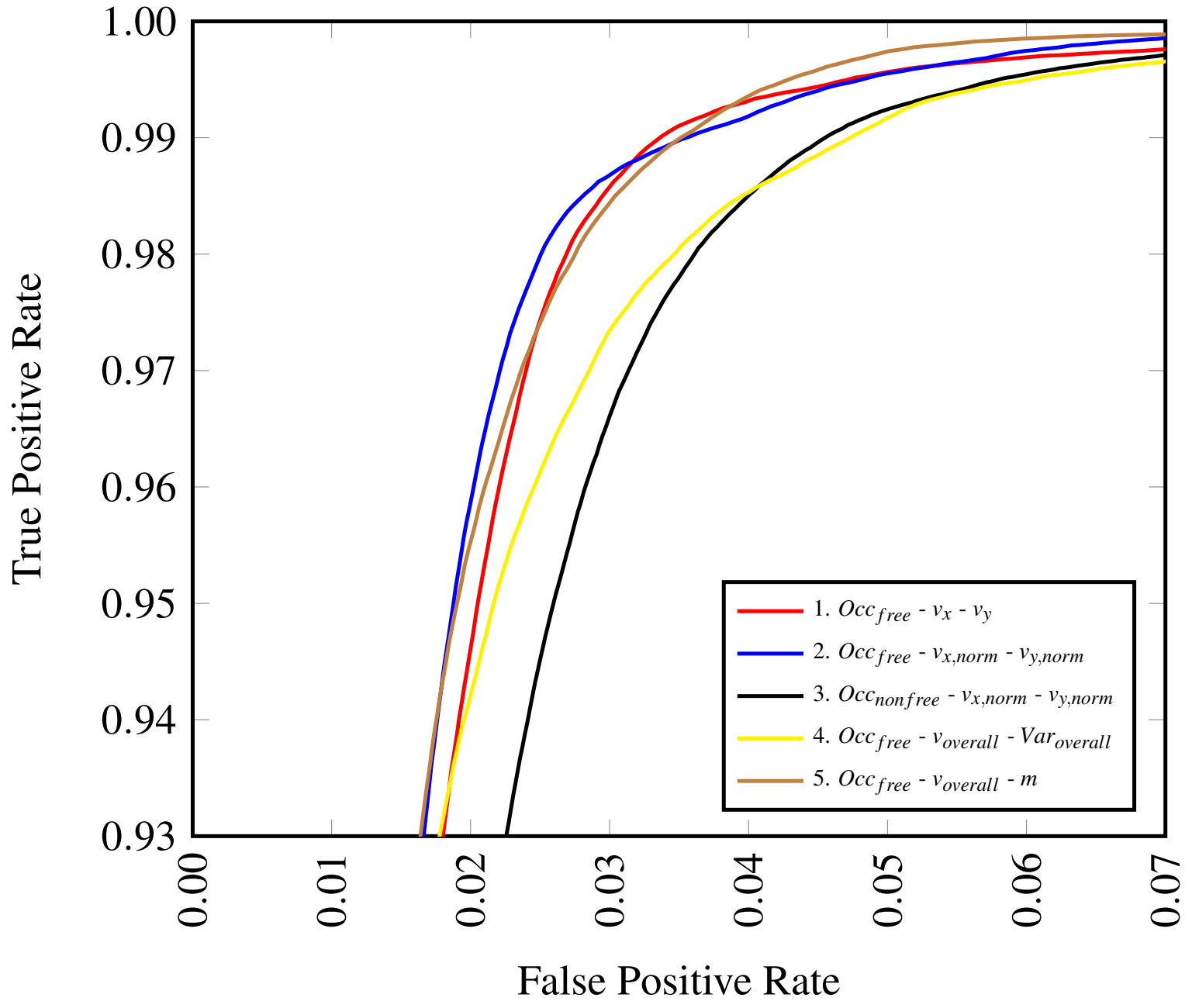

Figure 5.3.: ROC curves of different input configurations.

| Parameter | Value |
|---|---|
| Input | evaluated in this section |
| Input Range | $[-10, 10]$ |
| Network Structure | FCN-8s |
| Crop Size | $500 \times 500$ |
| Learning Rate | fixed policy with $2.14 \times 10^{-5}$ |
| Weight Factor | 60 |

Table 5.2.: Network configuration for optimization of the input.

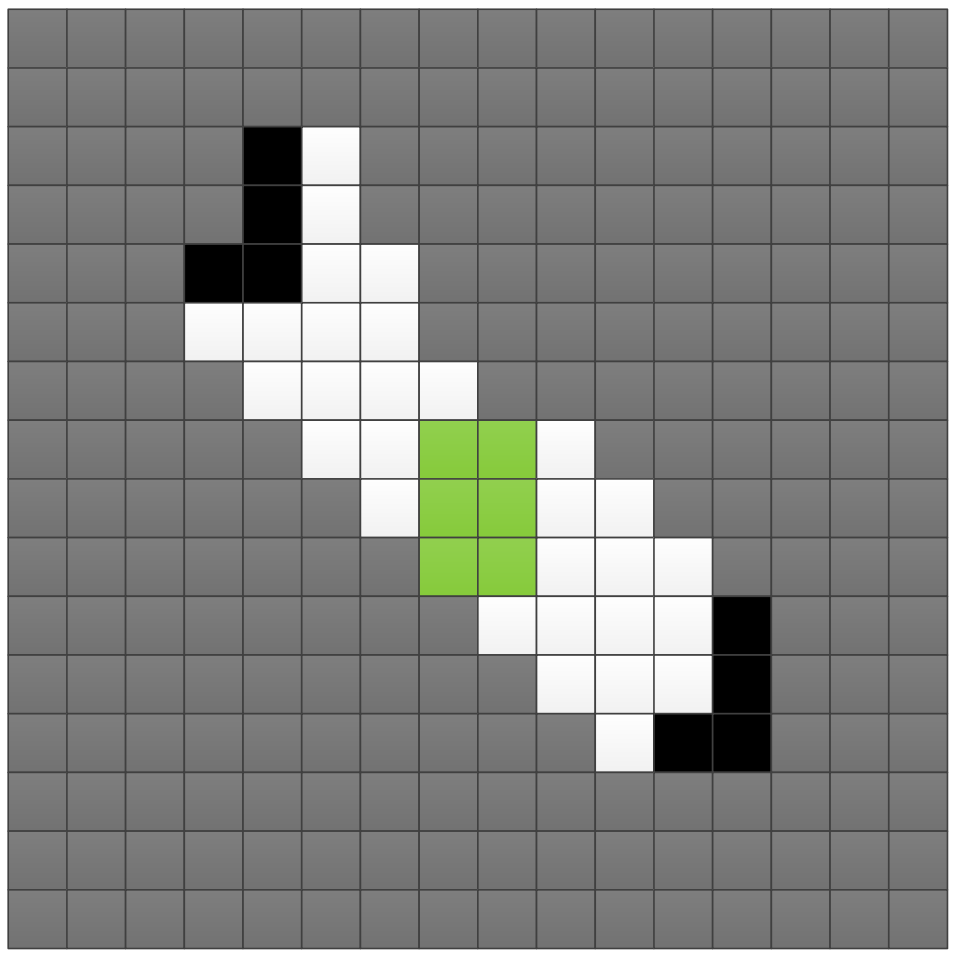

Figure 5.4.: Example of the recognition of a vehicle (upper left L-shape) and a wall (lower right L-shape) within a DOG by using the freespace information: the green object in the middle is the recording vehicle.

In general, the combinations with separation of velocities in $x$ and $y$ direction (combination 1 + 2) and the combination with the mahalanobis distance (combination 5) reach a better result than the combination with the separate overall variance (combination 4) and the combination without the freespace information (combination 3). That proves on the one side that the split of the normalized velocity of different direction into the overall velocity and the overall variance creates no information gain for the CNN. On the other side it proves, that the freespace information is an elementary information gain for recognizing moving obstacles. This can be explained with an example showed in Figure 5.4. There it is easy to distinguish between a corner of a wall, where at the inside of the L-shape is freespace information, and a possible vehicle. Without the freespace information, the two objects have the same appearance.

| Parameter | Value |
|---|---|
| Input | $Occ_{free}$ - $v_{x,norm}$ - $v_{y,norm}$ |
| Input Range | evaluated in this section |
| Network Structure | FCN-8s |
| Crop Size | $500 \times 500$ |
| Learning Rate | fixed policy with $2.14 \times 10^{-5}$ |
| Weight Factor | 60 |

Table 5.3.: Network configuration for optimization of the range.

Some smaller differences can be seen between the combination with the mahalanobis distance (combination 5) and the normalized and unnormalized approaches (combination 1 + 2). There, the mahalanobis distance has a slightly lower performance than the other two approaches. A suggestion for the reason is the fusion of the velocity in $x$ and in $y$ direction to the overall velocity. There the information of the moving direction is lost. So some specific motion model cannot be calculated. For example a vehicle can only drive forward and backward. This shows that an object with a vehicle appearance and with a motion to the side is probably clutter and should be labeled as static. This cannot be recognized without a moving direction.

The combination with the normalized velocities in $x$ and $y$ direction (combination 2) creates the best result. It is slightly better than the combination with the unnormalized velocities. That can be explained with the appearance of clutter. It can be observed, that clutter can reach a high velocity but has also a high variance. If the velocities are normalized, clutter can easier be detected as static.

As a result of the input configuration, the combination with the normalized velocity in $x$ and $y$ direction and with freespace information (combination 2) will be used for the final comparison.

### 5.1.3. Range

In this subsection the range configurations, defined in Section 3.1.2, are evaluated. The configuration of the network is shown in Table 5.3. The results of the ranges are presented in Figure 5.5 as ROC curves. There, the performance decreases with a larger range. This can be caused by the discretization of the input values while transferring the input values to discrete pixel values in the range of $[0,255]$. By increasing the range of the input values, the discretization error becomes greater, what could decrease the accuracy of the pixelwise labeling.

The only range, which does not match to the observed result that the performance decreases by increasing the range size, is the range $[-20,20]$. This range is slightly better than the smallest range. To verify, if there is a correlation between the range and the velocity distribution, Figure 5.6 shows the normalized velocity distribution in $x$ direction (driving direction of the recording vehicle) of dynamic grid cells (green) and static grid cells (red). The distribution of static grid cells can be described with a Gaussian curve. This is potentially caused by the particle filter of the DOG, where the process noise (see Equation 2.5) and the measurement noise influence the velocity distribution. The distribution of the dynamic grid cells has a specific appearance: above zero the distribution is relatively low and equally whereas below zero a distribution similar to a Gaussian curve with a negative mean can be recognized. This should be caused by the appearance of the obstacles in street environments. Vehicles that are following the recording vehicle can be recognized with different velocities, depending on the velocity of the recording vehicle. That is the reason why the velocity is relatively equal distributed in the positive $x$ direction. Obstacles that are driving towards the recording vehicle appear most of the time on the oncoming traffic lane. That causes the nearly same velocity for these vehicles and produces the Gaussian distribution in the negative $x$ direction.

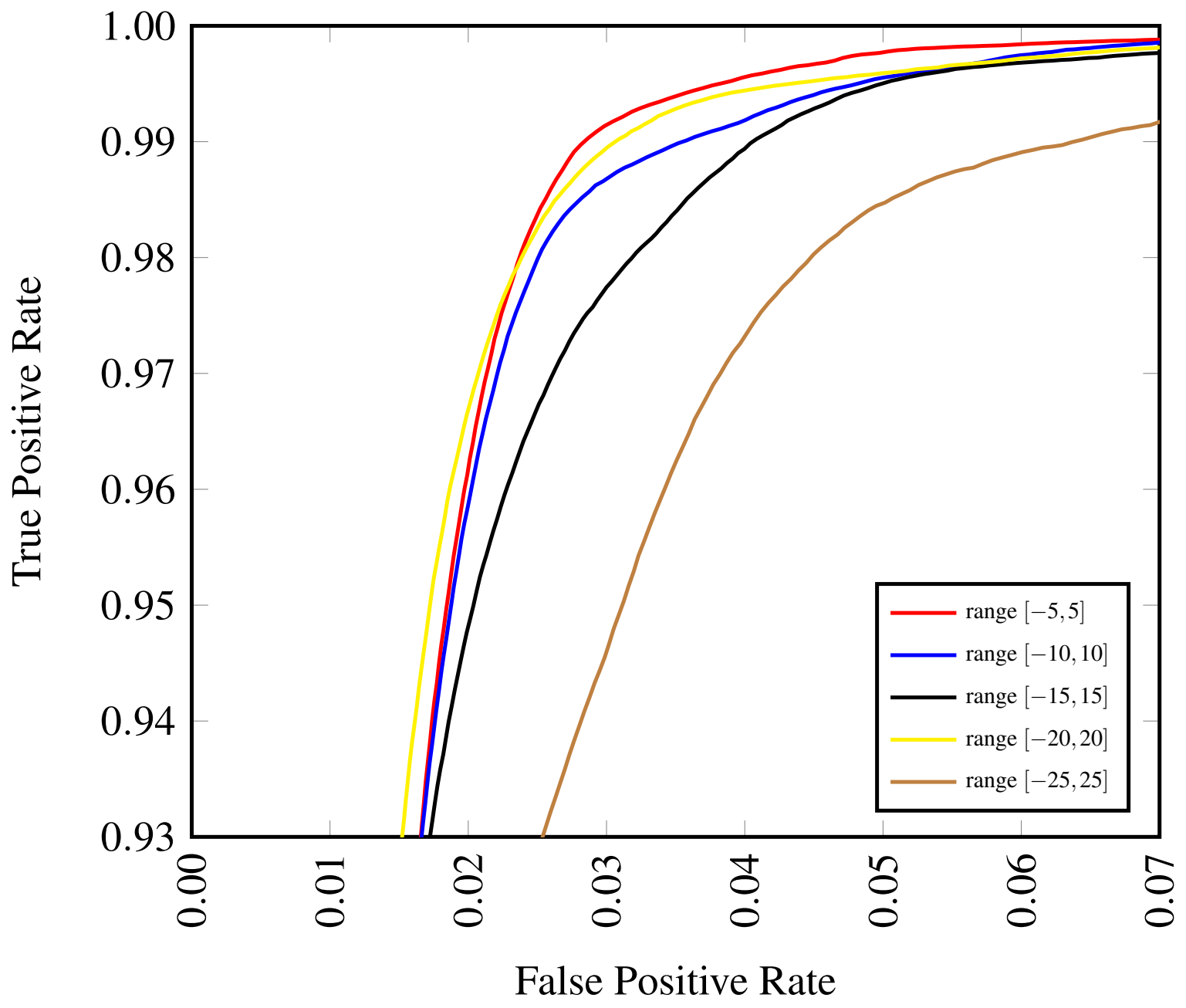

Figure 5.5.: ROC curves of different range configurations.

The distributions of the normalized velocity in $y$ direction are shown in Figure 5.7. There, the normalized velocities are all around zero, what should be also caused by the appearance of obstacles in the recorded street environment. Vehicles in crossing direction are only recognized on intersections, where on the recorded route only lower velocities were recognized.

The normalized velocity histograms do not show a relation to the range $[-20,20]$. So the cause of this unexpected result is not a specific distribution of the normalized velocities. One reason for this appearance could be the usage of the pre-trained networks. The pre-trained networks are trained on public datasets with a different type of input data. As a result, the weights of the network are adapted to a specific appearance of the input, which does not necessarily match the appearance of the input data within this thesis. By transferring the velocity to a certain range, the appearance of the input is changed. This can cause different and unexpected results.

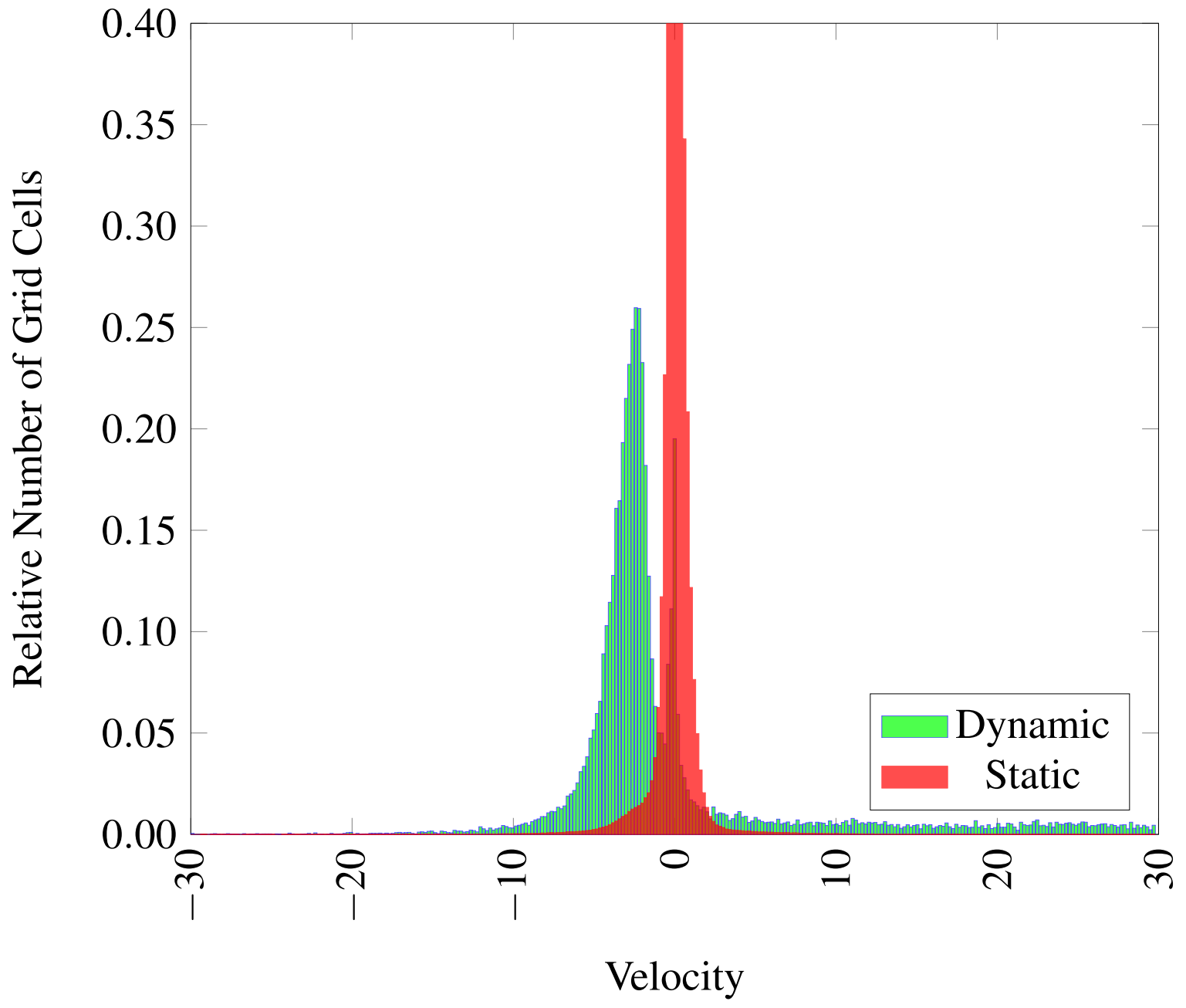

Figure 5.6.: Histogram of the normalized velocities in $x$ direction of the occupied grid cells within the validation dataset.

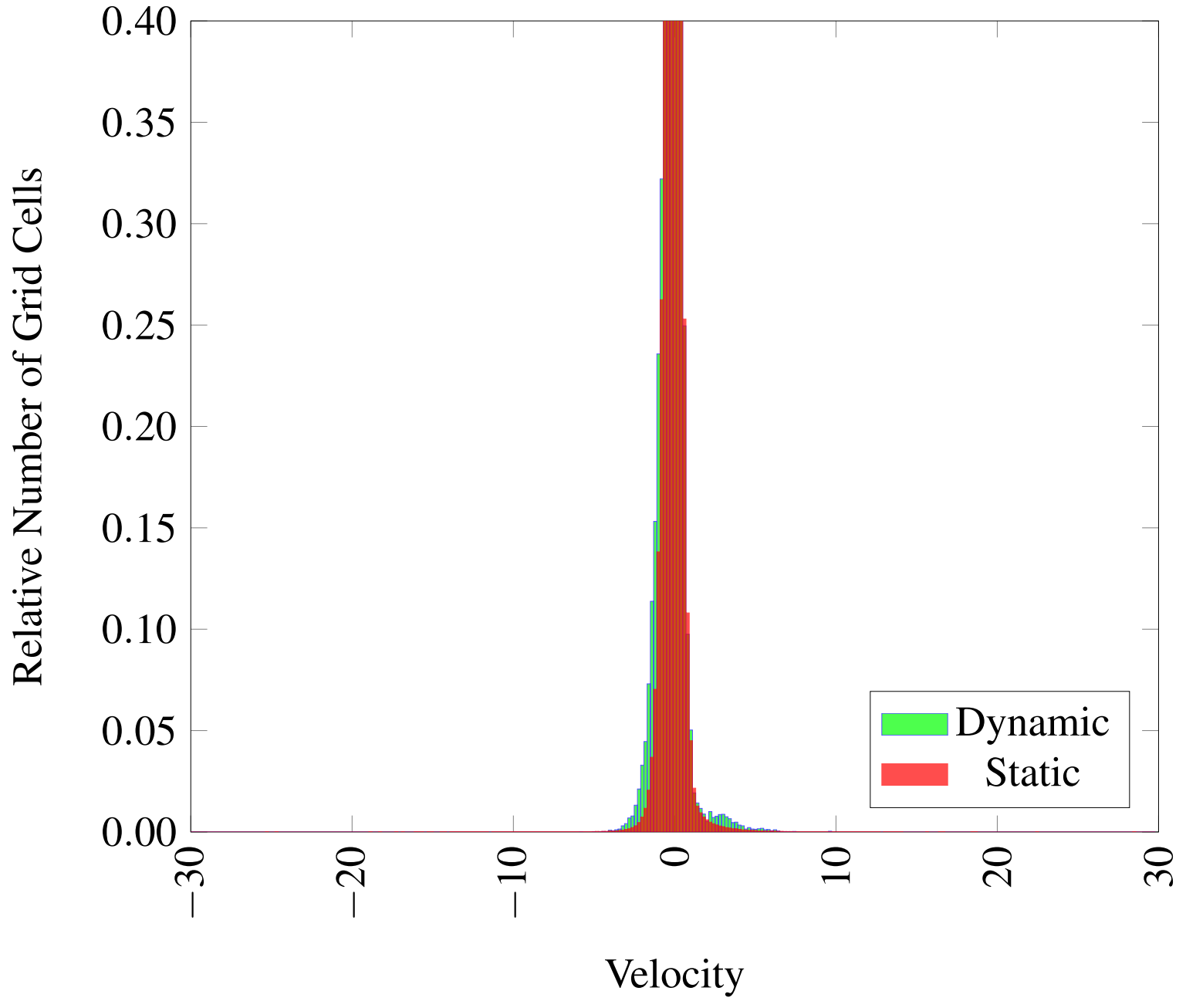

Figure 5.7.: Histogram of the normalized velocities in $y$ direction of the occupied grid cells within the validation dataset.

| Parameter | Value |
|---|---|
| Input | $Occ_{free}$ - $v_{x,norm}$ - $v_{y,norm}$ |
| Input Range | $[-10, 10]$ |
| Network Structure | evaluated in this section |
| Crop Size | $500 \times 500$ |
| Learning Rate | fixed policy with $2.14 \times 10^{-5}$(adapted for incremental trainings) |
| Weight Factor | 60 |

Table 5.4.: Network configuration for optimization of the structure.

As a result of the range configuration, the range $[-20, 20]$ is slightly better than the range $[-5, 5]$, what was also recognized in additional evaluations. Thus, the range $[-20, 20]$ will be used for the final comparison.

### 5.1.4. Structure

In this subsection, the structure configuration of Section 3.1.3 is evaluated. The configuration of the network is defined in Table 5.4. The first ROC curves in Figure 5.8 state the results of the incremental training of the FCN-16s. All the three network configurations are initialized with the pre-trained network. The FCN-16s non-incremental version means that the FCN-16s is trained directly from the pre-trained network. For the incremental versions the FCN-32s is trained first and afterwards the weights are transferred to the FCN-16s to continue the training. The two versions of the incremental training use different learning rates. Long et al. [28] recommend decreasing the learning rate in case of the incremental training. This is achieved with a decrease of the learning rate by factor ten. The other approach is implemented with a constant learning rate while applying the incremental training.

The incremental training for the FCN-16s shows that the incremental version with the recommendation of Long et al. is slightly better than the non-incremental training. The unchanged learning rate has the lowest performance in this comparison. The higher learning rate seems to deteriorate the learned model of the FCN-32s.

In Figure 5.9, the ROC curves of the FCN-8s are shown. There, the opposite can be seen. The incremental training with an unchanged learning rate gives the best result, while the incremental training with the decreased learning rate creates a lower performance. The reason for that seems to be the refinement with the deep jet, which seems to need more adjustment at the FCN-8s compared to the FCN-16s.

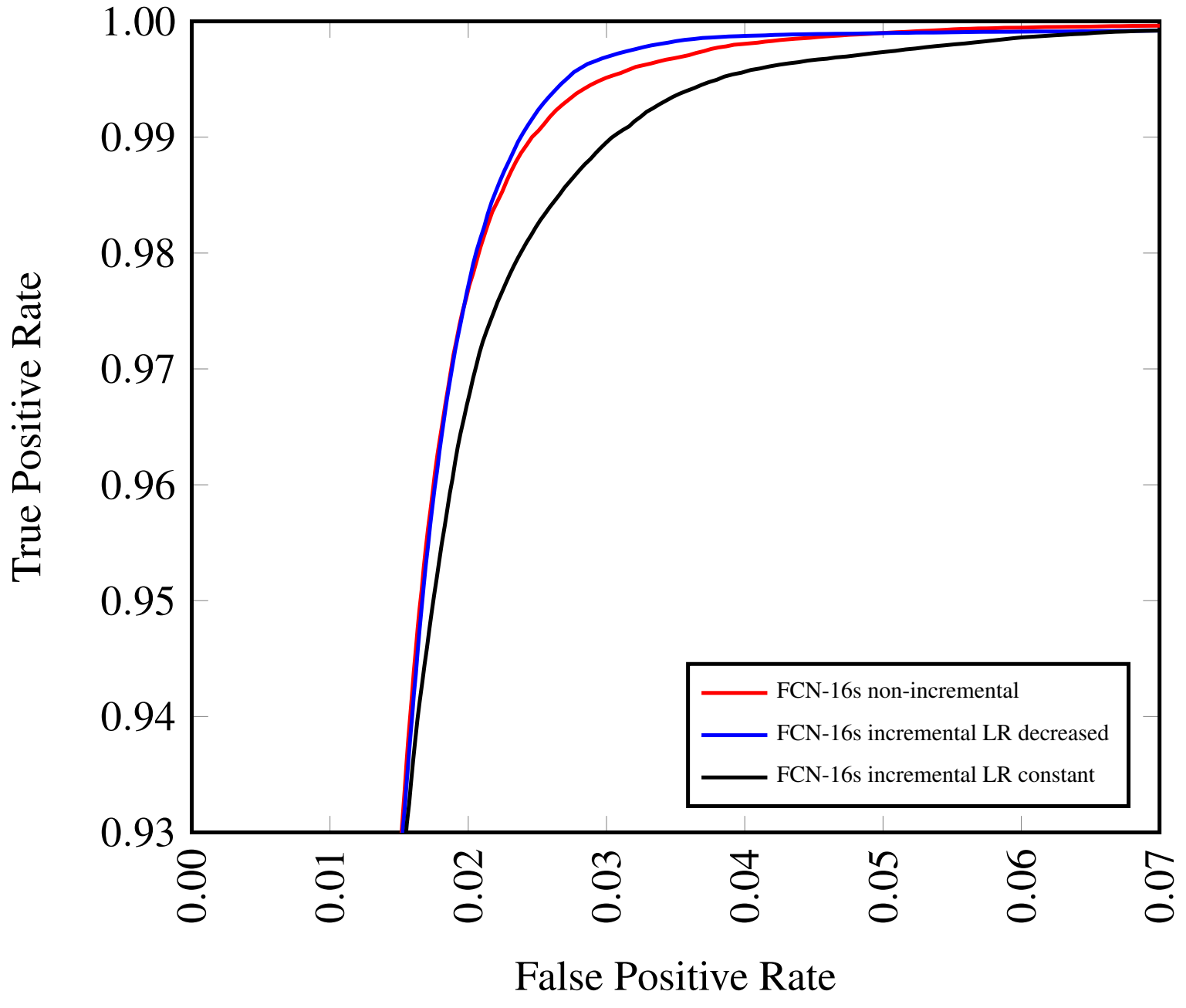

Figure 5.8.: ROC curves of different structure configurations for the incremental training of the FCN-16s.

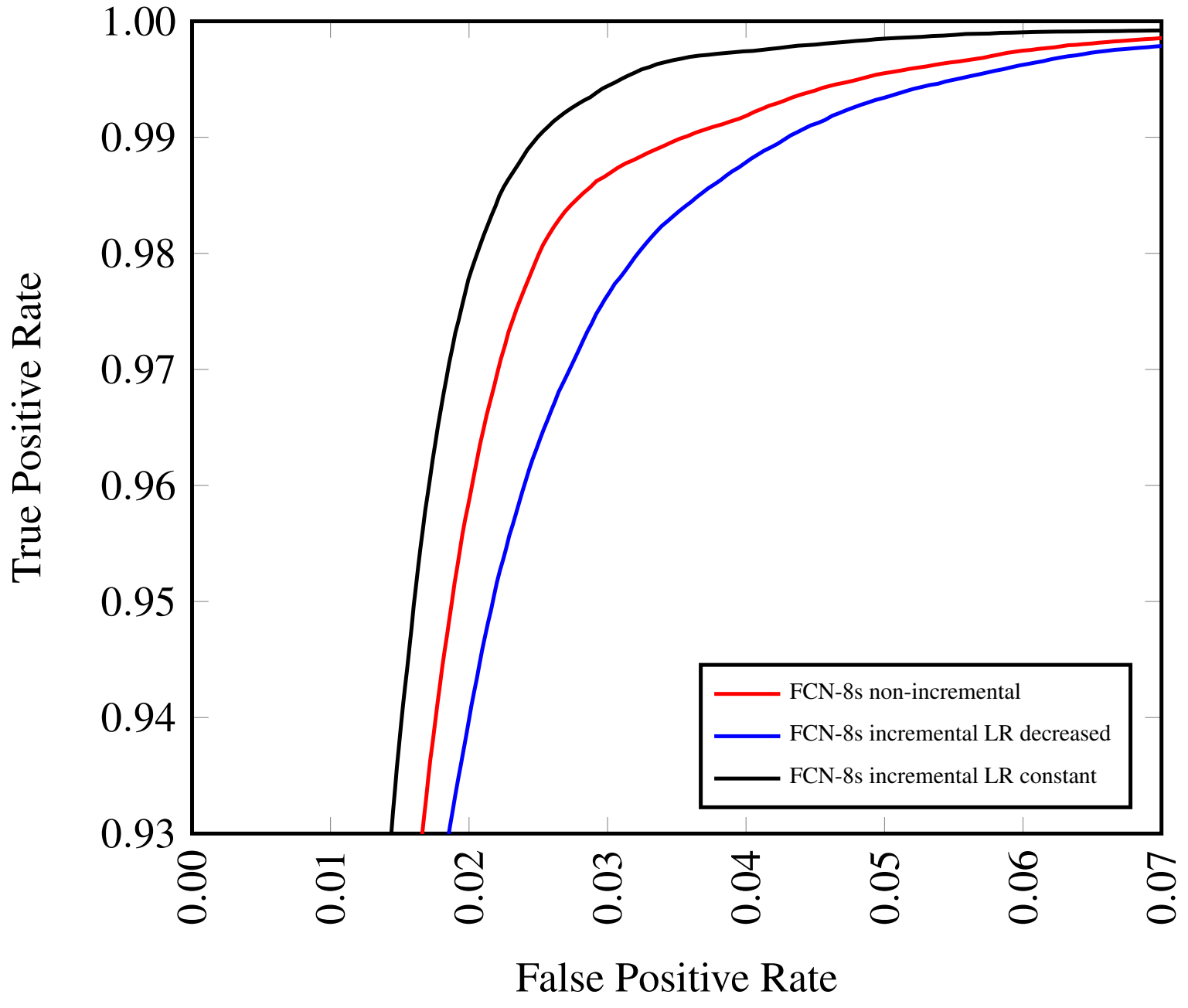

Figure 5.9.: ROC curves of different structure configurations for the incremental training of the FCN-8s.

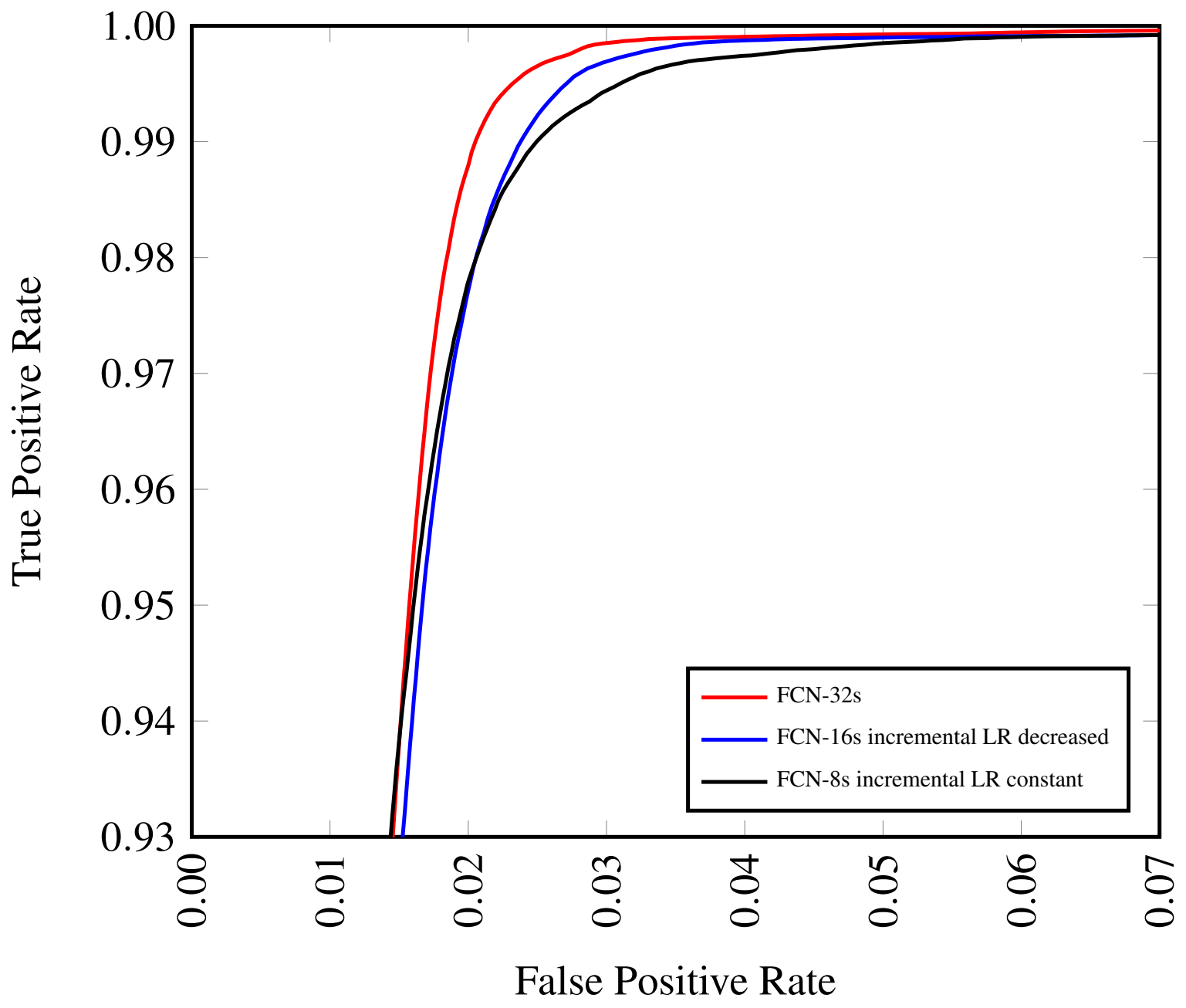

Figure 5.10.: Best ROC curves of different incremental trainings evaluated with the FCN-32s.

In comparison to the FCN-32s, the best incremental versions are shown in Figure 5.10. There, the FCN-32s is the best network configuration. This is an interesting result, because Long et al. [28] proved the opposite. They describe that the FCN-8s version creates finer details within the segmentation, what creates a better result. In general the finer details can be proved in this thesis as well, what is shown in Figure 5.11. There, the segmentation output of the FCN-32s, FCN-16s, FCN-8s, and the input data is provided. The FCN-32s created much coarser results than the FCN-8s, but it creates also less false positive clusters. Additionally, the filter sizes of the last deconvolutional layer can be recognized at the marked vehicle (yellow ellipse). There, the up-sampling artifacts can be recognized, which becomes smaller with a smaller stride as with the FCN-16s or the FCN-8s. The output of the CNN is not used directly for segmentation in this thesis. As mentioned in Section 3.1, the output of the CNN is refined with the occupancy information of the DOG. As a result, the FCN-32s generate a better ROC curve than the FCN-16s or FCN-8s caused by less false positive. That means the refinement with the occupancy information of the DOG is better than the refinement with the deep jet.

As mentioned in Section 3.1.3, the resulting quality should also be evaluated against the execution time. For this purpose, the smaller Alexnet is also evaluated in this context. The results are shown as ROC curves in Figure 5.12. The ROC curves show that the Alexnet in general has a lower performance than the FCN-Xs networks. This is as expected caused by the smaller structure of the network. It can also be recognized in Figure 5.13, where the outputs of the Alexnets are provided. There, the Alexnet creates a coarser segmentation as the FCN-Xs in Figure 5.11, especially at the stride 32 version the Alexnet combines two vehicles to one segment. This coarser segmentation is afterwards also refined with the occupancy information of the DOG. That results in a better output of the Alex-32s related to the Alex-4s as well as the FCN-32s related to the FCN-8s.

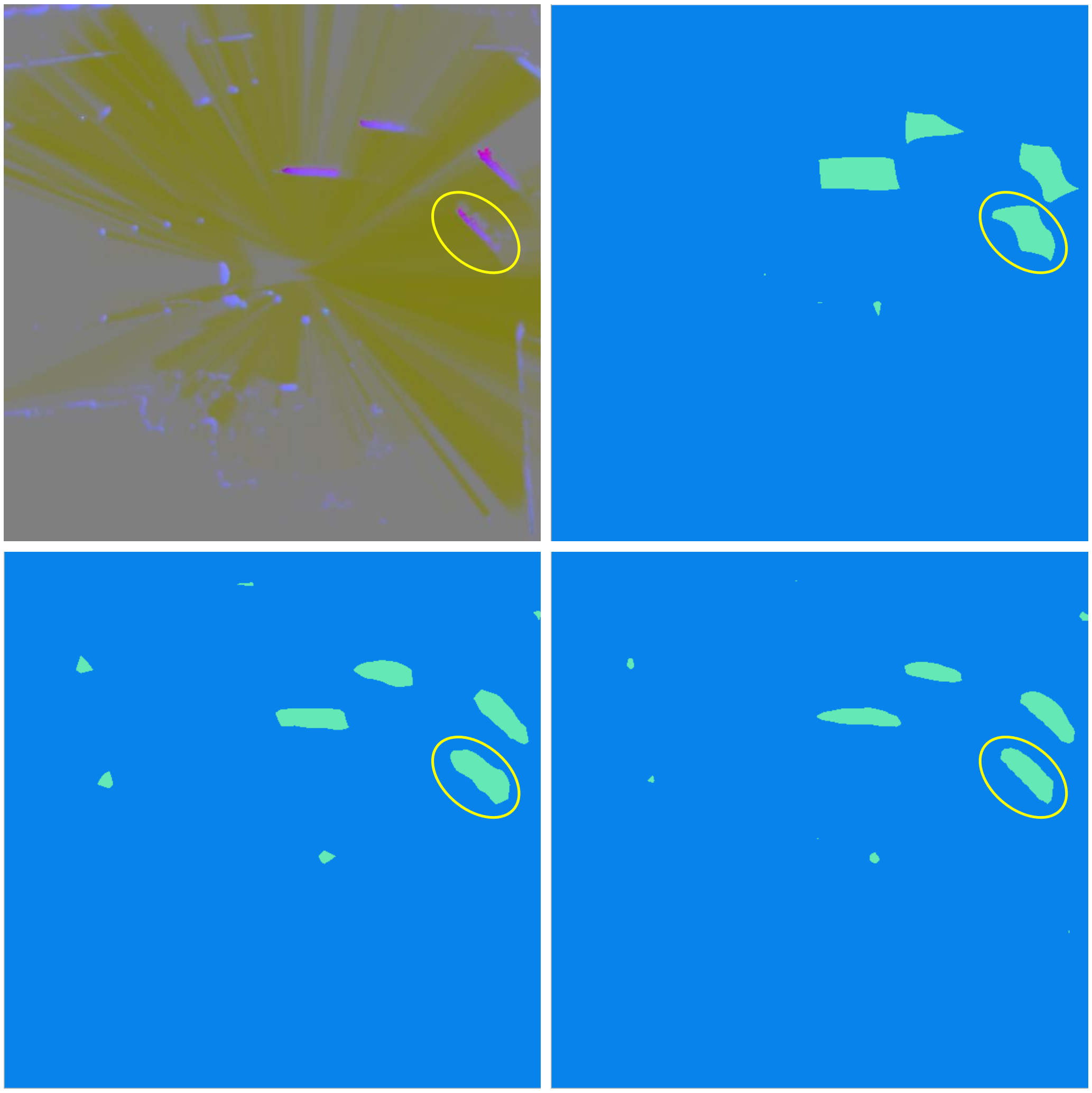

Figure 5.11.: Example of the result of deep jet of VGG-net: top left: input image, top right: output of the FCN-32s, bottom left: output of the FCN-16s, bottom right: output of the FCN-8s. In general, the smaller stride at the last layers (for example at FCN-8s or FCN-16s) produces finer details of objects, but increases also the number of false positive clusters.

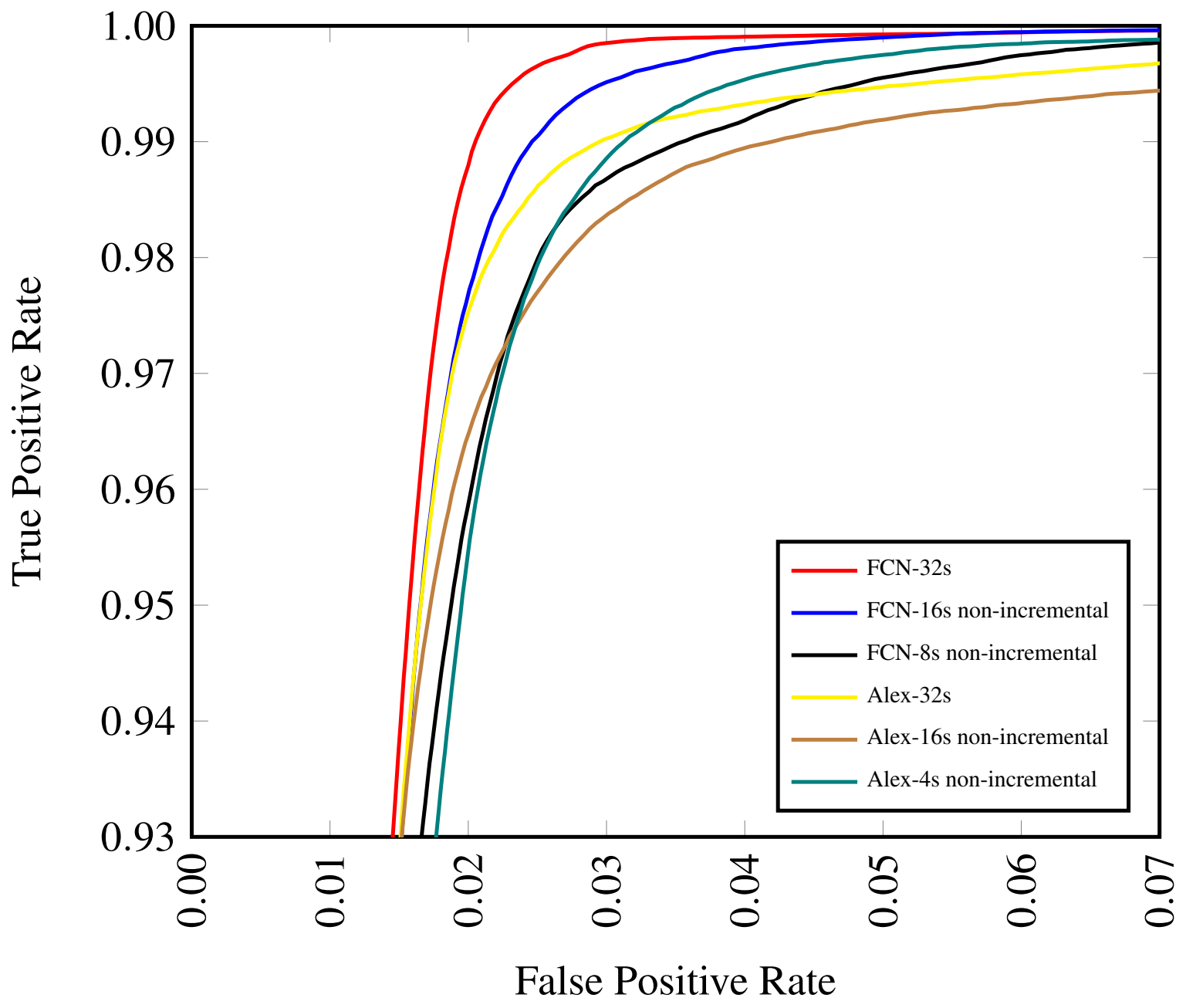

Figure 5.12.: ROC curves for comparison of the Alexnet and the VGG-net.

| Network Structure | Preprocessing and Postprocessing | Network Execution Time |
|---|---|---|
| FCN-32s | $\approx 100ms$ | $\approx 160ms$ |
| Alex-32s | | $\approx 20ms$ |

Table 5.5.: Execution time of different network structures.

The execution time of the Alexnet is much smaller, what can be seen in Table 5.5. There, the execution time is divided into two parts. The first part is the pre-processing and post-processing time, which is needed to create the specific input image for the CNN (see Section 3.1.1) and to refine the output with the occupancy information of the DOG. This time can be reduced by parallelizing the execution. The second part of the execution time is the processing of the CNN itself, which is already highly optimized on the GPU. The shorter execution time has to be seen in contrast to the quality of the output. As a result the Alexnet can be used for real-time execution, especially because the output is still good, but for further optimization in this thesis, the FCNs based on the VGG-net will be used. As a result of the structure configuration, the FCN-32s will be used for the final comparison.

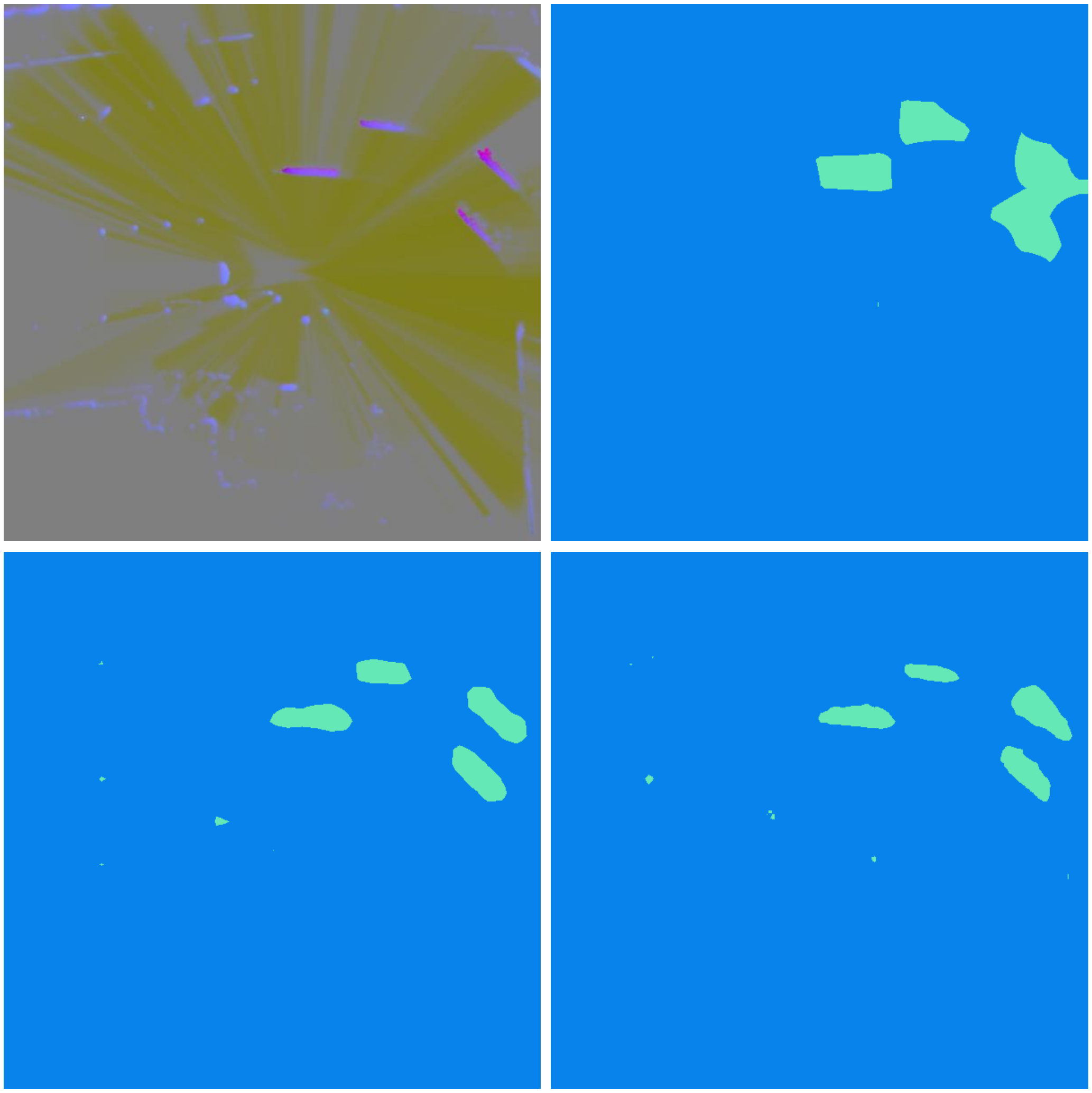

Figure 5.13.: Example of the result of deep jet of Alexnet: top left: input image, top right: output of the Alex-32s, bottom left: output of the Alex-16s, bottom right: output of the Alex-4s. The smaller stride at the last layers (at Alex-4s) produces finer details of objects, but increases also the number of false positive clusters.

| Parameter | Value |
|---|---|
| Input | $Occ_{free}$ - $v_{x,norm}$ - $v_{y,norm}$ |
| Input Range | $[-10, 10]$ |
| Network Structure | FCN-8s |
| Crop Size | evaluated in this section |
| Learning Rate | fixed policy with $2.14 \times 10^{-5}$ |
| Weight Factor | 60 |

Table 5.6.: Network configuration for optimization of the zooming.

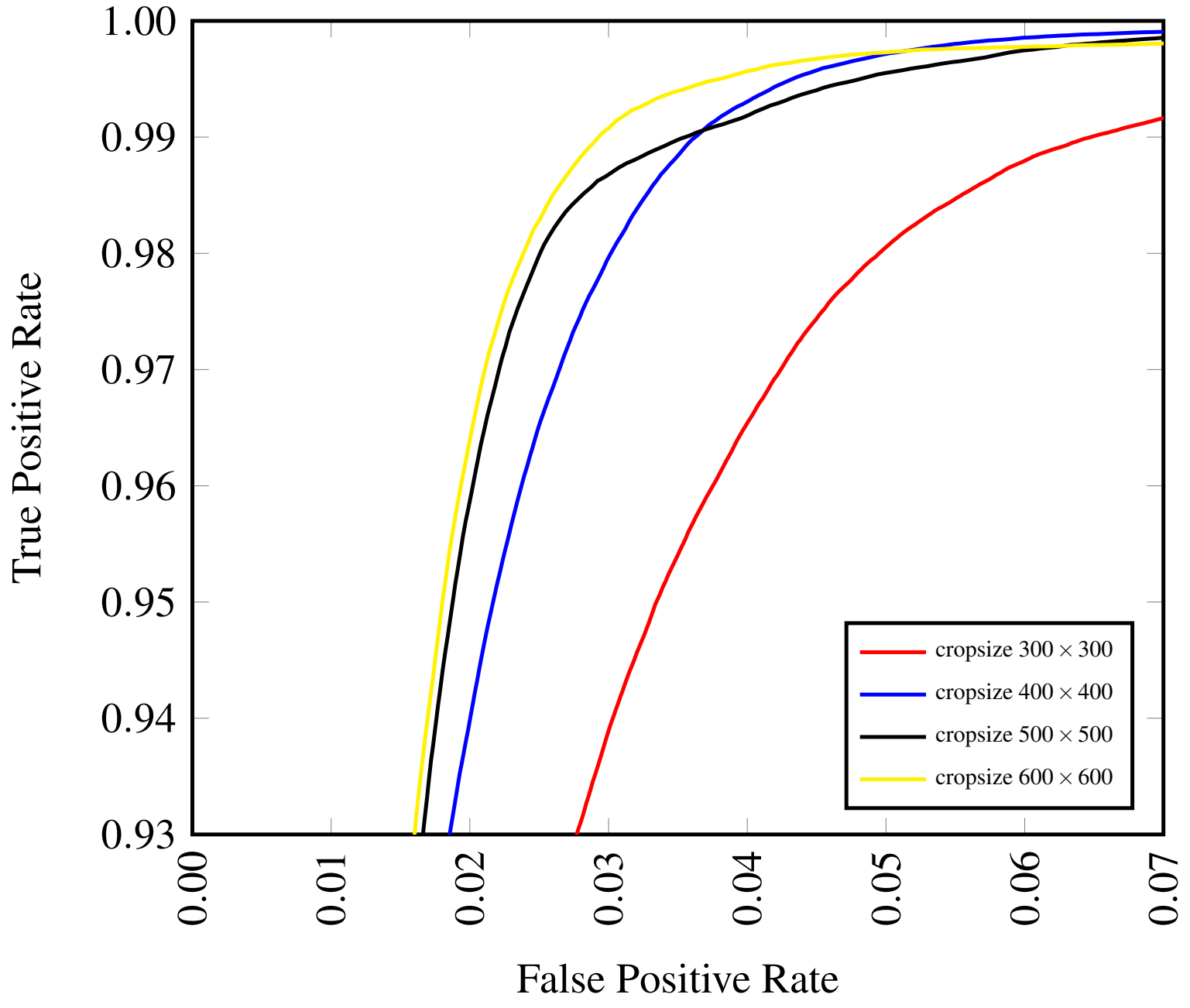

Figure 5.14.: ROC curves of different zooming configurations.

### 5.1.5. Zooming into the input

In this subsection the zooming configuration of the input of Section 3.1.4 is evaluated. The configuration of the network is given in Table 5.6. The result of the zooming is shown in Figure 5.14, where zooming into the input images creates a lower performance. This seems to be caused by the size of the object. The goal of the zooming was to enlarge the moving objects for a better recognition, but at the same time clutter and other objects are zoomed. These objects provoke false positives, what is shown in Figure 5.15. There, an input image is calculated with the different crop sizes. In the output image without zooming (crop size $600 \times 600$), no clutter is labeled as dynamic. By increasing the zooming, more clutter is labeled, what is marked with red ellipses. This signifies that the zooming into the images to increase the size of the objects does not improve the segmentation task.

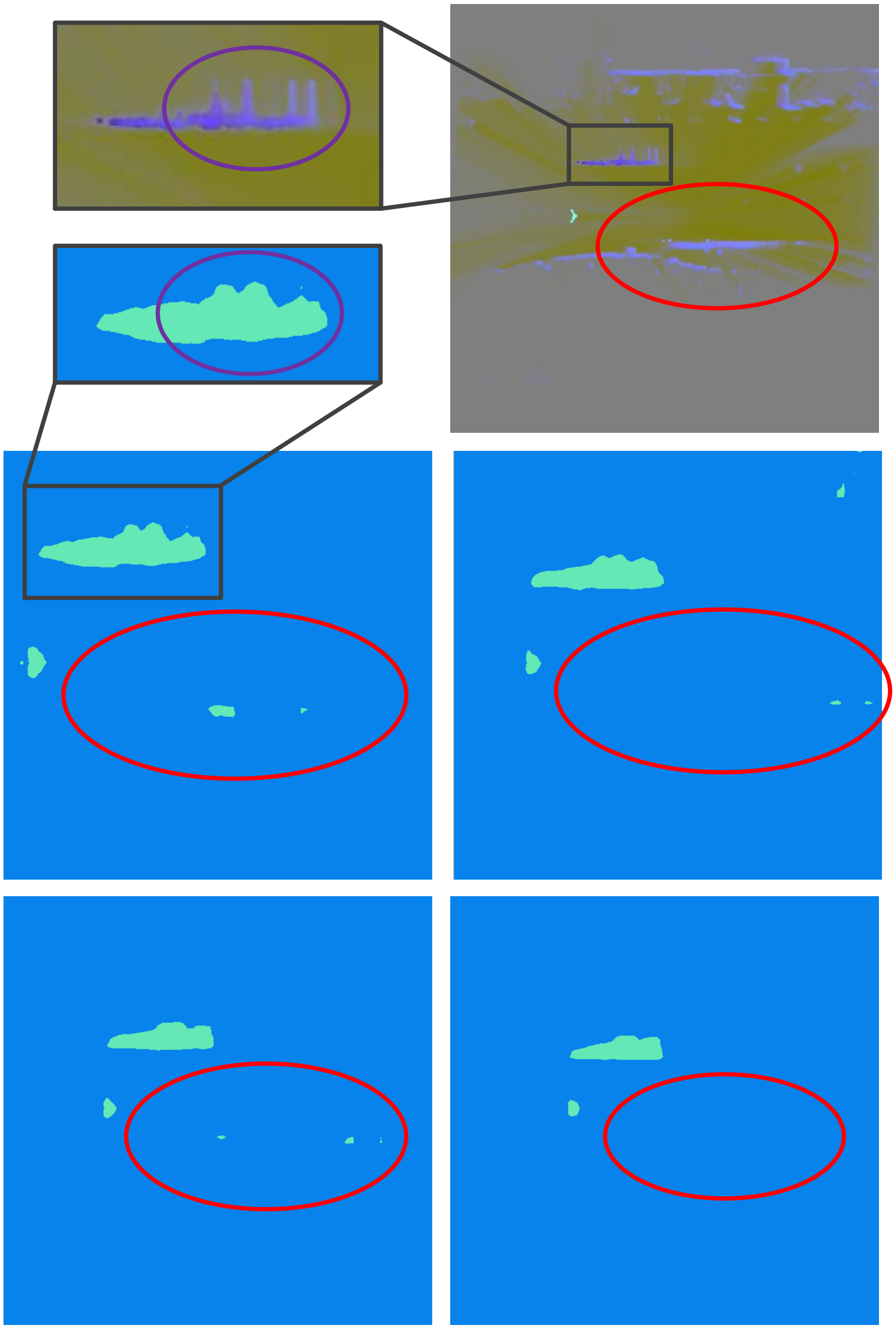

Figure 5.15.: Example of different crop sizes for one input image: top left: zoom of the input image and the detection of a vehicle with a crop size $300 \times 300$, top right: input image, middle left: crop size $300 \times 300$, middle right: crop size $400 \times 400$, bottom left: crop size $500 \times 500$, bottom right: crop size $400 \times 400$. With a smaller crop size, the number of false positive increases (red ellipse) and objects are recognized with more details, what can produce a partly recognition of an object if the velocity information differs over the object (violet ellipse).

| Parameter | Value |
|---|---|
| Input | $Occ_{free}$ - $v_{x,norm}$ - $v_{y,norm}$ |
| Input Range | $[-20, 20]$ |
| Network Structure | FCN-32s |
| Crop Size | $600 \times 600$ |
| Learning Rate | evaluated in this section |
| Weight Factor | 60 |

Table 5.7.: Network configuration for optimization of the learning rate.

Additionally the large objects receive more details by increasing the zooming, what can be seen at the borders of the recognized vehicle in Figure 5.15. This can also decrease the accuracy of the result, because the details become so strong, that the vehicle is not recognized as one obstacle. As a result, moving parts of the vehicles are recognized by itself, what can be seen at the crop size $300 \times 300$ (violet ellipse). There, the right part of the vehicle has lower velocity information as the left part of the vehicle (light blue instead of dark blue) within the DOG. This produces a different appearance of the output labels. In general it is important to choose the right level of details to recognize the vehicle as one obstacle and not with too many detailed structures. As a result of the zooming configuration, the crop size $600 \times 600$ will be used for the final comparison.

### 5.1.6. Learning Rate

In this subsection, the learning rate configuration from Section 3.1.5 is evaluated. The configuration of the network is defined in Table 5.7. The results of the fixed learning rates are stated in Figure 5.16. There it can be recognized, that the proposed learning rate of Long et al., transferred to a normalized learning rate, creates the best results. All other learning rates produce a lower performance. That is caused by the network structure and the way how to learn. A CNN learns by adapting its weights of the neurons. For that reason a gradient descent algorithm is used and the CNN converges to a minimum. The goal is to converge to a global minimum and instead of local minima. To reach that, the learning rate should be on the one side high enough to create a global optimization and on the other side low enough to reach the minimum. That can be recognized in the results, where higher or lower learning rates produce less performances.

The problem while training with a constant learning rate is the conflict between the global optimization and the convergence to a local minimum. For that purpose a step learning rate is evaluated as described in Section 3.1.5. In Figure 5.17 and Figure 5.18 the two different learning developments are shown. There, the step learning rate converges faster and is more stable in the end of the training. This is caused by the difference of the learning approaches. On the one side the step learning rate starts with a ten times higher learning rate, what creates a faster conversion. On the other side the learning rate decreases to a ten times lower learning rate at the end of the training, what causes the stability of the training in the end.

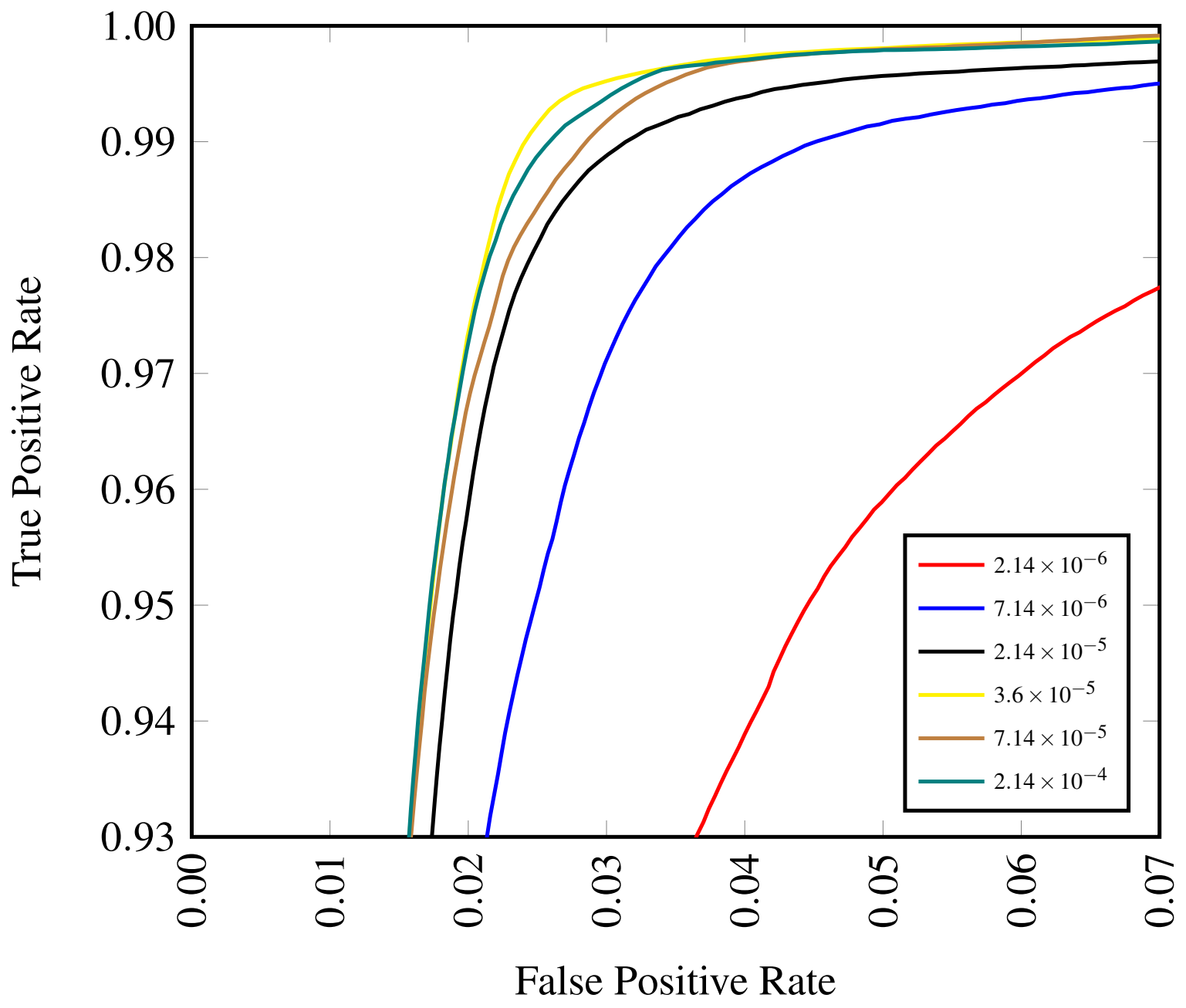

Figure 5.16.: ROC curves of different fixed learning rate configurations.

The step learning rate results are shown in Figure 5.19, where, equally to the fixed learning rate approach, the transferred learning rate of Long et al. generates the best result. The other approaches receive a lower performance in dependence of the distance to the optimal learning rate. The learning rate $\mu = 2.14 \times 10^{-3}$ is not shown in the figure, caused by numerical problems by using such a high learning rate.

To choose the best learning rate for the final comparison, the best learning rates of the two learning policies are evaluated against each other. That can be recognized in Figure 5.20. There, the step learning rate is better than the fixed learning rate approach. As a result the step learning rate of $\mu = 3.6 \times 10^{-4}$ will be used for the final comparison.

### 5.1.7. Weight Matrix

In this subsection, the weight matrix configuration of Section 3.1.6 is evaluated. The configuration of the network is defined in Table 5.8. To define the weight matrices for evaluation, the ratio between moving grid cells and non-moving grid cells of the validation dataset was calculated, which is nearly 1 : 200. For this reason the following factors $c^{(1)}$ for the moving grid cells were defined:

$$c^{(1)} \in \{1, 20, 40, 60, 80, 100, 120, 140, 160, 180, 200\} \quad . \tag{5.1}$$

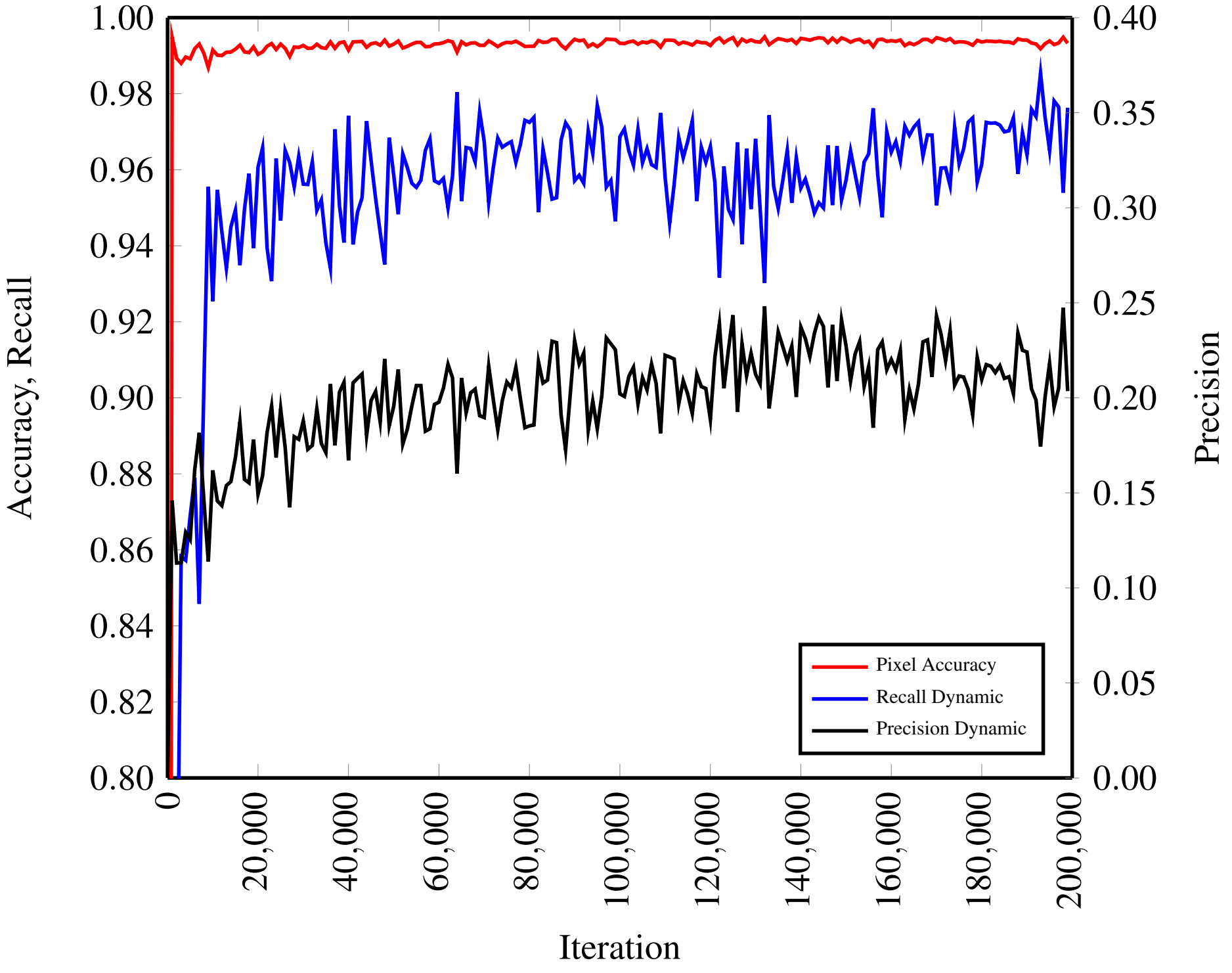

Figure 5.17.: Learning curve of the fixed learning rate $3.6 \times 10^{-5}$.

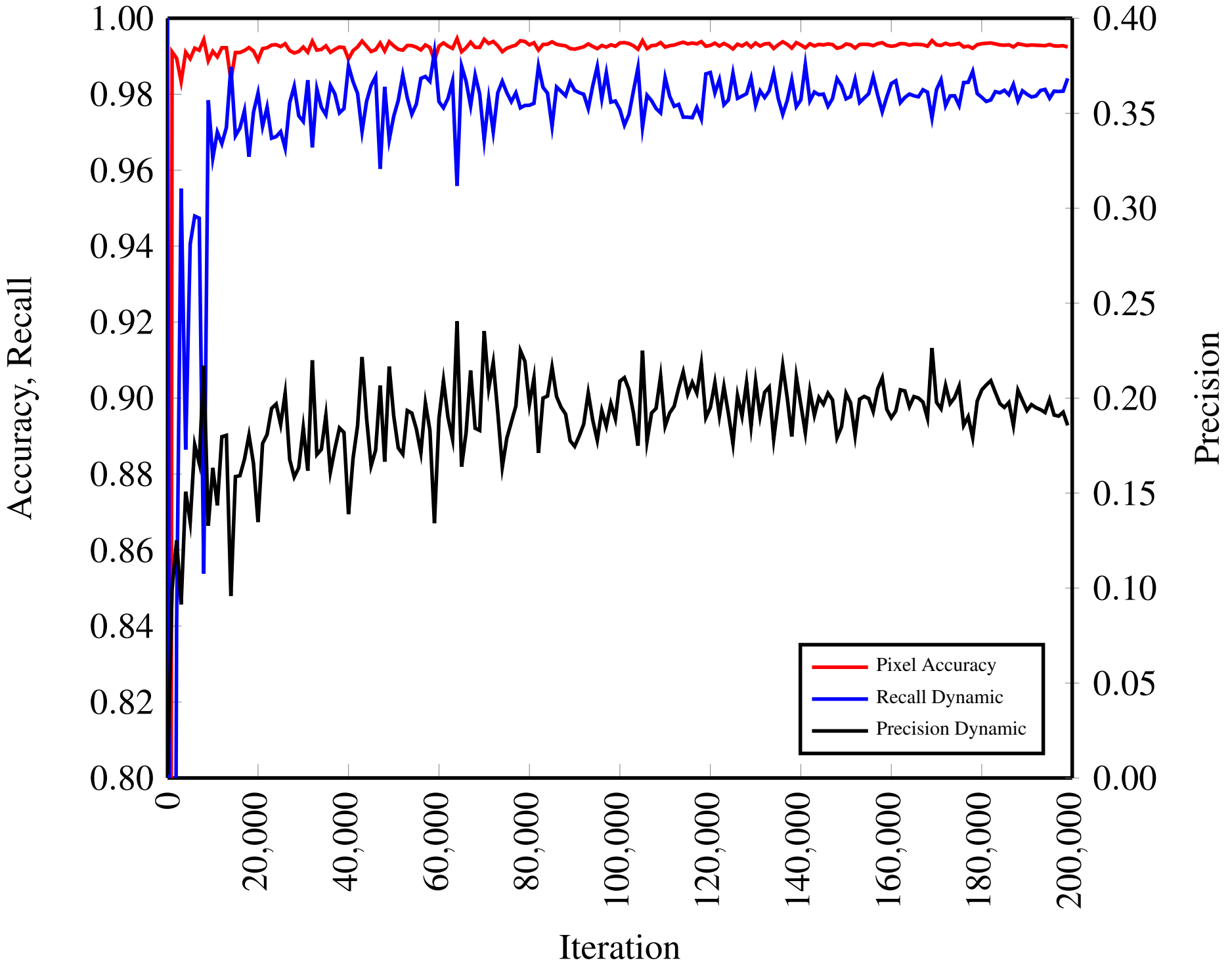

Figure 5.18.: Learning curve of the step learning rate $3.6 \times 10^{-4}$.

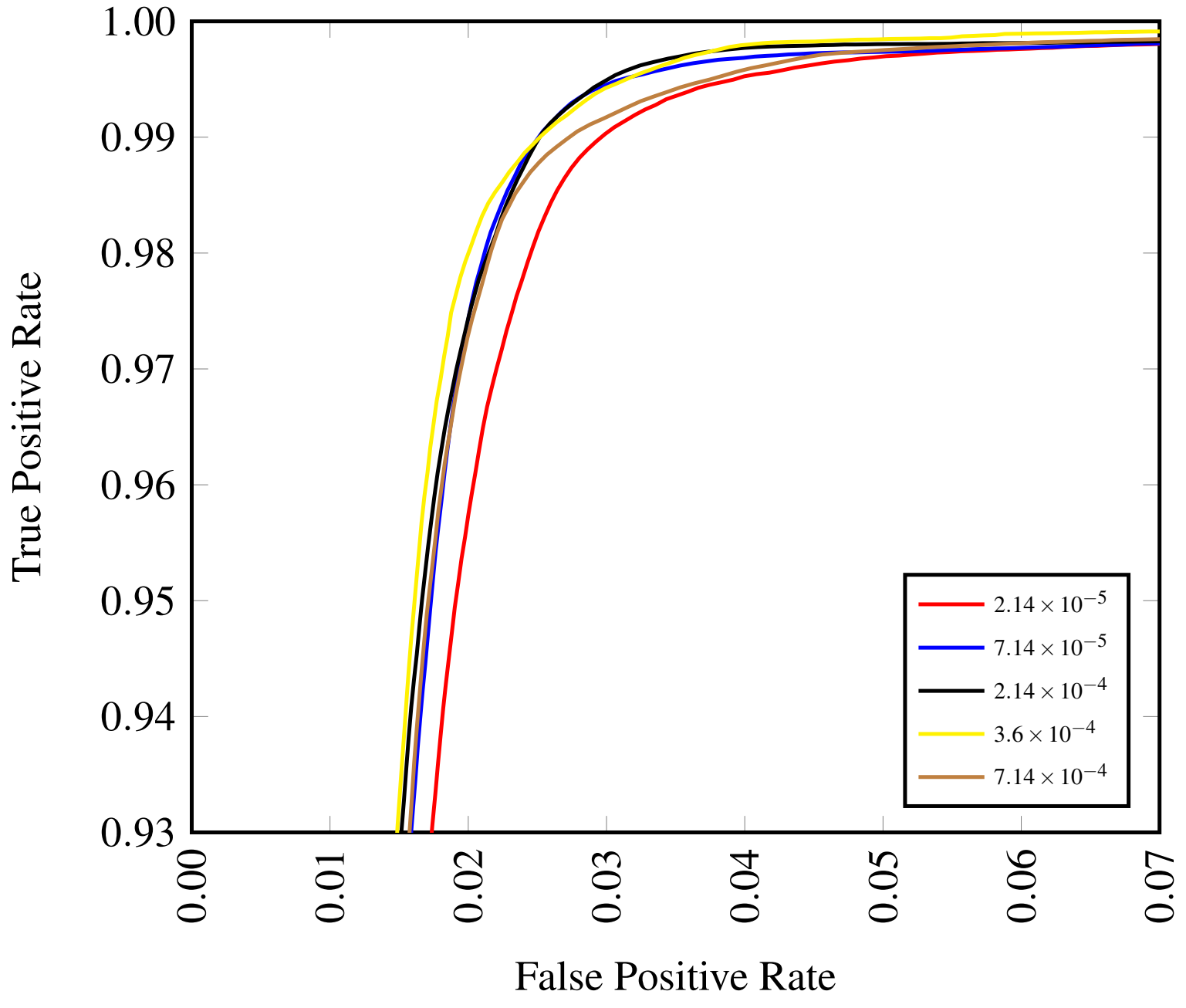

Figure 5.19.: ROC curves of different step learning rate configurations.

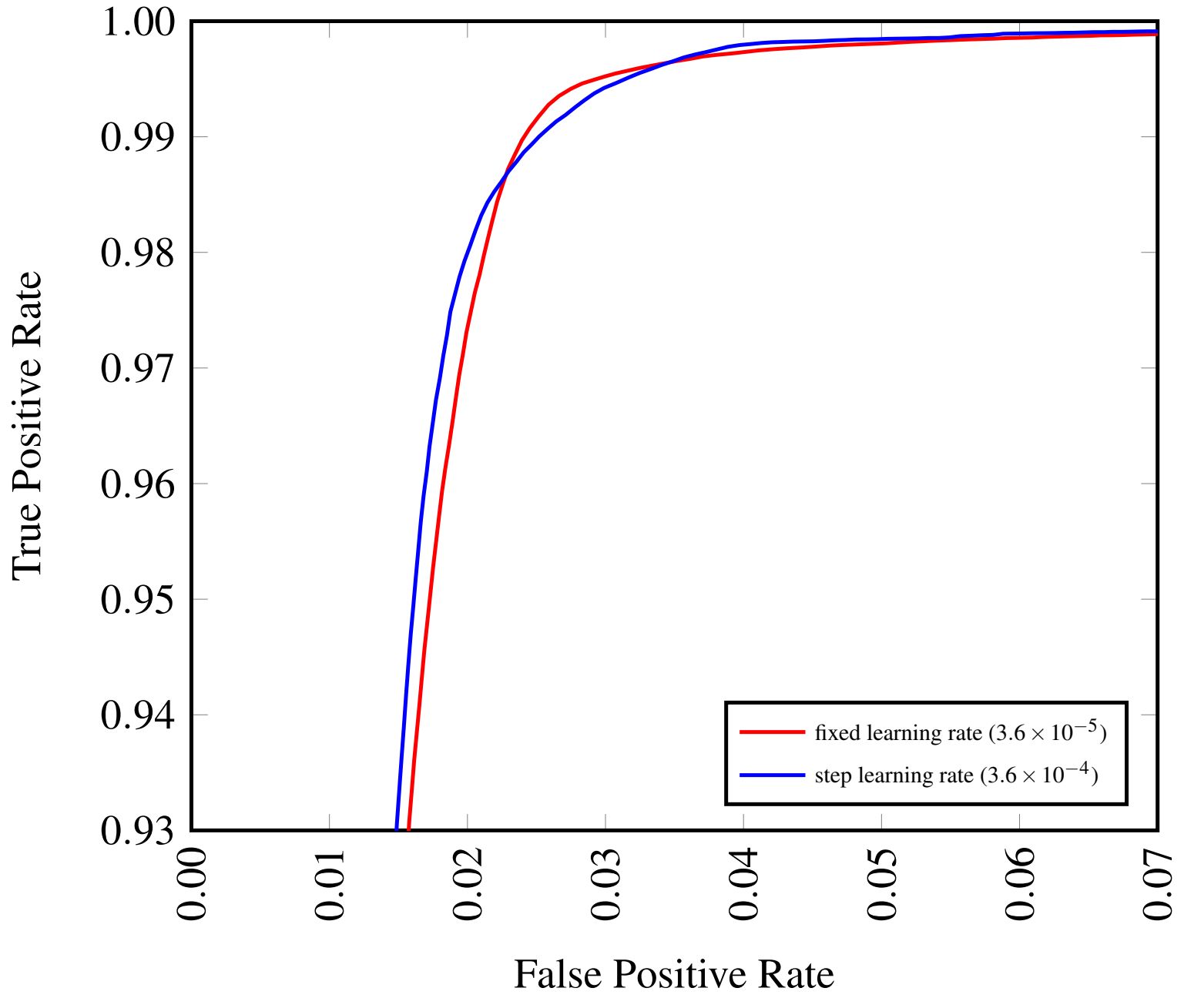

Figure 5.20.: Comparison between fixed and step learning rate policies.

| Parameter | Value |
|---|---|
| Input | $Occ_{free}$ - $v_{x,norm}$ - $v_{y,norm}$ |
| Input Range | $[-20, 20]$ |
| Network Structure | FCN-32s |
| Crop Size | $600 \times 600$ |
| Learning Rate | step policy with $3.6 \times 10^{-4}$ |
| Weight Factor | evaluated in this section |

Table 5.8.: Network configuration for optimization of the weight matrix.

Note that the factor for the non-moving grid cells is set to $c^{(2)} = 1$.

The result of the weight matrix adaption is shown in Figure 5.21 and Figure 5.22. There, it is shown that the factor $c^{(1)} = 1$ has the lowest performance and that the quality of the result increase with a growing factor. After a factor of $c^{(1)} = 40$ the quality decreases until the factor $c^{(1)} = 100$. Then it increases slightly before it decreases again. That result demonstrates the improvement of the quality by using a factor to solve the imbalance of the labeled data. Especially when the imbalance is as high as in this thesis, it is important to weight the labels. Otherwise the CNN would learn that the background is most important, as shown in Figure 5.23. There the difference between a factor of 1, 40 and 180 can be recognized. With an increasing factor the blobs around the moving objects become larger, but also false positive blobs appear. That is the reason why the factor has to be adapted to the database.

As a result the factor matrix

$$C = [1\ 40] \tag{5.2}$$

will be used for the final comparison.

## 5.2. Orientation Extraction

In this section the orientation extraction of Section 3.2 is evaluated. The configuration of the network is defined in Table 5.9 and is created by the optimized network of Section 5.1. The development of the learning process can be recognized in Figure 5.24. There, the two approaches for the orientation extraction (extraction over the CNN directly and over the velocity information) can be recognized. The figure shows the result of each approach against the ground truth data with the mean difference to the ground truth and the deviation of the difference to the ground truth. The ground truth is generated as described in Section 4.1.2. The goal is to reach a mean difference of zero and a deviation of zero.

The results show that after large changes (until $\approx 80\,000$ iterations), the CNN converges to a better result than the velocity approach. The large changes are caused by the first weight adaption, where the segmentation part (see Section 5.1) and the angle extraction part of the CNN generate weight updates separately. This produces different weight adaption, which converges over time. The segmentation part of the CNN is used to create clusters for the orientation extraction, as descibed in Section 3.2. Caused by the weight adaption of this part, the orientation extraction with the velocity changes in the first iterations (until $\approx 100\,000$) and stays relatively constant afterwards. The training in this situation stops at $200\,000$ iterations. By observing the development of the training, the orientation extraction of the CNN could converge to a better result by training more iterations, what could be evaluated in future work.

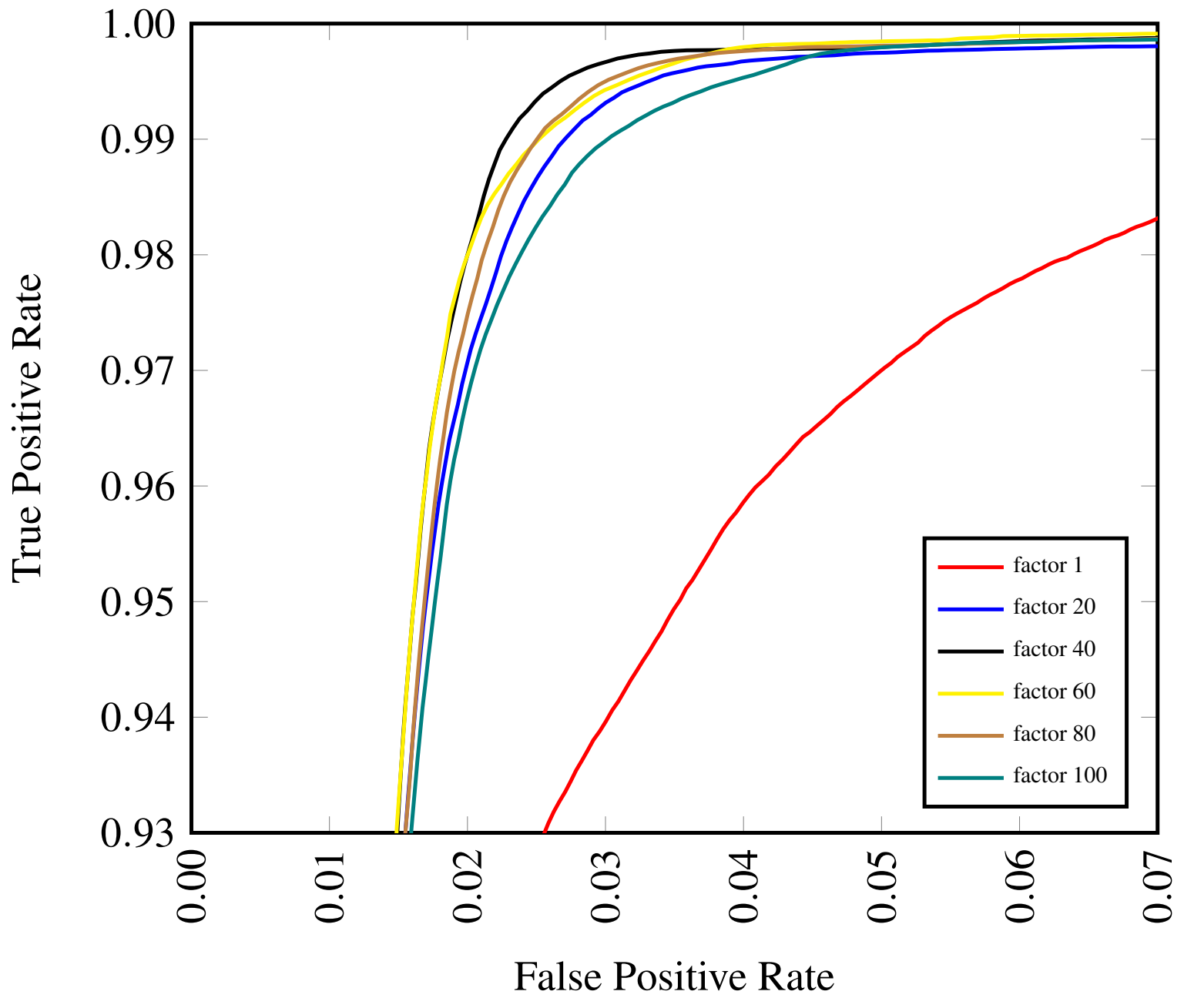

Figure 5.21.: ROC curves of different weight factor configurations in the range $[1, 100]$.

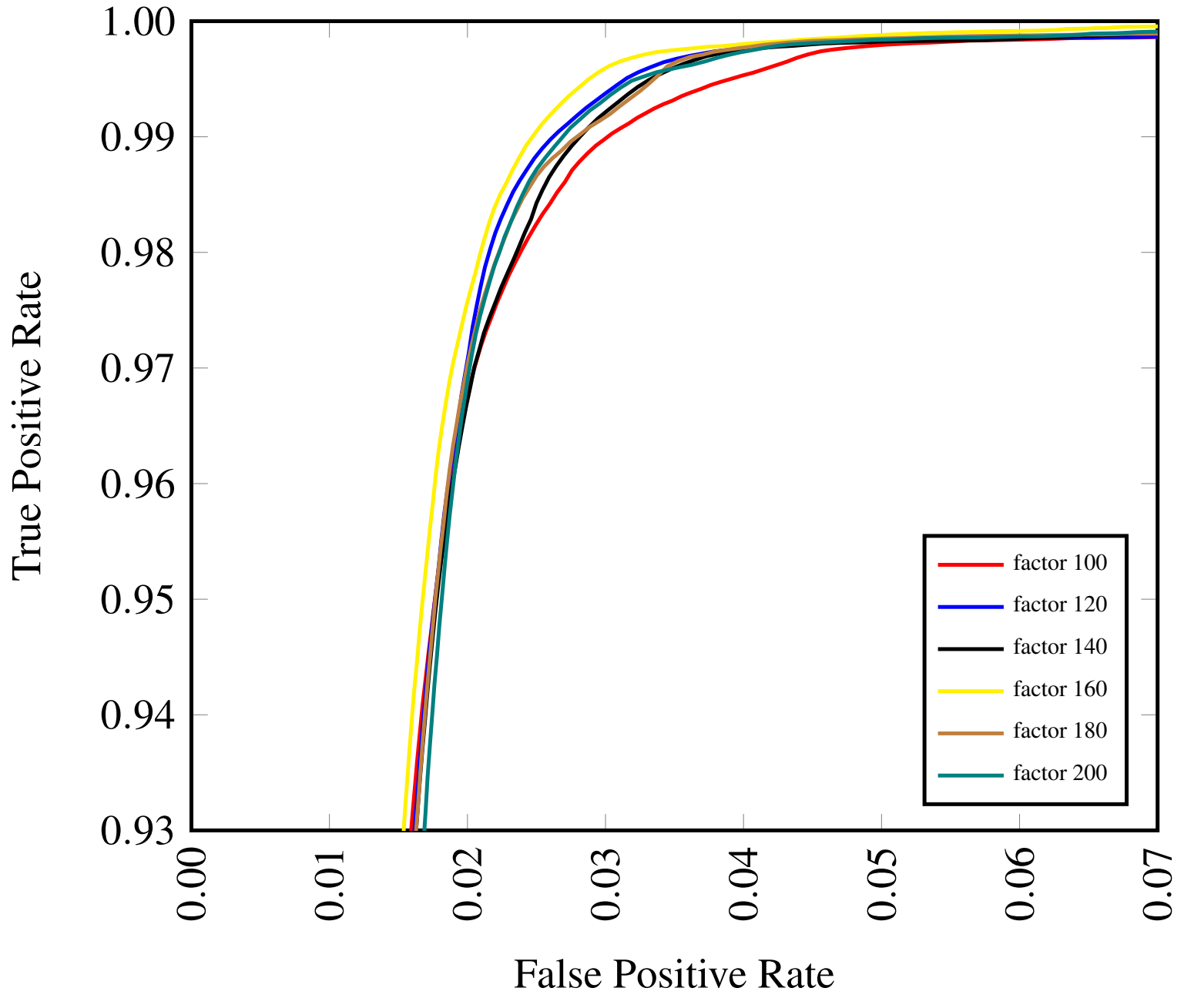

Figure 5.22.: ROC curves of different weight factor configurations in the range $[100, 200]$.

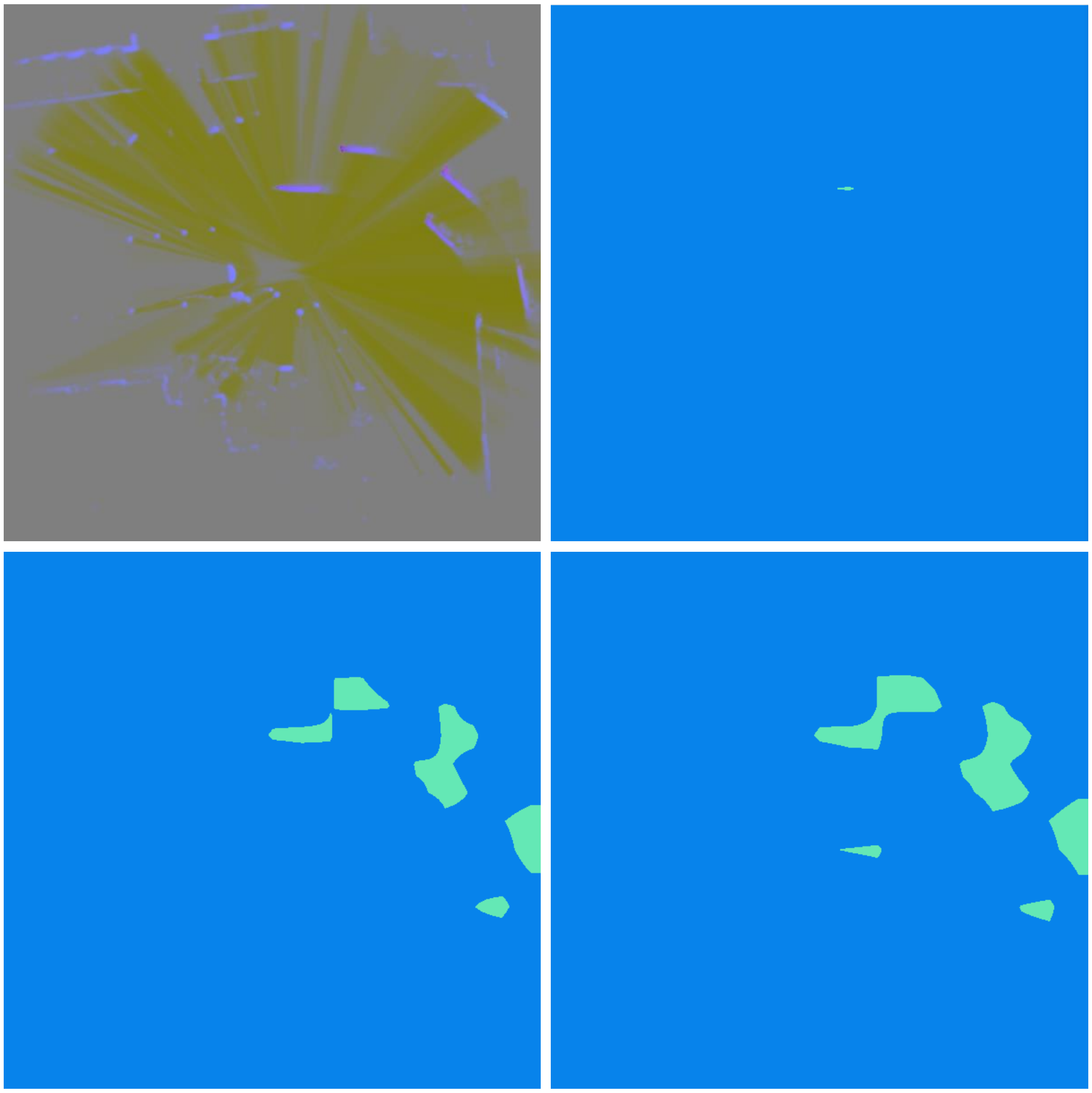

Figure 5.23.: Example of different weight matrices for one input image: top left: input image, top right: weight factor $c^{(1)} = 1$, bottom left: weight factor $c^{(1)} = 40$, bottom right: weight factor $c^{(1)} = 180$. By using a weight factor $c^{(1)} = 1$ the CNN interpret the background as most important. With an increasing weight factor, the blobs around the moving objects become larger, but also false positives appear.

| Parameter | Value |
|---|---|
| Input | $Occ_{free}$ - $v_{x,norm}$ - $v_{y,norm}$ |
| Input Range | $[-20, 20]$ |
| Network Structure | FCN-32s |
| Crop Size | $600 \times 600$ |
| Learning Rate | step policy with $3.6 \times 10^{-4}$ |
| Weight Factor | 40 |

Table 5.9.: Network configuration with the optimized parameters of Section 5.1.

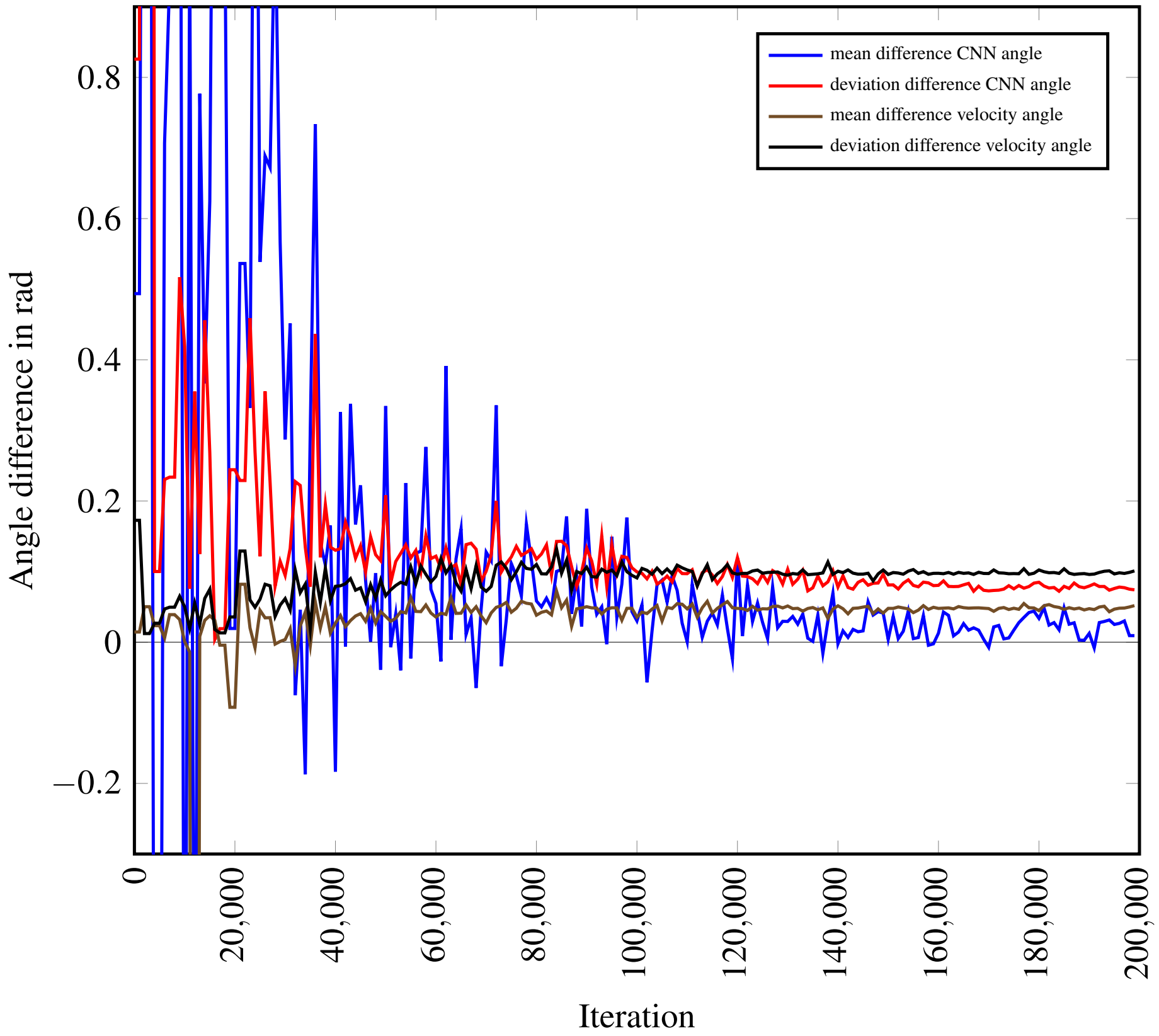

Figure 5.24.: Development of the angle extraction over the iteration number.

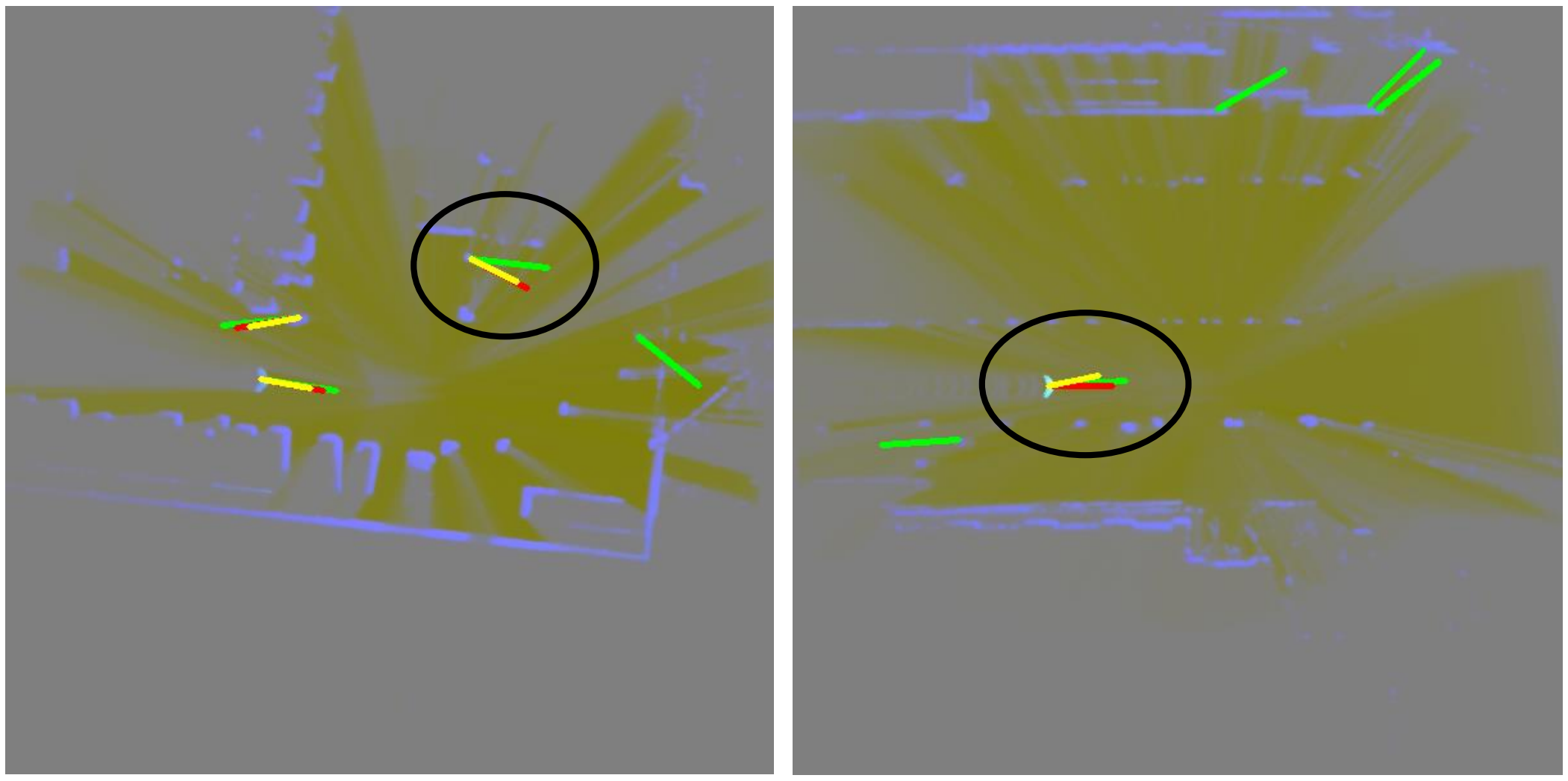

Figure 5.25.: Examples of the orientation extraction: green line: orientation extraction with the CNN, red line: ground truth, yellow line: orientation extraction with the velocity information, green line without comparison to the ground truth: false positive in the segmentation. The CNN seams to have problems with small slow moving objects (left image), but it uses the occupancy information to extract a better orientation with uncertainty velocity information (right image).

Figure 5.25 shows two examples of the orientation extraction, where the green line displays the orientation output of the CNN, the red line presents the ground truth, and the yellow line indicates the velocity approach. Clusters with only a green line are false positives, where no ground truth is available. The CNN seems to have problems with small slow moving objects like pedestrians. There, a large difference of the CNN to the ground truth can be recognized. The velocity approach shows nearly no differences for these kind of objects. On the other side objects with uncertain velocity information exist. This results in a lower performance of the orientation extraction with the velocity approach. In this example the CNN can use the occupancy information to extract the right orientation from the appearance of the object, as shown in Figure 5.25.

All in all the orientation extraction shows a better result with the CNN approach. Consequently, tracking algorithms can be improved by using the orientation information directly extracted from the CNN.

## 5.3. Semi-Supervised Learning

In this section, the proposed semi-supervised learning algorithm of Section 3.3 is evaluated. First, an algorithm according to Scudder et al. [40] without a false positive suppression is evaluated. For this purpose, the best network configuration of Section 5.1 (see Table 5.9) is used and a threshold for the CNN is extracted from the corresponding ROC curve. The automated labeling is then executed with a recording, which is different to the manually created recording of Section 4.1.1. This recording produces 9466 additional images for the training, which are again rotated to receive 340776 input images. From these automatically labeled images, every fifth image is used in the first step to be merged with the manual annotated dataset. The result can be seen in Figure 5.26 together with the non semi-supervised learning approach. The number of iterations to train the CNN is adapted accordingly to the number of training samples. The result shows, that the automated labeled data without a false positive suppression does not improve the result of the CNN. This can be caused by too many false positives, which are produced by the automated labeling.

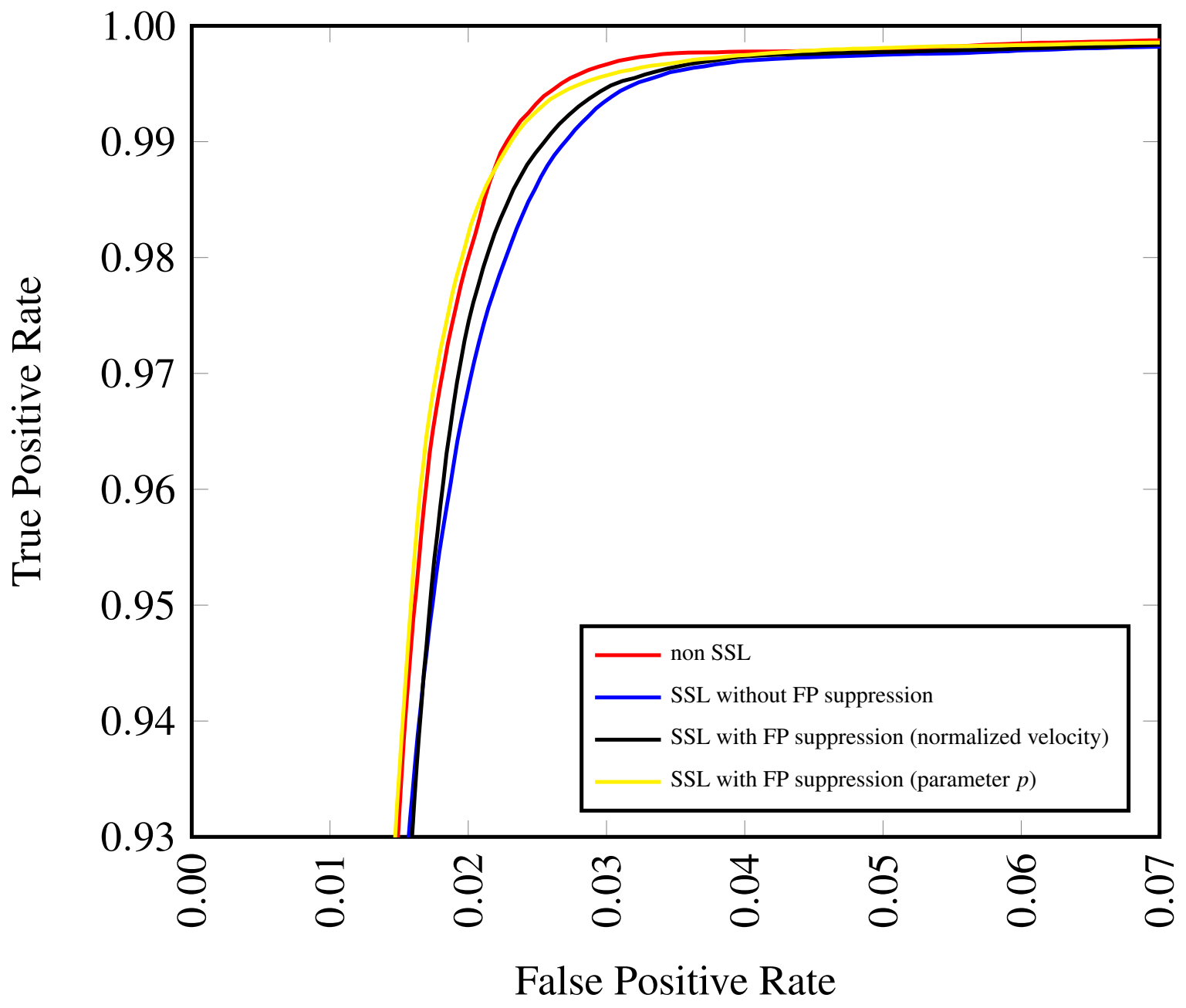

Figure 5.26.: ROC curves of the semi supervised learning approach.

To improve the semi-supervised learning approach a false positive suppression is inserted. For this purpose, the automated labels are clustered and a threshold for the normalized velocity and for the combination $p$ of the mahalanobis distance and the mean velocity is extracted to remove false positives, as described in Section 3.3. Additionally, small clusters are removed. The result is shown in Figure 5.26 as well, where the false positive suppression with the normalized velocity has a lower performance than the false positive suppression with the combined parameter $p$ by using every fifth automated labeled image. The suppression with the combined parameter has nearly the same result as the non-semi-supervised approach. That means the produced labeled data has nearly the same quality as the manually created data. By evaluating the changes of the output labels, it can be observed that the labels of the semi-supervised approach create on the one side a better output by suppressing small false positive clusters (see Figure 5.27). This can be recognized at small curb stones. On the other side false positives are inserted (see Figure 5.28). These false positives could be the reason why the usage of all the automated labeled data in combination with the manual created data generates no better results.

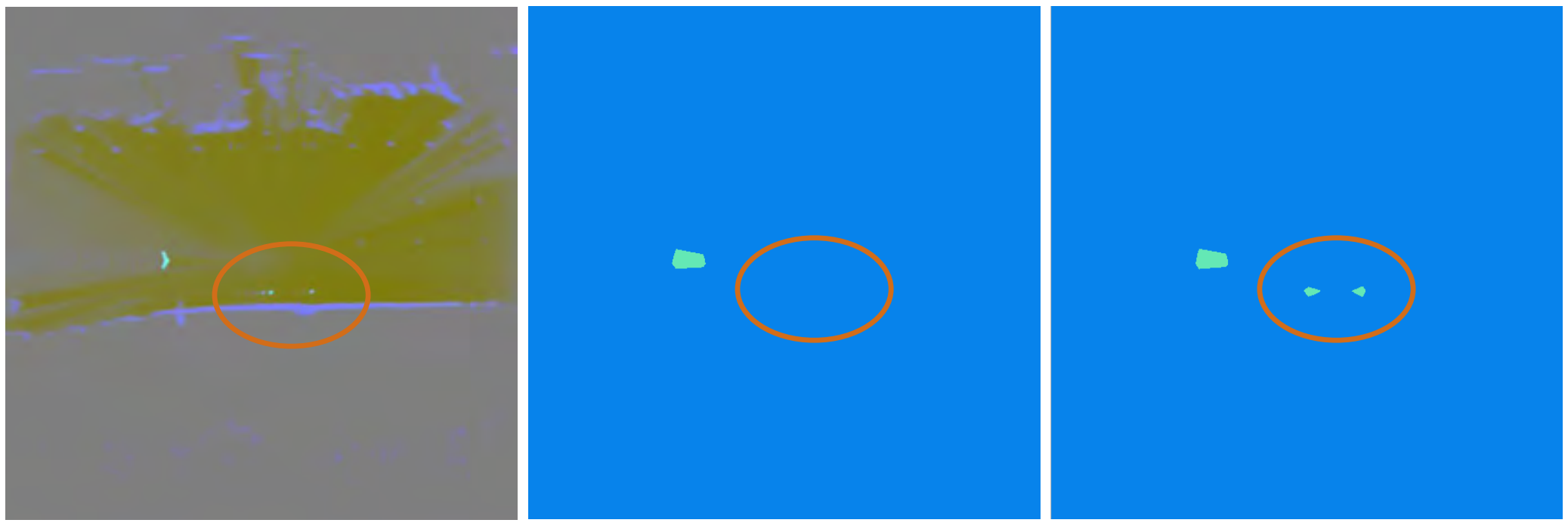

Figure 5.27.: Suppression of small curb stones with the semi-supervised learning approach: left: input image, middle: output of the semi-supervised learning approach, right: output of the CNN without semi-supervised training.

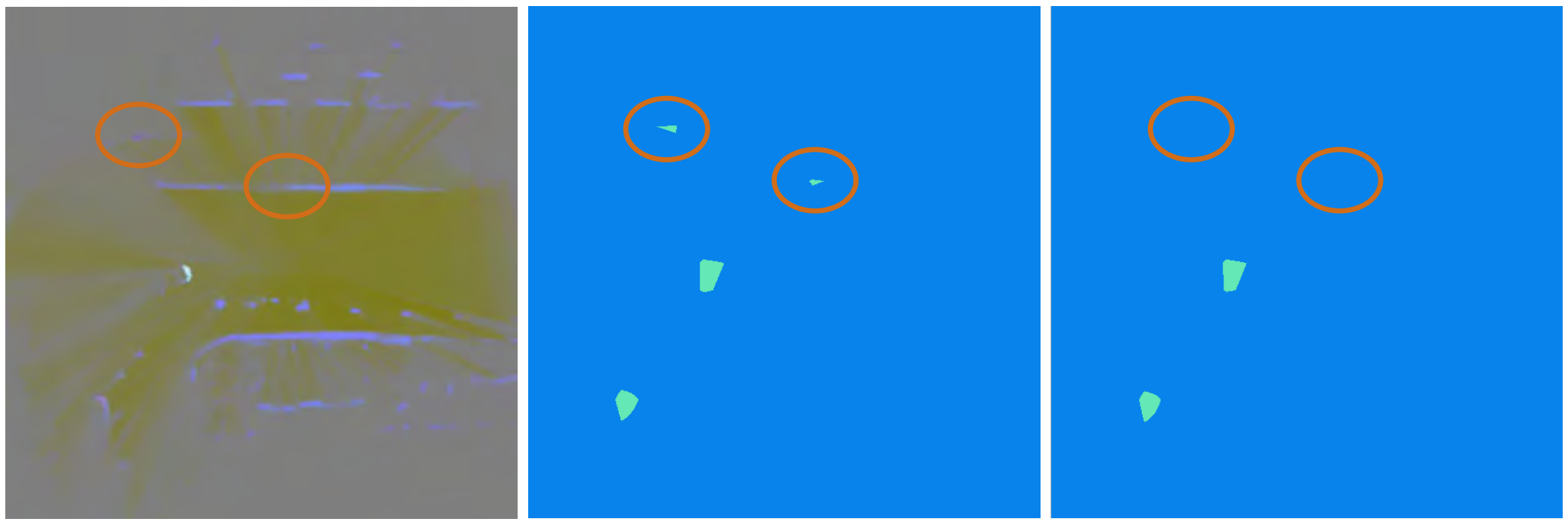

Figure 5.28.: Insertion of false positives with the SSL approach: left: input image, middle: output of the semi-supervised learning approach, right: output of the CNN without semi-supervised training.

For further reducing the false positive, the threshold for the probability of the CNN to generate the labeled images is variated, what can be recognized in Figure 5.29. There, the best result is still the optimal threshold of Section 5.1. That means the number of false positives may be reduced with a variation of the threshold, but the number of false negatives increases instead, what produces a lower overall performance.

Additionally the training dataset is increased by using every automatic annotated image instead of every fifth, to see the influence of the labeled data. The result is shown in Figure 5.30, where the number of iterations is adapted according to the number of training data. As a result, the automatic labeled data creates a lower performance, by introducing more of them in the training process. That proves the suggestion, that the false positives of the automatic labeled data have a negative effect on the training process.

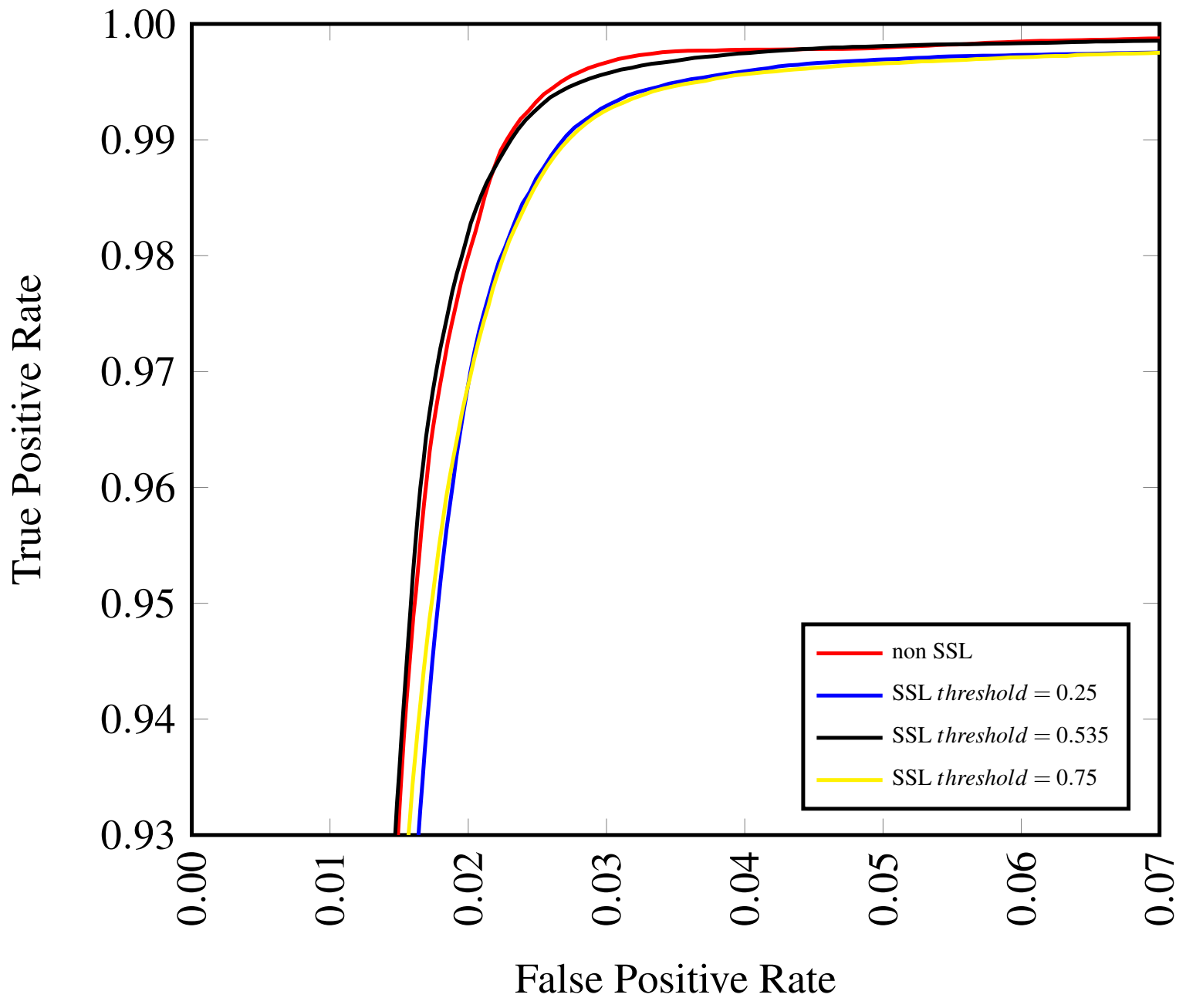

Figure 5.29.: ROC curves of different threshold of the CNN probability with false positive suppression.

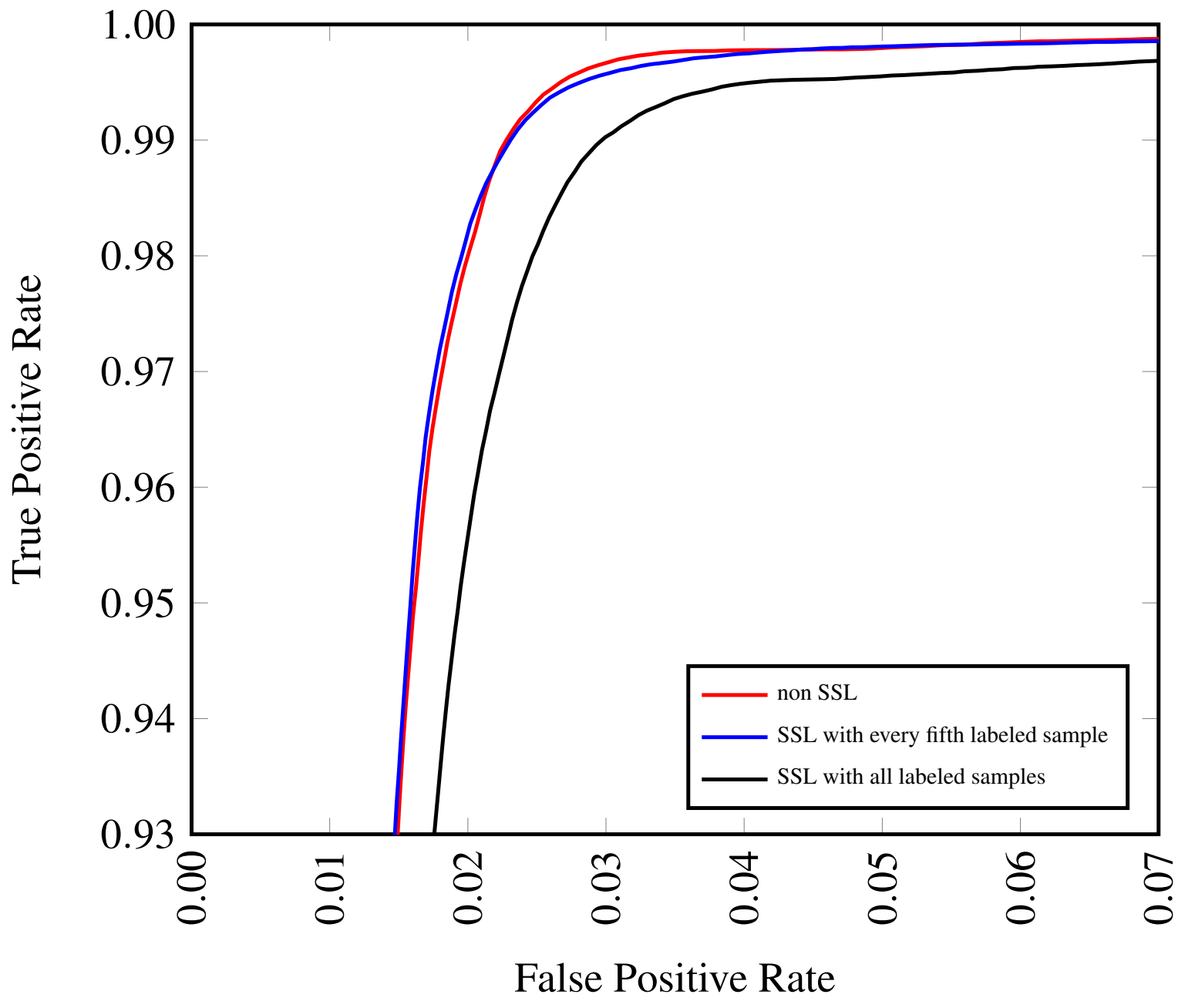

Figure 5.30.: ROC curves of different number of generated training data with the semi-supervised learning approach ($threshold = 0.535$).

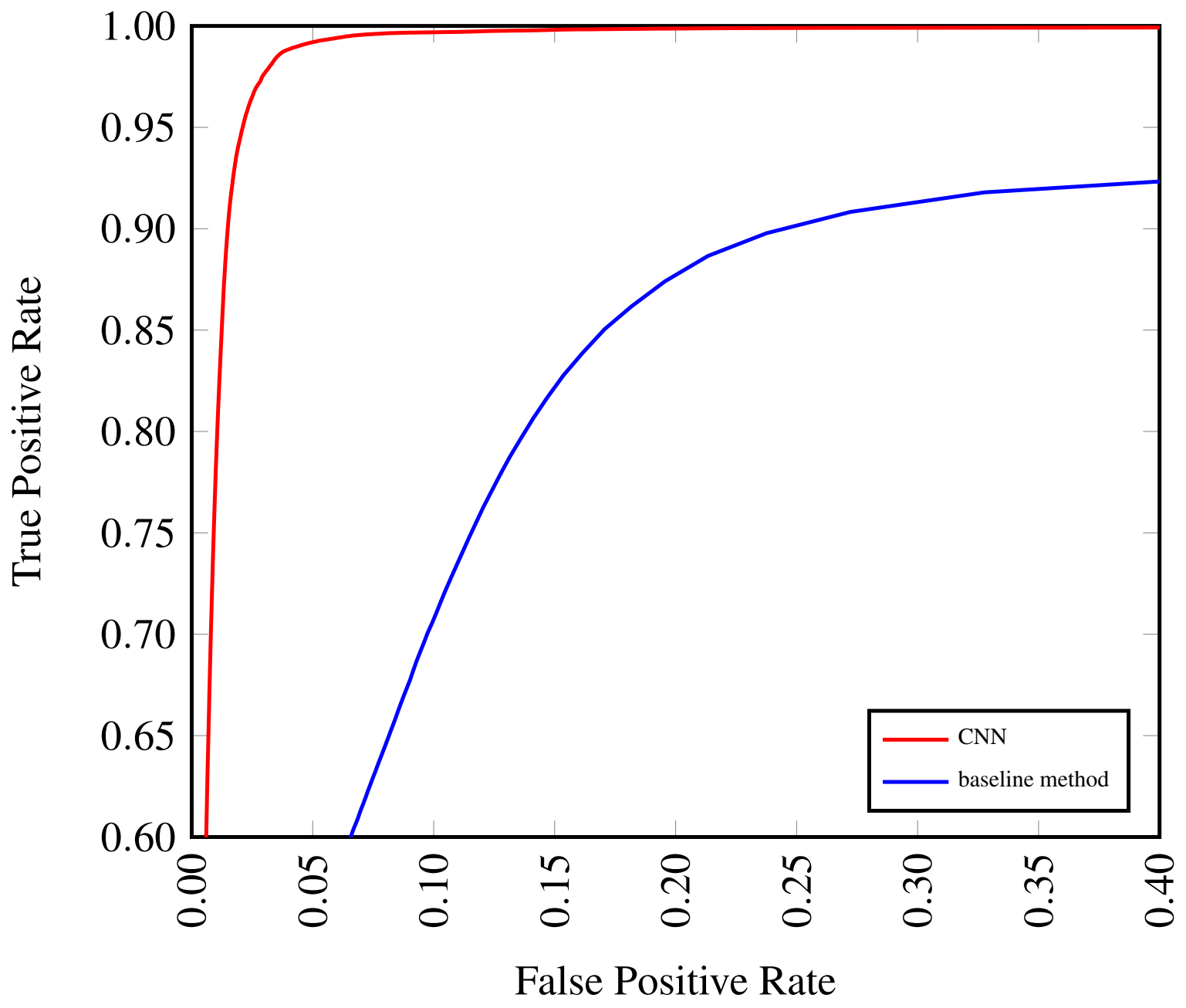

Figure 5.31.: ROC curves of the final comparison with Nuss et al. [36] as baseline method.

As a result, the semi-supervised learning approach of this thesis cannot be used for improvement of the CNN. For this reason, the semi-supervised learning approach is not used for the final comparison.

## 5.4. Comparison to other Approaches

For the final comparison to the baseline method, which is described in Section 4.2, the optimized network configuration of the segmentation evaluation (see Table 5.9) is used. The CNN is evaluated against the baseline method introduced by Nuss et al. [36] by distinguishing dynamic and static grid cells. The results are shown in Figure 5.31, where a general improvement can be recognized. The accuracy of 72.1% for the baseline method and 97.2% for the CNN approach, results in a relative improvement of 34.8%. That means a better performance by using the appearance of an object while introducing the occupancy information in the decision process of movement.

The baseline method creates a decision for each grid cell independently. That results in a complete rotation invariance of objects. The CNN internally uses filters at the convolutional layer of a specific size, which results in a specific field of view. That can affect the rotation invariance of a CNN. For this reason the input images are rotated for the training as described in Section 4.1.4. To examine the rotation invariance in this section, the test input images are rotated as well, where one test sample is shown in Figure 5.32. The figure shows that the precision, recall and the accuracy do not essentially change over the different rotation angles. Thus, a rotation invariance is proved. Additionally it can be recognized, that the precision and the recall have in a distance of 90 degrees the same curve progression. That is caused by the filters of the convolutional layers, which are squared. By rotating the input image, the squared filter changes the field of view caused by the corners of the square. To reduce this and enforce the rotation invariance, a circled filter shape can be used in future works.

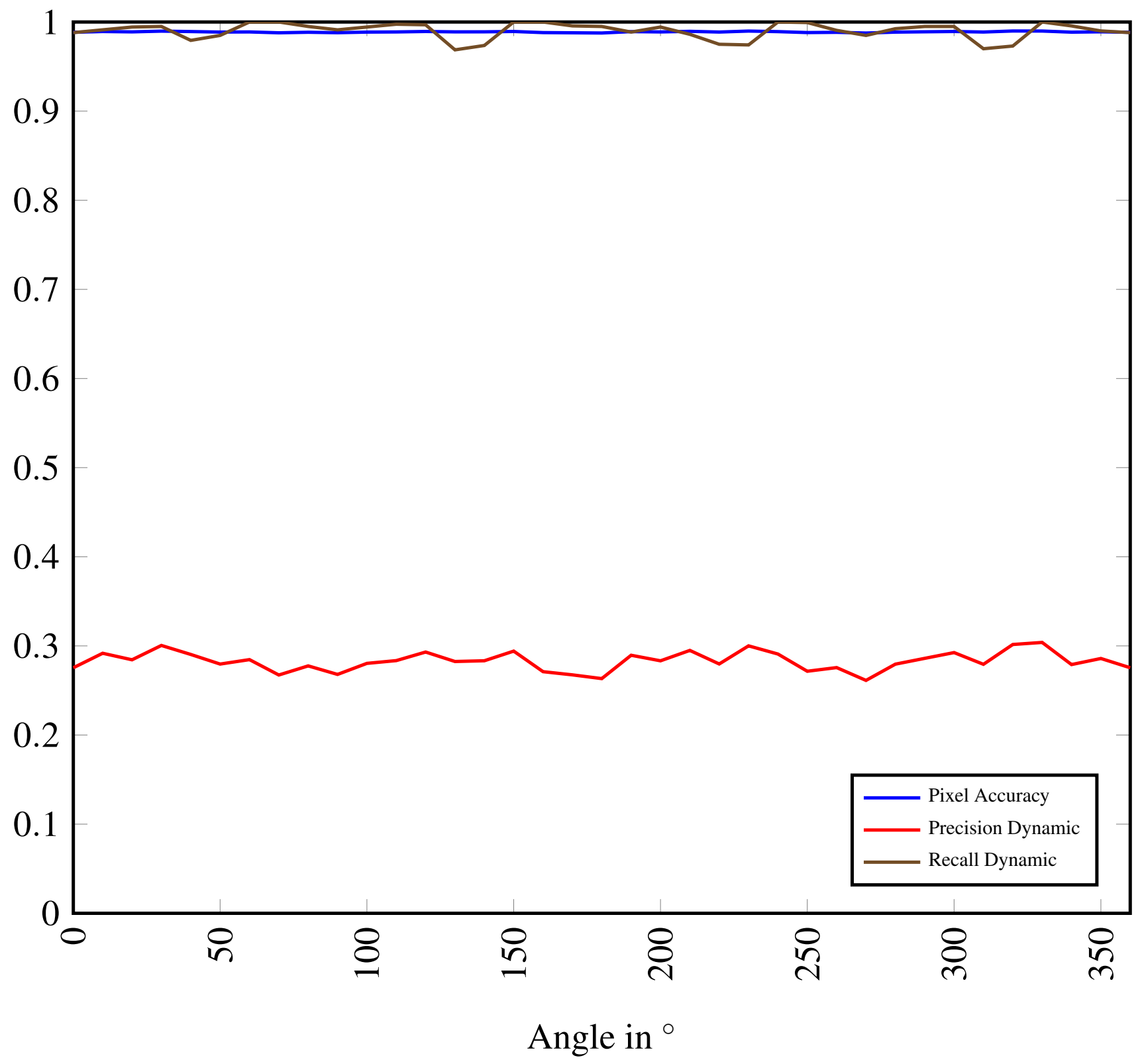

Figure 5.32.: Evaluation of orientation invariance by rotating one input image of the test dataset.

# 6. Conclusion

An elementary part of environment perception in the field of autonomous driving is tracking of objects. For this purpose a distinction between dynamic and static objects has to be achieved, what is made a subject in this thesis. The differentiation is based on a dynamic occupancy grid map (DOG), where for each grid cell a decision has to be made.

In contrast to other approaches, this thesis uses Convolutional Neural Networks (CNNs) to predict the class affiliation of each grid cell. To achieve that, ground truth datasets had to be created. For this purpose a semi-automatic labeling toolchain was implemented, where the mahalanobis distance of each grid cell was used to create clusters of dynamic grid cells. Afterwards the generated data was corrected manually to create the training, validation and test dataset. The primary goal was to use additional information like the occupancy information of the DOG to generate an appearance-based decision of the movement and improve other approaches.

The CNN was initialized with the pre-trained VGG-net to shorten the training time and to use specific filters of the pre-trained networks. Afterwards, the network was transferred to a Fully Convolutional Network (FCN) to create a pixelwise prediction of the movement. Caused by the transfer of the pre-trained network from video captured input to the DOG as input, the CNN had to be optimized at the following parameters:

- type of input data,
- input rage,
- structure of the network,
- zooming into the input,
- learning rate,
- weight matrix.

With this optimization, different problems like the imbalance of the training data or the different appearance of objects were solved. Additionally a smaller network structure (Alexnet) was evaluated to prove the real-time feasibility.

Another goal of this thesis was the orientation extraction of the detected objects to improve tracking algorithms with this information. This was done with two different approaches: the extraction of the orientation directly through the CNN and the extraction of the orientation through the velocity information for each detected moving cluster. For the prediction of the orientation angle with the CNN, the biternion representation was used, to solve the problem of periodicity. The results presented a better overall performance of the orientation extraction through the CNN.

A general problem while training machine learning approaches is the need of labeled data. For this reason a semi-supervised learning algorithm with false positive suppression was implemented. This algorithm created automatic labeled data based on a trained CNN. The results of the semi-supervised learning approach showed no improvement of the CNN when using additional automatic labeled data even after reducing the number of false positives.

Finally a comparison of the trained CNN with the approach of Nuss et al. [36] as a baseline method of distinction between static and dynamic grid cells was executed. There, a relative improvement of 34.8% for the accuracy was proven. Additionally a rotation invariance of the CNN was demonstrated, what allows a realization of this CNN in a real environment.

All in all, this thesis showed a feasible approach for grid cell based distinction between dynamic and static obstacles, which outperforms the current baseline method, which represents the state of the art. This could improve tracking algorithms in the field of autonomous driving, what helps to approach the vision zero.

# 7. Outlook

This thesis presented a general improvement of the approach of Nuss et al. [36] to distinguish between dynamic and static objects, but the result can still be improved. First, the network structure can be adapted or completely developed from scratch to create a network, which has on the one side a good performance and on the other side a real-time applicable execution time. This could be implemented by adapting the filter sizes according to the size of obstacles. Especially for recognition of an obstacle as a whole, compared to recognizing different parts, the filter size could be increased within the first layers. At the same time, the number of layers could be decreased, creating a shorter execution time. Especially the last convolutional layers (before the deconvolutional layer) could be removed, because these layers were introduced into the public network structures like VGG-net [41] or Alexnet [26] to create a classification of the whole image. This is not needed for a pixelwise classification. The rejection of these layers would also cause less upscaling in the end, what creates a more detailed output.

The adaption of the network structure would cause that pre-trained network cannot be used for initialization. For this purpose, an own pre-training has to be performed, what needs a larger number of training samples. To solve this problem, an unsupervised pre-training like an autoencoder could be used. This would create a problem-specific pre-trained network instead of using a pre-trained network, which was trained on another type of input data. As a result, discretization or range adaptions would not be needed anymore.

Furthermore, by training from scratch, the types of input can be combined in different ways to obtain a better result. For example the velocity of the recording vehicle could be provided as a separate input, what could improve the recognition of clutter caused by a fast moving environment. Also the restriction of three input channels would be removed, what enables a lot of possibilities for combinations of the inputs.

Additionally, the structure can be adapted, as discussed in Section 5.4, with circular filter shapes. That could improve the rotation invariance, which is an elementary requirement in street environment.

A general problem of training neural networks is the limited number of training samples. For this reason a larger training dataset can be created, by recording more data on different environments, like a highway with a high speed limit or different cities, where the appearance of the environment changes. This should further improve the results of the CNN. Also the labeling process itself can be improved by using further semi-automatic labeling approaches, which use the CNN for prelabeling new recordings. This would create more labeled data by a supervision approach, what could improve the performance of the CNN. For example a recording with non-moving objects can be recorded, what should reduce the false positives. For this kind of recording the labeled data can easily be generated.

Finally, the semi-supervised learning (SSL) approach can be improved to generate automatic labeled data for increasing the performance of the CNN. Especially the false positive rejection can be improved to create labeled data with a higher quality.

# A. List of Figures

# B. List of Tables

# C. Bibliography

[1] A. Asvadi, P. Peixoto, and U. Nunes. Detection and Tracking of Moving Objects Using 2.5D Motion Grids. In *IEEE International Conference on Intelligent Transportation Systems (ITSC)*, 2015.

[2] R. H. Baxter, M. J. V. Leach, S. S. Mukherjee, and N. M. Robertson. An Adaptive Motion Model for Person Tracking with Instantaneous Head-Pose Features. *IEEE Signal Processing Letters*, 22(5):578–582, 2015.

[3] L. Beyer, A. Hermans, and B. Leibe. Biternion Nets: Continuous Head Pose Regression from Discrete Training Labels. In *German Conference on Pattern Recognition (GCPR)*, 2015.

[4] O. Chapelle, B. Schoelkopf, and A. Zien. *Semi-Supervised Learning*. MIT Press, Massachusetts, 2006.

[5] C. Chen and J. M. Odobez. We are not Contortionists: Coupled Adaptive Learning for Head and Body Orientation Estimation in Surveillance Video. In *IEEE Conference on Computer Vision and Pattern Recognition (CVPR)*, 2012.

[6] L.-C. Chen, G. Papandreou, I. Kokkinos, K. Murphy, and A. L. Yuille. Semantic Image Segmentation with Deep Convolutional Nets and Fully Connected CRFs. In *International Conference on Learning Representations (ICLR)*, 2015.

[7] D. C. Ciresan, A. Giusti, L. M. Gambardella, and J. Schmidhuber. Deep Neural Networks Segment Neuronal Membranes in Electron Microscopy Images. In *Conference on Neural Information Processing Systems (NIPS)*, 2012.

[8] P. Cloutier, C. Tibirna, B. P. Grandjean, and J. Thibault. NNFit version 2.0 User's Manual, 1997.

[9] M. Cordts, M. Omran, S. Ramos, T. Rehfeld, M. Enzweiler, R. Benenson, U. Franke, S. Roth, and B. Schiele. The Cityscapes Dataset for Semantic Urban Scene Understanding. In *IEEE Conference on Computer Vision and Pattern Recognition (CVPR)*, 2016.

[10] M. Cordts, M. Omran, S. Ramos, T. Scharwächter, M. Enzweiler, R. Benenson, U. Franke, S. Roth, and B. Schiele. The Cityscapes Dataset. In *IEEE Conference on Computer Vision and Pattern Recognition (CVPR)*, 2015.

[11] C. Coue, C. Pradalier, C. Laugier, T. Fraichard, and P. Bessiere. Bayesian Occupancy Filtering for Multitarget Tracking: An Automotive Application. *The International Journal of Robotics Research*, 25(1):19–30, 2006.

[12] J. Dai, K. He, and J. Sun. BoxSup: Exploiting Bounding Boxes to Supervise Convolutional Networks for Semantic Segmentation. In *IEEE International Conference on Computer Vision (ICCV)*, 2015.

[13] J. Deng, W. Dong, R. Socher, L.-J. Li, K. Li, and L. Fei-Fei. ImageNet: A Large-Scale Hierarchical Image Database. In *IEEE Conference on Computer Vision and Pattern Recognition (CVPR)*, 2009.

[14] E. Dickmanns, R. Behringer, D. Dickmanns, T. Hildebrandt, M. Maurer, F. Thomanek, and J. Schiehlen. The Seeing Passenger Car 'VaMoRs-P'. In *IEEE Intelligent Vehicles Symposium (IV)*, 1994.

[15] E. D. Dickmanns, B. Mysliwetz, and T. Christians. An Integrated Spatio-Temporal Approach to Automatic Visual Guidance of Autonomous Vehicles. In *IEEE Transactions on Systems, Man and Cybernetics (SMC)*, 1990.

[16] A. Elfes. Using Occupancy Grids for Mobile Robot Perception and Navigation. *IEEE Computer Magazine*, 22(6):46–57, 1989.

[17] M. Ester, H. P. Kriegel, J. Sander, and X. Xu. A Density-Based Algorithm for Discovering Clusters in Large Spatial Databases with Noise. In *International Conference on Knowledge Discovery and Data Mining (KDD)*, 1996.

[18] M. Everingham, L. Van Gool, C. K. I. Williams, J. Winn, and A. Zisserman. The PASCAL Visual Object Classes Challenge, 2012.

[19] P. Fischer, A. Dosovitskiy, E. Ilg, P. Haeusser, C. Hazirbas, V. Golkov, P. van der Smagt, D. Cremers, and T. Brox. FlowNet: Learning Optical Flow with Convolutional Networks. In *IEEE International Conference on Computer Vision (ICCV)*, 2015.

[20] U. Franke, S. Mehring, A. Suissa, and S. Hahn. The Daimler-Benz Steering Assistant: a Spin-off from Autonomous Driving. In *IEEE Intelligent Vehicles Symposium (IV)*, 1994.

[21] K. Fukushima. Neocognitron: A Self-organizing Neural Network Model for a Mechanism of Pattern Recognition Unaffected by Shift in Position. *Biological Cybernetics*, 36(4):193–202, 1980.

[22] R. Girshick, J. Donahue, T. Darrell, and J. Malik. Region-based Convolutional Networks for Accurate Object Detection and Segmentation. *IEEE Transactions on Pattern Analysis and Machine Intelligence (TPAMI)*, 38(1):142–158, 2016.

[23] I. J. Goodfellow, Y. Bengio, and A. Courville. Deep Learning. *Book in preparation for MIT Press available at http://www.deeplearningbook.org*, 2016.

[24] B. Hariharan, P. Arbel, and R. Girshick. Hypercolumns for Object Segmentation and Fine-grained Localization. In *IEEE Conference on Computer Vision and Pattern Recognition (CVPR)*, 2015.

[25] A. Krizhevsky. *Learning Multiple Layers of Features from Tiny Images*. Masterthesis, University of Toronto, 2009.

[26] A. Krizhevsky, I. Sulskever, and G. E. Hinton. ImageNet Classification with Deep Convolutional Neural Networks. In *Conference on Neural Information Processing Systems (NIPS)*, 2012.

[27] Y. LeCun, L. Bottou, Y. Bengio, and P. Haffner. Gradient-Based Learning Applied to Document Recognition. *Proceedings of the IEEE (Journal)*, 86(11):2278–2324, 1998.

[28] J. Long, E. Shelhamer, and T. Darrell. Fully Convolutional Networks for Semantic Segmentation. In *IEEE Conference on Computer Vision and Pattern Recognition (CVPR)*, 2015.

[29] J. Masci, U. Meier, D. Ciresan, and J. Schmidhuber. Stacked Convolutional Auto-Encoders for Hierarchical Feature Extraction. In *International Conference on Artificial Neural Networks (ICANN)*, 2011.

[30] O. Matan, C. J. C. Burges, Y. LeCun, and J. S. Denker. Multi-Digit Recognition Using a Space Displacement Neural Network. In *Conference on Neural Information Processing Systems (NIPS)*, 1991.

[31] D. Maturana and S. Scherer. VoxNet: A 3D Convolutional Neural Network for Real-Time Object Recognition. In *IEEE/RSJ International Conference on Intelligent Robots and Systems (IROS)*, 2015.

[32] T. Mitchell and A. Blum. Combining Labeled and Unlabeled Data with Co-Training. In *Conference on Computational Learning Theory (COLT)*, 1998.

[33] T. M. Mitchell. The Role of Unlabeled Data in Supervised Learning. In *International Colloquium on Cognitive Science (ICCS)*, 1999.

[34] A. Y. Ng, J. Ngiam, C. Y. Foo, Y. Mai, and C. Suen. Unsupervised Feature Learning and Deep Learning Tutorial - Softmax Regression, 2013.

[35] K. Nigam, A. Mccallum, S. Thrun, and T. Mitchell. Learning to Classify Text from Labeled and Unlabeled Documents. In *Conference on Innovative Applications of Artificial Intelligence (IAAI)*, 1998.

[36] D. Nuss, S. Reuter, M. Thom, T. Yuan, G. Krehl, M. Maile, A. Gern, and K. Dietmayer. A Random Finite Set Approach for Dynamic Occupancy Grid Maps with Real-Time Application. *ArXiv e-prints*, 2016.

[37] D. Nuss, T. Yuan, G. Krehl, M. Stuebler, S. Reuter, and K. Dietmayer. Fusion of Laser and Radar Sensor Data with a Sequential Monte Carlo Bayesian Occupancy Filter. In *IEEE Intelligent Vehicles Symposium (IV)*, 2015.

[38] G. Papandreou, L.-C. Chen, K. Murphy, and A. L. Yuille. Weakly- and Semi-Supervised Learning of a DCNN for Semantic Image Segmentation. *IEEE International Conference on Computer Vision (ICCV)*, 2015.

[39] J. Peukert. *Object Generation from a Dynamic Occupancy Gridmap Considering Autocorrelation*. Masterthesis, Karlsruhe Institute of Technology, 2016.

[40] H. Scudder. Probability of Error of Some Adaptive Pattern-Recognition Machines. *IEEE Transactions on Information Theory*, 11(3):363–371, 1965.

[41] K. Simonyan and A. Zisserman. Very Deep Convolutional Networks for Large-Scale Image Recognition. In *International Conference on Learning Representations (ICLR)*, 2015.

[42] C. Szegedy, W. Liu, Y. Jia, P. Sermanet, S. Reed, D. Anguelov, D. Erhan, V. Vanhoucke, and A. Rabinovich. Going Deeper with Convolutions. In *IEEE Conference on Computer Vision and Pattern Recognition (CVPR)*, 2015.

[43] R. Szeliski. *Computer Vision: Algorithms and Applications*. Springer Science & Business Media, London, 2010.

[44] The California Department of Motor Vehicles (DMV). Autonomous Vehicle Disengagement Reports, 2016.

[45] S. J. Thorpe and M. Fabre-Thorpe. Seeking Categories in the Brain. *Science*, 291(5502):260–263, 2001.

[46] S. Thrun, W. Burgard, and D. Fox. *Probabilistic Robotics (Intelligent Robotics and Autonomous Agents)*. MIT Press, Cambridge, 2005.

[47] S. Thrun, M. Montemerlo, H. Dahlkamp, D. Stavens, A. Aron, J. Diebel, P. Fong, J. Gale, M. Halpenny, and G. Hoffmann. Stanley, the Robot that Won the DARPA Grand Challenge. *Journal of Field Robotics*, 23(9):661–692, 2006.

[48] C. Tingvall and N. Haworth. Vision Zero - An ethical approach to safety and mobility. In *Road Safety & Traffic Enforcement*, 1999.

[49] C. Urmson, C. Baker, J. Dolan, P. Rybski, B. Salesky, W. Whittaker, D. Ferguson, and M. Darms. Autonomous Driving in Traffic: Boss and the Urban Challenge. *AI Magazine*, 30(2):17–28, 2009.

[50] T.-D. Vu, O. Aycard, and N. Appenrodt. Online Localization and Mapping with Moving Object Tracking in Dynamic Outdoor Environments. In *IEEE Intelligent Vehicles Symposium (IV)*, 2007.

[51] C. Wang, C. Thorpe, and S. Thrun. Online Simultaneous Localization and Mapping with Detection and Tracking of Moving Objects: Theory and Results from a Ground Vehicle in Crowded Urban Areas. In *IEEE International Conference on Robotics and Automation (ICRA)*, 2003.

[52] Q. Wang, J. Zhang, X. Hu, and Y. Wang. Automatic Detection and Classification of Oil Tanks in Optical Satellite Images Based on Convolutional Neural Network. In *International Conference on Image and Signal Processing (ICISP)*, 2016.

[53] J. Wieczner. Why self-driving cars are crashing: humans. *Time Inc.*, 2015.

[54] R. Wolf and J. C. Platt. Postal Address Block Location Using A Convolutional Locator. In *Conference on Neural Information Processing Systems (NIPS)*, 1994.

[55] World Health Organization (WHO). *Global Status Report On Road Safety*. World Health Organization (WHO), Villars-sous-Yens, 2015.

[56] T. Yuan, B. Duraisamy, T. Schwarz, and M. Fritzsche. Track Fusion with Incomplete Information for Automotive Smart Sensor Systems. In *IEEE Radar Conference (RadarConf)*, 2016.

[57] M. D. Zeiler, D. Krishnan, G. W. Taylor, and R. Fergus. Deconvolutional Networks. In *IEEE Conference on Computer Vision and Pattern Recognition (CVPR)*, 2010.

[58] S. Zheng, S. Jayasumana, B. Romera-Paredes, V. Vineet, Z. Su, D. Du, C. Huang, and P. Torr. Conditional Random Fields as Recurrent Neural Networks. In *IEEE International Conference on Computer Vision (ICCV)*, 2015.

[59] J. Ziegler, P. Bender, M. Schreiber, H. Lategahn, T. Strauss, C. Stiller, T. Dang, U. Franke, N. Appenrodt, C. G. Keller, E. Kaus, R. G. Herrtwich, C. Rabe, D. Pfeiffer, F. Lindner, F. Stein, F. Erbs, M. Enzweiler, C. Knoppel, J. Hipp, M. Haueis, M. Trepte, C. Brenk, A. Tamke, M. Ghanaat, M. Braun, A. Joos, H. Fritz, H. Mock, M. Hein, and E. Zeeb. Making Bertha Drive - An Autonomous Journey on a Historic Route. *IEEE Intelligent Transportation Systems Magazine*, 6(2):8–20, 2014.